\documentclass{article} 
\usepackage{iclr2025_conference,times}
\iclrfinalcopy

\usepackage{amsmath,amsfonts,bm}

\def\eqref#1{equation~\ref{#1}}

\def\1{\bm{1}}

\def\ermX{{\textnormal{X}}}
\def\ermY{{\textnormal{Y}}}

\def\vtheta{{\bm{\theta}}}

\def\vg{{\bm{g}}}

\def\vm{{\bm{m}}}

\def\vv{{\bm{v}}}

\DeclareMathAlphabet{\mathsfit}{\encodingdefault}{\sfdefault}{m}{sl}
\SetMathAlphabet{\mathsfit}{bold}{\encodingdefault}{\sfdefault}{bx}{n}

\newcommand{\Var}{\mathrm{Var}}

\newcommand{\Cov}{\mathrm{Cov}}

\usepackage{url}
\usepackage{times}
\usepackage{epsfig}
\usepackage{graphicx}
\usepackage{amsmath}
\usepackage{multicol}
\usepackage{colortbl}
\usepackage{multirow}
\usepackage{subcaption}
\usepackage{makecell}
\usepackage{amssymb}
\usepackage{setspace} 
\usepackage{comment}
\usepackage{wrapfig}
\usepackage{algorithm} 
\usepackage{caption}
\usepackage{float}
\usepackage{caption}
\usepackage{algorithmic}
\usepackage{mathtools}
\usepackage{xcolor}
\usepackage{pythonhighlight}
\usepackage{tabularx}
\definecolor{gray}{gray}{0.95}
\definecolor{gray3}{gray}{0.90}
\usepackage{hhline}
\usepackage{booktabs}
\usepackage{microtype}
\usepackage{graphicx}
\usepackage{amsthm}

\definecolor{green}{HTML}{009000} 
\definecolor{red}{HTML}{ea4335}  
\definecolor{orange}{HTML}{ff9900}
\definecolor{gray2}{HTML}{9e9e9e}
\definecolor{blue}{HTML}{0000ff}
\definecolor{red2}{HTML}{ff0000}

\newcommand{\hlg}[1]{\textcolor{green}{#1}}
\newcommand{\hlr}[1]{\textcolor{red}{#1}}
\newcommand{\better}[1]{\hlg{$\downarrow\,$#1}}
\newcommand{\worse}[1]{\hlr{$\uparrow\,$#1}}
\newcommand{\betterinv}[1]{\hlg{$\uparrow\,$#1}}
\newcommand{\worseinv}[1]{\hlr{$\downarrow\,$#1}}

\newcommand{\snapshot}{{\vtheta}^\text{past}}

\newcommand{\approach}{$\alpha$-SVRG}

\newcolumntype{x}[1]{>{\centering\arraybackslash}p{#1pt}}
\newcolumntype{y}[1]{>{\raggedright\arraybackslash}p{#1pt}}
\newcolumntype{z}[1]{>{\raggedleft\arraybackslash}p{#1pt}}

\definecolor{citecolor}{HTML}{0071BC}
\definecolor{linkcolor}{HTML}{ED1C24}
\usepackage{tikz}
\usepackage[pagebackref=false, breaklinks=true, letterpaper=true, colorlinks, citecolor=citecolor, linkcolor=linkcolor, bookmarks=false]{hyperref}

\author{
Yida Yin$^1$\quad Zhiqiu Xu$^2$\quad Zhiyuan Li$^3$\quad Trevor Darrell$^1$\quad Zhuang Liu$^4$ \vspace{0.3ex}\\
$^1$UC Berkeley \quad $^2$University of Pennsylvania \quad $^3$TTIC \quad $^4$Meta AI Research}

\title{A Coefficient Makes SVRG Effective}
\begin{document}

\maketitle
\begin{abstract}
   Stochastic Variance Reduced Gradient (SVRG), introduced by~\citet{johnson2013accelerating}, is a theoretically compelling optimization method. However, as~\cite{defazio2019ineffectiveness} highlight, its effectiveness in deep learning is yet to be proven. In this work, we demonstrate the potential of SVRG in optimizing real-world neural networks. Our empirical analysis finds that, for deeper neural networks, the strength of the variance reduction term in SVRG should be smaller and decrease as training progresses. Inspired by this, we introduce a multiplicative coefficient $\alpha$ to control the strength and adjust it through a linear decay schedule. We name our method \approach. Our results show \approach\ better optimizes models, consistently reducing training loss compared to the baseline and standard SVRG across various model architectures and multiple image classification datasets. We hope our findings encourage further exploration into variance reduction techniques in deep learning. Code is available at \url{github.com/davidyyd/alpha-SVRG}.
\end{abstract}

\section{Introduction}
A decade ago,~\citet{johnson2013accelerating} proposed a simple approach for reducing gradient variance in SGD---Stochastic Variance Reduced Gradient (SVRG). SVRG keeps a snapshot model and uses it to form a variance reduction term to adjust the gradient of the current model. This variance reduction term is the difference between the snapshot's stochastic gradient and its full gradient on the whole dataset. Utilizing this term, SVRG can reduce gradient variance of SGD and accelerate it to almost as fast as the full-batch gradient descent in strongly convex settings. 

Over the years, numerous SVRG variants have emerged. Some focus on further accelerating convergence in convex settings~\citep{xiao2014proximal, lin2015universal, defazio2016simple}, while others are tailored for non-convex scenarios~\citep{allenzhu2016variance, reddi2016stochastic, lei2017nonconvex, fang2018spider}. SVRG and its variants have shown effectiveness in optimizing simple machine learning models like logistic regression and ridge regression~\citep{allenzhu2018katyusha, lei2017nonconvex}.

Despite the theoretical value of SVRG and its subsequent works, they have seen limited practical success in training neural networks. Most SVRG research in non-convex settings is restricted to modest experiments: training basic models like Multi-Layer Perceptrons (MLP) or simple CNNs on small datasets like MNIST and CIFAR-10. These studies usually exclude evaluations on more capable and deeper networks. More recently,~\citet{defazio2019ineffectiveness} have exploited several variance reduction methods, including SVRG, to deep vision models. They found that SVRG fails to reduce gradient variance for deep neural networks because the model updates so quickly on the loss surface that the snapshot model becomes outdated and ineffective at variance reduction.

\begin{figure}[h]
\vspace{-1.5em}
    \centering
    \begin{tikzpicture}
\node[above right] (img) at (0,0) {\includegraphics[width=\linewidth, keepaspectratio]{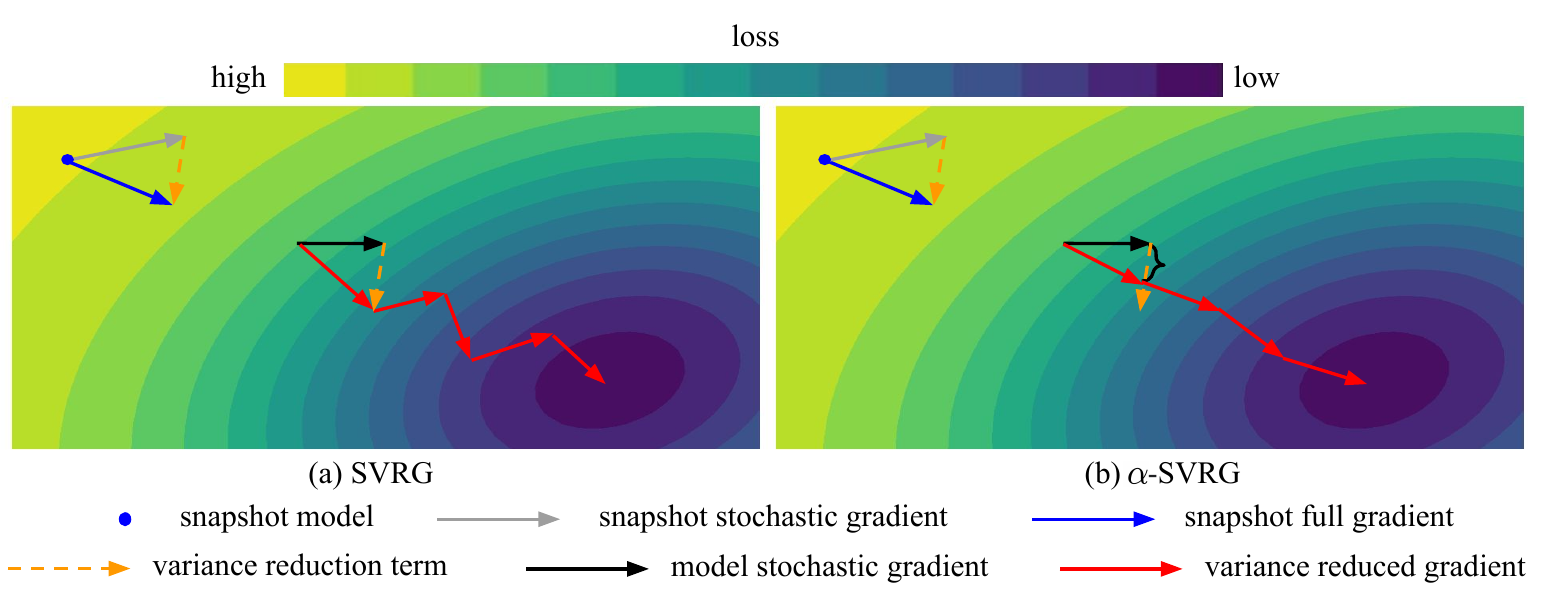}};
\node at (304pt,89pt) {$\alpha$};
\end{tikzpicture}
\vspace{-2em}
\caption{\textbf{SVRG vs. \(\boldsymbol{\alpha}\)-SVRG.}  Both SVRG (left) and \approach\ (right) use the difference between snapshot stochastic gradient (\textcolor{gray2}{gray}) and snapshot full gradient (\textcolor{blue}{blue}) to form a variance reduction term (\textcolor{orange}{orange}), which modifies model stochastic gradient (black) into variance reduced gradient (\textcolor{red2}{red}). But \approach\ employs a coefficient \(\alpha\) to modulate the strength of the variance reduction term. With this coefficient, \approach\ reduces the gradient variance and results in faster convergence.}
\label{fig:teaser}
\end{figure}
In this work, we show that adding a multiplicative coefficient to SVRG's variance reduction term can make it effective for deep neural networks. Our exploration is motivated by an intriguing observation: SVRG can only reduce gradient variance in the initial training stages but \emph{increases} it later. To tackle this problem, we mathematically derive the optimal coefficient for the variance reduction term to minimize the gradient variance. Our empirical analysis then leads to two key observations about this optimal coefficient: (1) as the depth of the model increases, the optimal coefficient becomes smaller; (2) as training advances, the optimal coefficient decreases, dropping well below the default coefficient of 1 in SVRG. These findings help explain why a constant coefficient of 1 in SVRG, while initially effective, eventually fails to reduce gradient variance.

Based on these observations, we introduce a linearly decaying coefficient $\alpha$ to control the strength of the variance reduction term in SVRG. We call our method \approach\ and illustrate it in Figure~\ref{fig:teaser}. \approach\ decreases gradient variance stably across both early and late training and helps optimize the model better.  We evaluate \approach\ on a range of architectures and image classification datasets. \approach\ achieves a lower training loss than the baseline and the standard SVRG. Our results highlight the value of SVRG in deep learning. We hope our work can offer insights about SVRG and stimulate more research in variance reduction approaches for optimization in neural networks.

\section{Motivation: SVRG may not always reduce variance}
\label{sec:motivation}
\textbf{SVRG formulation.} We first introduce the basic formulation of SVRG. We adopt the following notation: \(t\) is the iteration index, \(\vtheta^t\) represents the current model parameters, and \(\nabla f_i(\cdot)\) denotes the gradient of loss function $f$ for the $i$-th mini-batch. In SVRG's original work~\citep{johnson2013accelerating}, this corresponds to the $i$-th data point. When the subscript $i$ is omitted, \(\nabla f(\cdot)\) represents the full gradient across the entire dataset. A key concept in SVRG is the snapshot model, represented as \(\snapshot\). It is a snapshot of the model at a previous iteration before $t$. We store its full gradient \(\nabla f(\snapshot)\). This snapshot is taken periodically. SVRG defines the variance reduced gradient \(\vg_i^t\), as follows:
\begin{gather}
    \vg_i^t = \nabla f_i(\vtheta^t)-\underbrace{(\nabla f_i(\snapshot)-\nabla f(\snapshot))}_\text{variance reduction term}\label{eq:svrg}.
\end{gather}
Intuitively, SVRG uses the difference between the mini-batch gradient and full gradient of a past model to modify the current mini-batch gradient. This could make \(\vg_i^t\) better aligned with the current full gradient $\nabla f(\vtheta^t)$ and thus stabilize each update.

SVRG was initially introduced in the context of vanilla SGD. Recent work~\citep{adasvrg,adamsvrg} has integrated SVRG into alternative base optimizers. Following them, we input the variance reduced gradient \(\vg_i^t\) into the base optimizer and ensure a fair comparison by using the same base optimizer for SVRG and the baseline. We also follow the practice in~\cite{defazio2019ineffectiveness}, taking snapshot for SVRG once per training epoch.

\begin{table}[bp]
\centering
\small
\vspace{-.5em}
\begin{tabular}{c|cc}
name & formula & description\\
\Xhline{0.7pt}
metric 1* & $\frac{2}{N(N-1)}\sum_{i\neq j}\frac{1}{2}(1-\frac{\langle \vg_i^t,\vg_j^t\rangle}{\|\vg_i^t\|_2\|\vg_j^t\|_2})$& the directional variance of the gradients\\[3pt]
metric 2$\dagger$ & $\sum_{k=1}^d\Var(g_{i, k}^t)$& the variance of gradients across each component \\[4pt]
metric 3$\ddagger$ & $ \lambda_{max}(\frac{1}{N}\sum_{i=1}^N(\vg_i^t-\vg^t)(\vg_i^t-\vg^t)^T)$ & the magnitude of the most significant variation \\
\end{tabular}
\vspace{-.5em}
\caption{\textbf{Metrics.} $\vg^t$ is the mean of the mini-batch gradients $\vg_i^t$. $k$ indexes the $k$-th component of gradient $g_{i, k}^t$. References: $*$~\citet{liu2023dropout}, $\dagger$~\citet{defazio2019ineffectiveness}, $\ddagger$~\citet{jastrzebski2020breakeven}}
\label{tab:metric}
\end{table}
\textbf{Gradient variance.} Our goal is to assess SVRG's effectiveness in reducing gradient variance. To this end, we gather $N$ mini-batch gradients, denoted as \(\{\vg_i^t|i\in\{1,\cdots,N\}\}\), by performing back-propagation on checkpoints of the model at the iteration \(t\) with randomly selected $N$ mini-batches. For SVRG, each of these gradients is modified based on Equation~\ref{eq:svrg}. To present a comprehensive view, we employ three metrics from prior studies to quantify gradient variance in Table~\ref{tab:metric}.

In this part, we compare our three metrics. Metric 1 calculates the cosine distance between pairwise mini-batch gradients, therefore only capturing variance in gradient directions rather than gradient magnitudes. This is very important for scale-invariant optimizers, such as Adagrad~\citep{adagrad} and Adam~\citep{adam}. In contrast, metric 2 focuses on both gradient directions and magnitudes by summing the variance of each component of gradients. This metric has been the standard tool to measure gradient variance in various optimization literature~\citep{allenzhu2016variance, defazio2019ineffectiveness}. Metric 3 considers the largest eigenvalue in gradient covariance matrix, characterizing the most dominant part in gradient variance. We also average each gradient variance metric across three runs, with shaded regions in figures representing the standard deviation.

\textbf{SVRG's effect on gradient variance.} To understand how SVRG affects training, we examine two simple models: a linear layer (Logistic Regression) and a 4-layer Multi-Layer Perceptron (MLP-4). We train them over 30 epochs on CIFAR-10. We compare SVRG to a baseline using only SGD.

\begin{figure}[h]
\centering
\vspace{-.5em}
\begin{minipage}{0.48\textwidth}
    \centering
    \includegraphics[width=\linewidth,keepaspectratio]{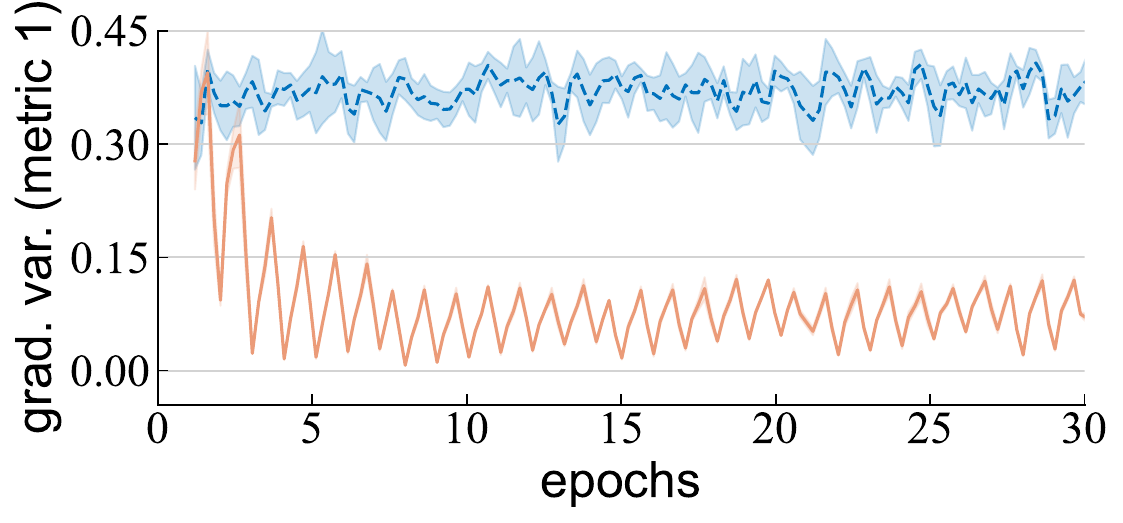}
\end{minipage}%
\begin{minipage}{0.48\textwidth}
    \centering
    \includegraphics[width=\linewidth,keepaspectratio]{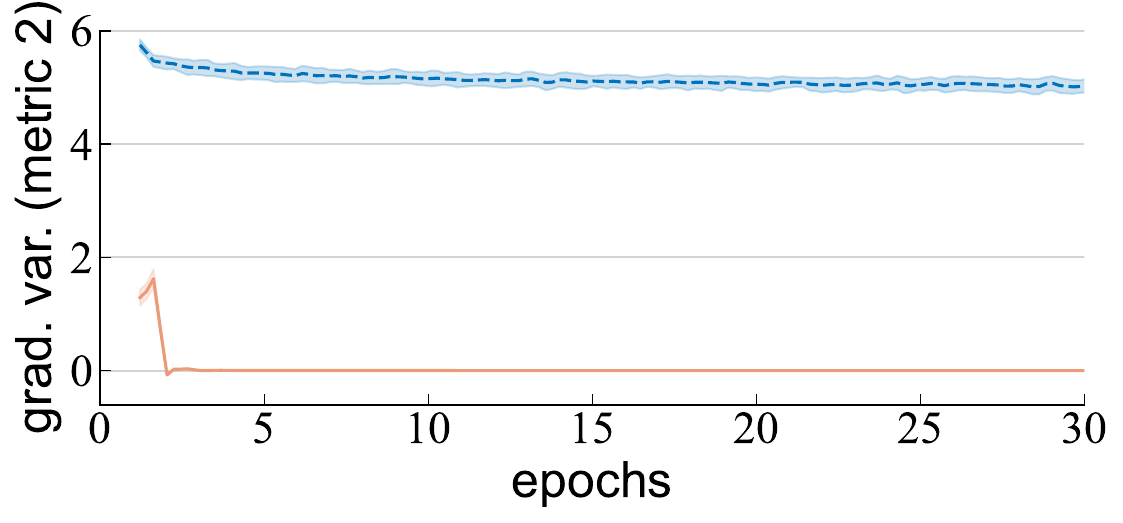}
\end{minipage}
\begin{minipage}{0.48\textwidth}
    \centering
    \includegraphics[width=\linewidth,keepaspectratio]{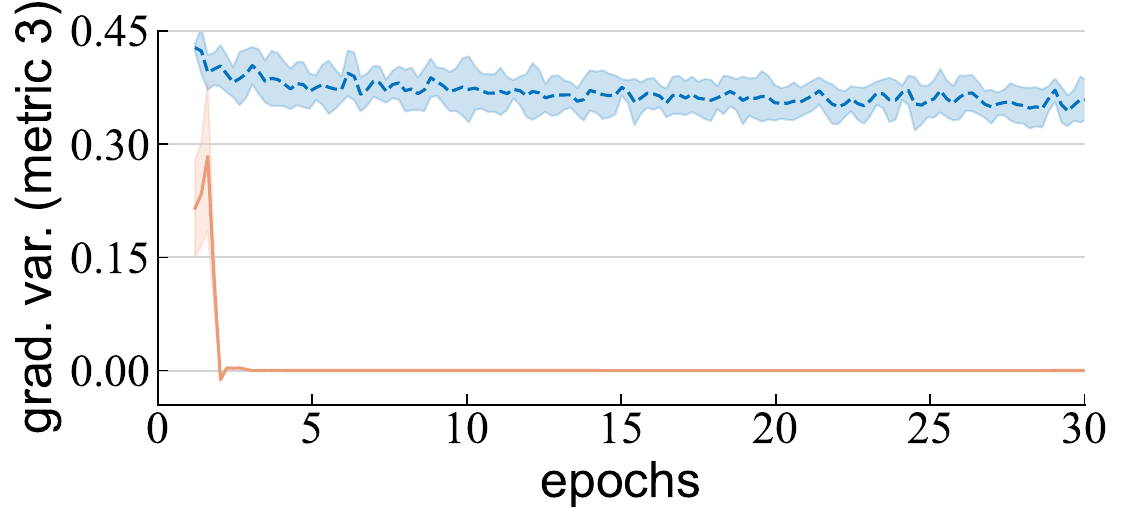}
\end{minipage}%
\begin{minipage}{0.48\textwidth}
    \centering
    \includegraphics[width=\linewidth,keepaspectratio]{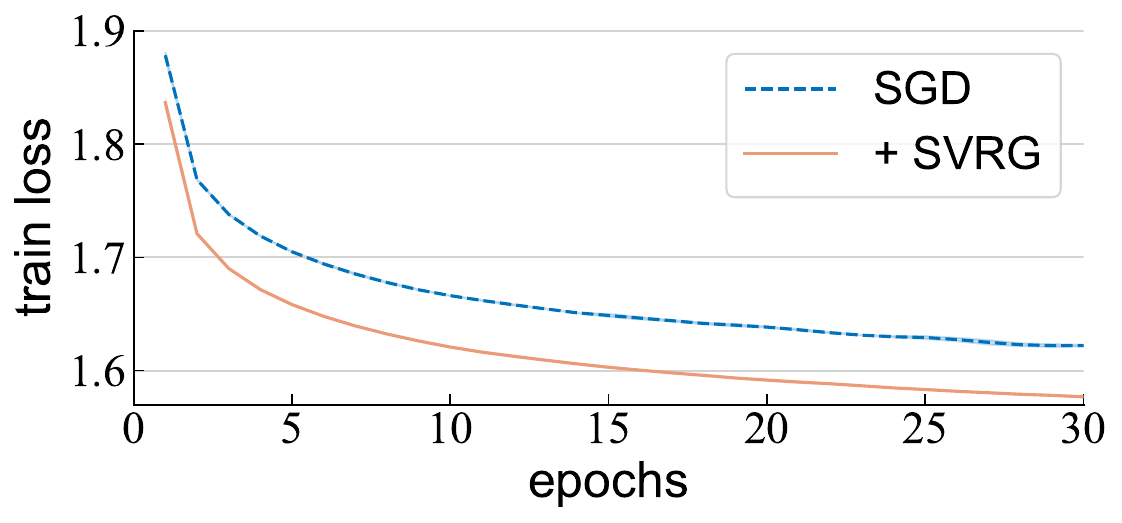}
\end{minipage}
\vspace{-.5em}
\caption{\textbf{SVRG on Logistic Regression.} SVRG effectively reduces the gradient variance for Logistic Regression, leading to a lower training loss than the baseline.}
\label{fig:baseline_vs_svrg_1}
\end{figure}

We plot Logistic Regression's gradient variance (top two and bottom left) and training loss (bottom right) in Figure~\ref{fig:baseline_vs_svrg_1}. For Logistic Regression, SVRG can reduce the gradient variance throughout the entire training process and achieve a lower training loss than the baseline.

\begin{figure}[!bh]
\centering
\vspace{-.5em}
\begin{minipage}{0.48\textwidth}
    \centering
    \includegraphics[width=\linewidth,keepaspectratio]{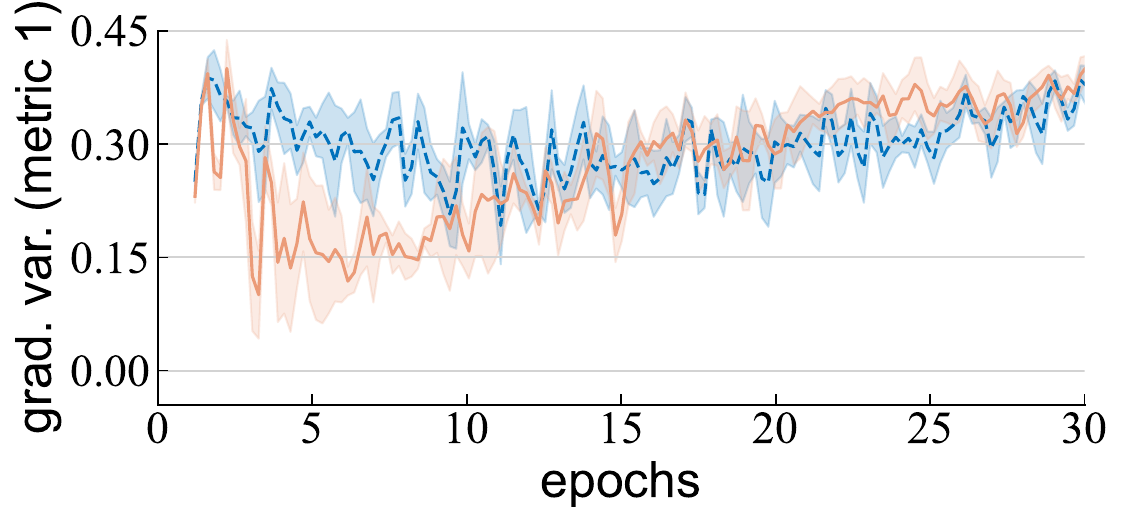}
\end{minipage}%
\begin{minipage}{0.48\textwidth}
    \centering
    \includegraphics[width=\linewidth,keepaspectratio]{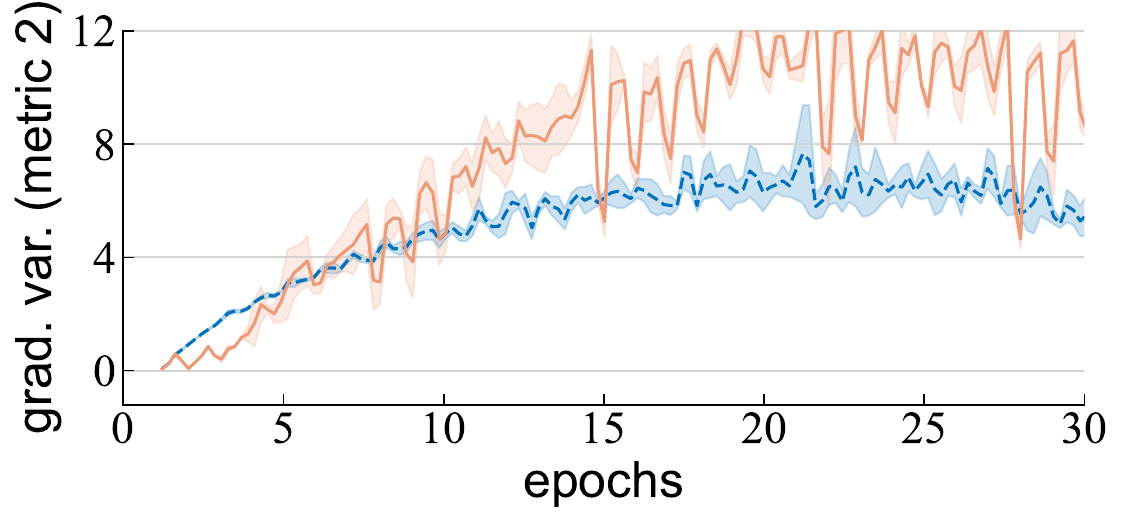}
\end{minipage}
\begin{minipage}{0.48\textwidth}
    \centering
    \includegraphics[width=\linewidth,keepaspectratio]{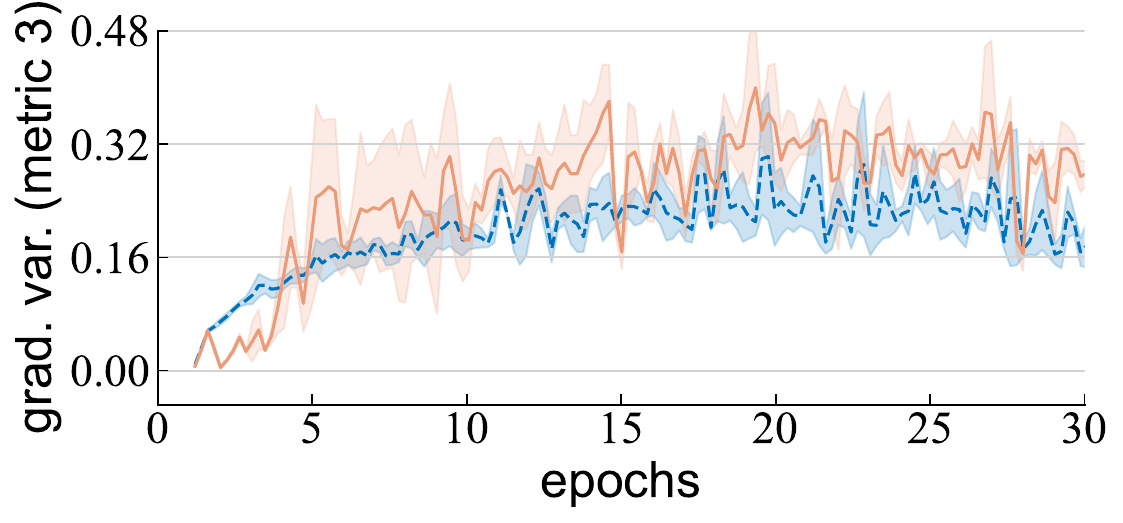}
\end{minipage}%
\begin{minipage}{0.48\textwidth}
    \centering
    \includegraphics[width=\linewidth,keepaspectratio]{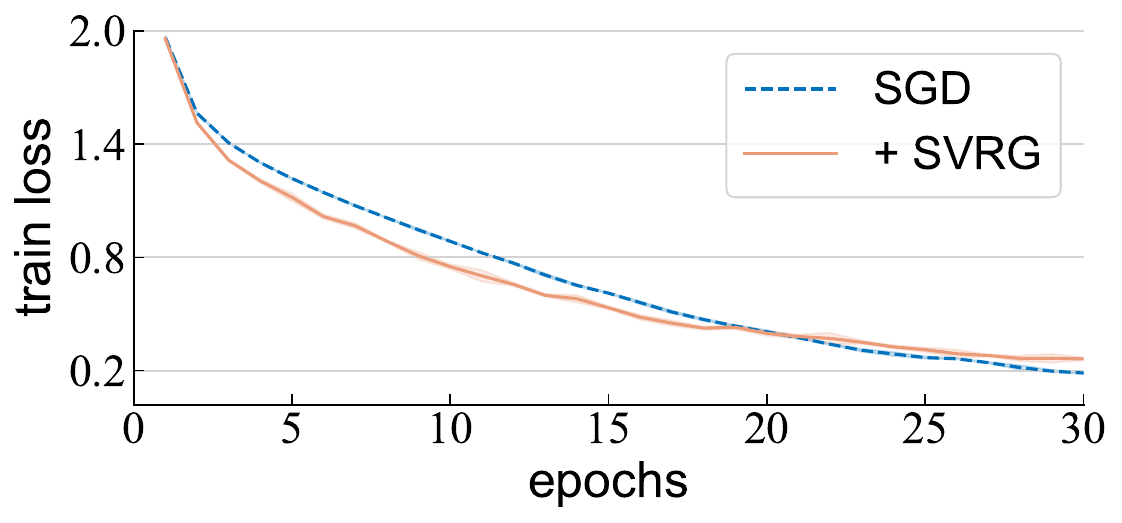}
\end{minipage}
\vspace{-.5em}
\caption{\textbf{SVRG on MLP-4.} In the first few epochs, SVRG reduces the gradient variance for MLP-4, but afterward, SVRG increases it, well above the baseline. As a result, SVRG exhibits a higher training loss than the baseline at the end of training.}
\label{fig:baseline_vs_svrg_4}
\vspace{-1em}
\end{figure}
In contrast, for MLP-4, SVRG may not always reduce gradient variance. As shown in Figure~\ref{fig:baseline_vs_svrg_4}, SVRG can only decrease the gradient variance for the first five epochs but then increases it. Consequently, SVRG has a larger final training loss than the baseline. This indicates that the increase in gradient variance caused by SVRG hinders the convergence of MLP-4's training loss.

This surprising empirical observation in a slightly deeper model leads us to question whether SVRG may alter the gradient too excessively at certain phases of training. Can we mitigate this adverse effect? We explore these questions starting from a theoretical framework in the next section.

\section{A Closer Look at Control Variates in SVRG}
\label{sec:3}
Control variates~\citep{lavenberg1977application} is a technique initially developed in Monte Carlo methods to reduce variance. We aim to estimate the expected value of a random variable \(\ermX\). The variance of this estimate usually depends on \(\Var(\ermX)\). To form a less variate estimate \(\ermX^*\), we can use a control variate \( \ermY \) that correlates with \(\ermX \) and a coefficient \( \alpha \) to regulate the influence of \( \ermY \) and \( \mathbb{E}[\ermY] \) :
\begin{gather}
    \ermX^* = \ermX - \alpha (\ermY - \mathbb{E}[\ermY]).\label{eq:control}
\end{gather}
This estimate remains unbiased for any value of \( \alpha \). The coefficient that minimizes the variance of the estimate can be derived as:
\begin{gather}
\alpha^* = \frac{\Cov(\ermX, \ermY)}{\Var(\ermY)}=\rho(\ermX, \ermY)\frac{\sigma(\ermX)}{\sigma(\ermY)},\label{eq:control_opt}
\end{gather}
where \(\rho(\ermX,\ermY)\) represents the correlation coefficient between \(\ermX\) and \(\ermY\); \(\sigma(\cdot)\) denotes the standard deviation. The derivation is detailed in Appendix~\ref{appendix:derivation}. 
The minimized variance becomes \(\Var(\ermX^*) = (1-\rho(\ermX,\ermY)^2)\Var(\ermX)\). The higher the correlation is, the lower the variance of the estimate is.

Note that SVRG uses control variates to reduce variance in each component of the gradient. This variance reduction occurs at each iteration \(t\). Take a closer look at Equation~\ref{eq:svrg} and~\ref{eq:control}: the model stochastic gradient \(f_i(\vtheta^t)\) is the random variable \(\ermX\); the snapshot stochastic gradient \(f_i(\snapshot)\) is the control variate \(\ermY\); and the snapshot full gradient \(f(\snapshot)\) is the expectation \( \mathbb{E}[\ermY] \). 

A key difference between SVRG and control variates is that SVRG omits the coefficient \(\alpha\), defaulting it to 1. This is possibly because the gradient distribution does not change drastically in strongly convex settings~\citep{johnson2013accelerating}. Yet, SVRG's subsequent studies, even those addressing non-convex cases, have neglected the coefficient and formulated their theories based on Equation~\ref{eq:svrg}. 

Motivated by this, we introduce a time-dependent coefficient vector \( \boldsymbol{\alpha}^t \in \mathbb{R}^d \) in SVRG:
\begin{gather}
    \vg_i^t = \nabla f_i(\vtheta^t)-\boldsymbol{\alpha}^t\odot(\nabla f_i(\snapshot)-\nabla f(\snapshot)),\label{eq:approach}
\end{gather}
where \( \odot \) represents the element-wise multiplication.

\textbf{Optimal coefficient.}
We adopt the same gradient variance definition as~\citet{defazio2019ineffectiveness} (metric 2 in Table~\ref{tab:metric}) and aim to determine the optimal $\boldsymbol{\alpha}^{t*}$ that minimizes it at each iteration. Specifically, our objective is to minimize the sum of variances across each component of \( \vg_i^t \). Let $k$ index the $k$-th component \( \alpha^{t*}_k \) and the $k$-th component of the gradient \(\nabla f_{\cdot, k}(\cdot)\). For clarity, we omit the mini-batch index $i$. This can be formally expressed as follows:
\begin{gather}
    \min_{\boldsymbol{\alpha}^t} \sum_{k=1}^d \Var(g^t_{\cdot, k})=\sum_{k=1}^d \min_{\alpha^t_k} \Var(g^t_{\cdot, k}).\label{eq:switch}
\end{gather}
We can switch the order of minimization and summation in Equation~\ref{eq:switch} because the variance of the $k$-th component of the gradient only depends on the $k$-th component of the coefficient. Applying Equation~\ref{eq:control_opt} yields the optimal coefficient \(\alpha^{t*}_k\):
\begin{gather}
    \alpha^{t*}_k = \frac{\Cov(\nabla f_{\cdot,k}(\snapshot), \nabla f_{\cdot,k}(\vtheta^t))}{\Var(\nabla f_{\cdot,k}(\snapshot))} = \rho(\nabla f_{\cdot, k}(\snapshot), \nabla f_{\cdot, k}(\vtheta^t))\frac{\sigma(\nabla f_{\cdot, k}(\vtheta^t))}{\sigma(\nabla f_{\cdot, k}(\snapshot))}.\label{eq:optimal}
\end{gather}
A stronger correlation between the snapshot and model gradients leads to a larger optimal coefficient. 

For small networks like MLP-4, calculating the optimal coefficient at each iteration is feasible by gathering all mini-batch gradients for both the current and snapshot models. For larger networks, however, this method becomes impractical; we will address this challenge later in the paper.

\begin{figure}[t]
    \centering
    \begin{minipage}{0.49\linewidth}
        \centering
        \includegraphics[width=\linewidth,keepaspectratio]{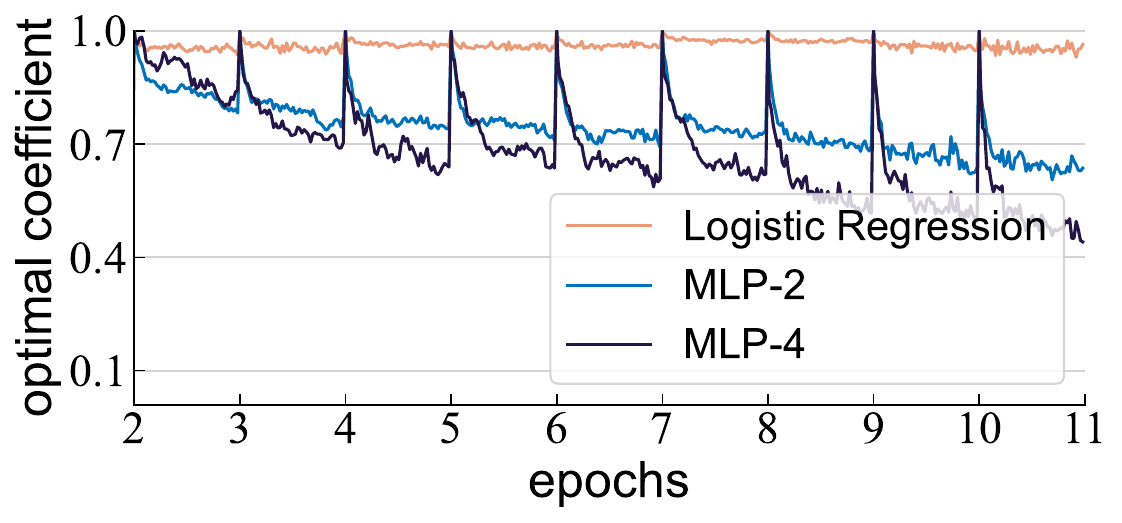}
        \vspace*{-0.7cm}
        \centering
        \caption*{\hspace{0.05\linewidth} (a) SGD}
    \end{minipage}~
    \begin{minipage}{0.49\linewidth}
        \centering
        \includegraphics[width=\linewidth,keepaspectratio]{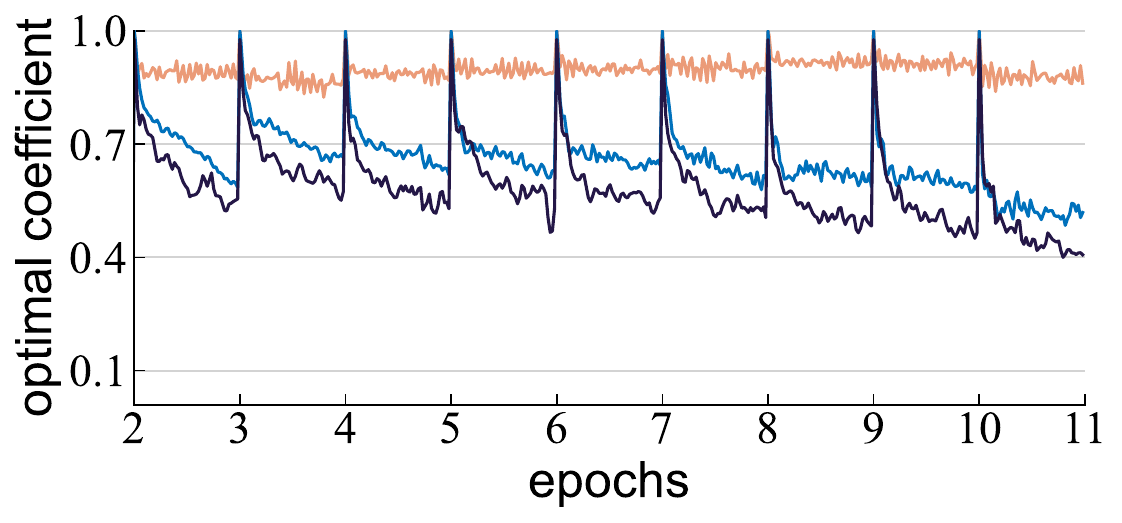}
        \vspace*{-0.7cm}
        \centering
        \caption*{\hspace{0.05\linewidth} (b) AdamW}
    \end{minipage}
    \caption{\textbf{Optimal coefficient.} At the start of each epoch, a snapshot is taken. Consequently, the optimal coefficient initiates at a value of 1 and results in a periodic upward jump.}
    \label{fig:depth}
\end{figure}

\textbf{Observations on optimal coefficient.}
To explore how the optimal coefficient evolves in a normal training setting, we train 1, 2, and 4-layer MLPs (Logistic Regression, MLP-2, and MLP-4) using SGD and AdamW~\citep{adamw} on CIFAR-10 \textit{without using SVRG}. Given the small size of these models, we can analytically compute the optimal coefficient at each iteration. We plot its mean value over all indices $k$ in Figure~\ref{fig:depth}. 
 We can make two notable observations as below.

\emph{Observation 1: a deeper model has a smaller optimal coefficient.} For Logistic Regression, the optimal coefficient remains relatively stable, hovering near 1. For MLP-2, the coefficient deviates from 1, dropping to about 0.6. For MLP-4, it decreases more sharply, reaching approximately 0.4.

\emph{Observation 2: the average optimal coefficient of a deeper model in each epoch generally decreases as training progresses.} This suggests that each epoch's average correlation between snapshot gradients and model gradients ($\rho(\nabla f_{\cdot, k}(\snapshot), \nabla f_{\cdot, k}(\vtheta^t))$ in Equation~\ref{eq:optimal}) decreases as the model becomes better trained. We further analyze this decreasing trend of the correlation term in Appendix~\ref{appendix:corr}.

These observations shed light on why the standard SVRG struggles to reduce gradient variance or training loss in later training stages (Figure~\ref{fig:baseline_vs_svrg_4}). A default coefficient of 1 proves to be too high, and the weakening correlation between snapshot and model gradients necessitates a smaller coefficient. Without a suitable coefficient, gradient variance may increase, leading to oscillations in SGD.

\begin{figure}[th]
\centering
\begin{minipage}{0.48\textwidth}
    \centering
    \includegraphics[width=\linewidth,keepaspectratio]{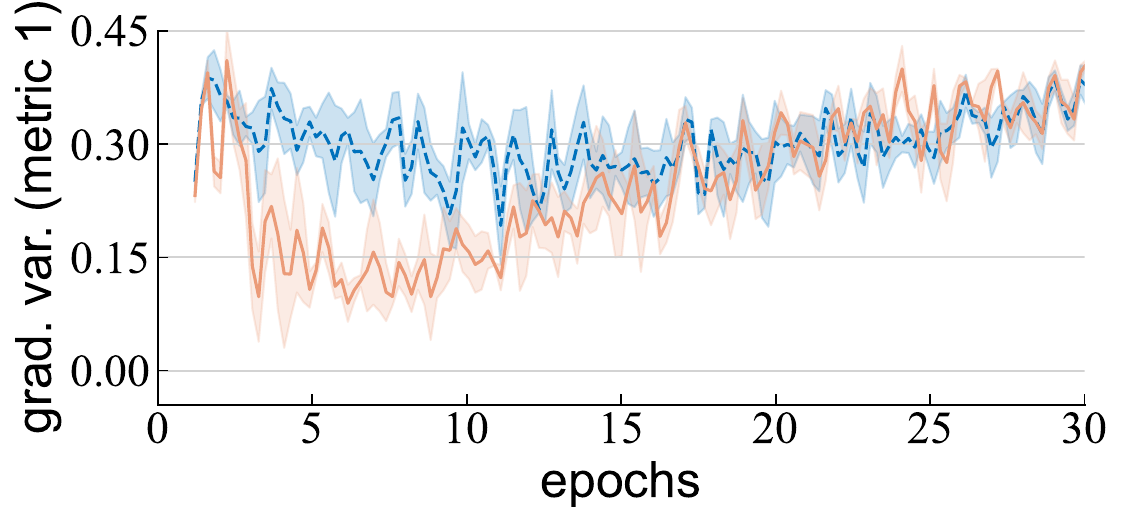}
\end{minipage}%
\begin{minipage}{0.48\textwidth}
    \centering
    \includegraphics[width=\linewidth,keepaspectratio]{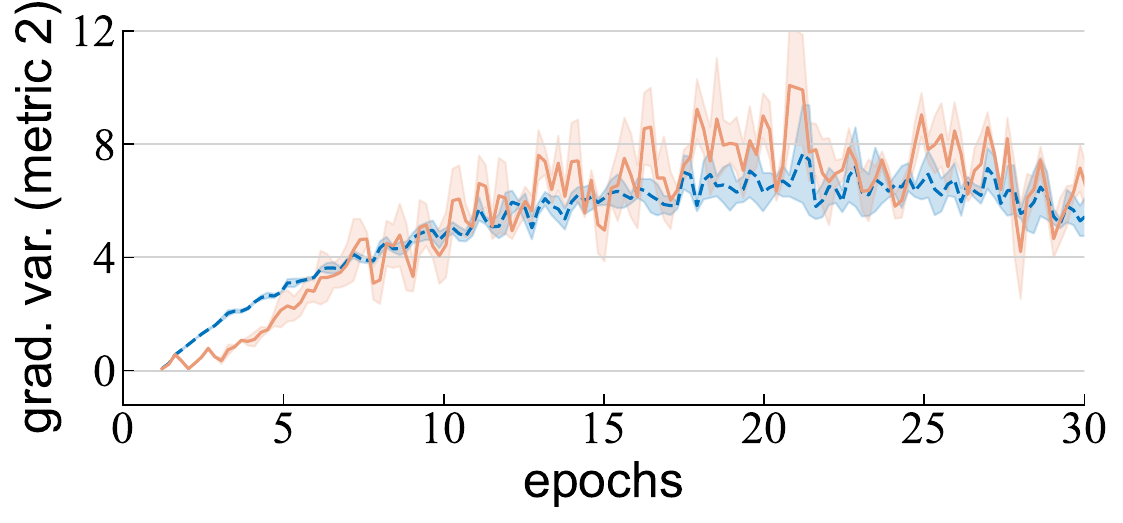}
\end{minipage}
\centering
\begin{minipage}{0.48\textwidth}
    \centering
    \includegraphics[width=\linewidth,keepaspectratio]{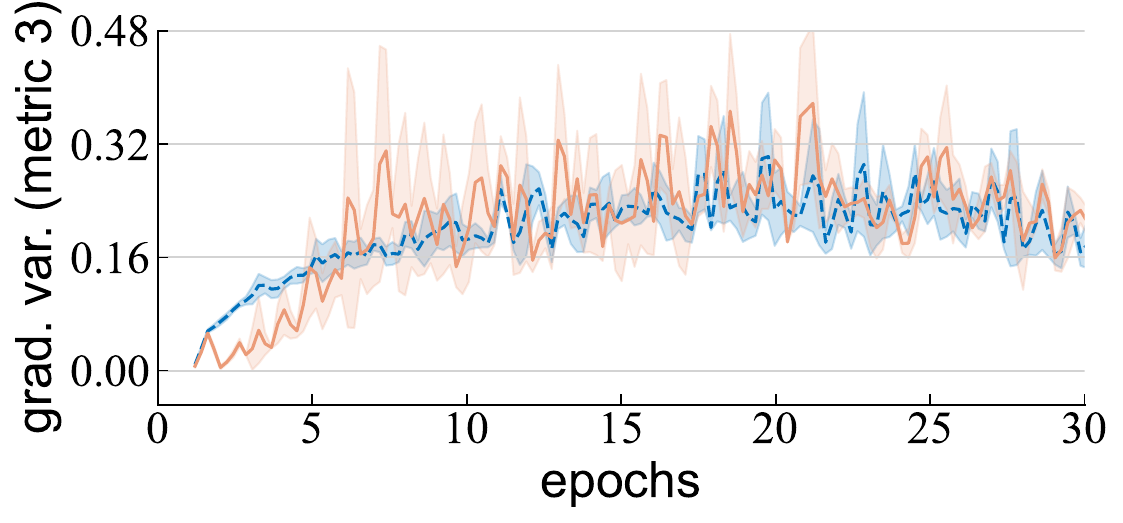}
\end{minipage}%
\begin{minipage}{0.48\textwidth}
    \centering
    \includegraphics[width=\linewidth,keepaspectratio]{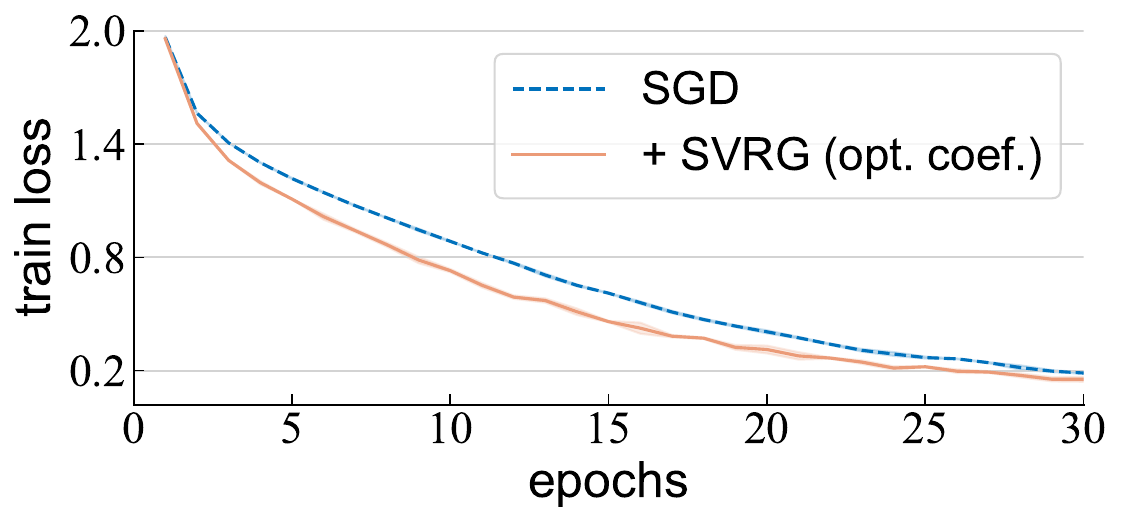}
\end{minipage}
\caption{\textbf{SVRG with optimal coefficient on MLP-4.} SVRG with the optimal coefficient reduces gradient variance stably and achieves a lower training loss than the baseline SGD.}
\label{fig:svrg_vs_optimal_svrg}
\end{figure}

\textbf{Optimal coefficient's effect on gradient variance.} We evaluate whether optimal coefficient can make SVRG more effective in reducing gradient variance. Specifically, we use SVRG with optimal coefficient to train an MLP-4 by computing optimal coefficient (Equation~\ref{eq:optimal}) and adjusting the gradient (Equation~\ref{eq:approach}) at each iteration. In Figure~\ref{fig:svrg_vs_optimal_svrg}, we compare SVRG with optimal coefficient to an SGD baseline. Using the optimal coefficient enables SVRG to reduce gradient variance in the early stages of training without uplifting it later. This yields a consistently lower training loss than the baseline.

\section{\approach}
\label{sec:approach}
From our analysis above, it becomes clear that the best coefficient for SVRG is not necessarily 1 for deep neural networks. However, computing the optimal coefficient at each iteration would result in a complexity of full-batch gradient descent. This approach quickly becomes impractical for larger networks like ResNet~\citep{He2016}. In this section, we show how using a preset schedule of $\alpha$ values can achieve a similar effect of using the computed optimal coefficients.

\textbf{\(\boldsymbol{\alpha}\)-SVRG.}  Given the decreasing trend (Figure~\ref{fig:depth}) and the computational challenge, we propose to apply a linearly decreasing scalar coefficient for SVRG, starting from an initial value $\alpha_0$ and decreasing to 0. This is our main method in this paper. We name it \approach. More results of enabling {\approach} only during the early stage of training are in Appendix~\ref{appendix:early_svrg}. The pseudocode for \approach\ with SGD and AdamW as base optimizers is provided in Appendix~\ref{appendix:pseudocode}.

\begin{figure}[h]
\centering
\begin{minipage}{0.48\textwidth}
    \centering
    \includegraphics[width=\linewidth,keepaspectratio]{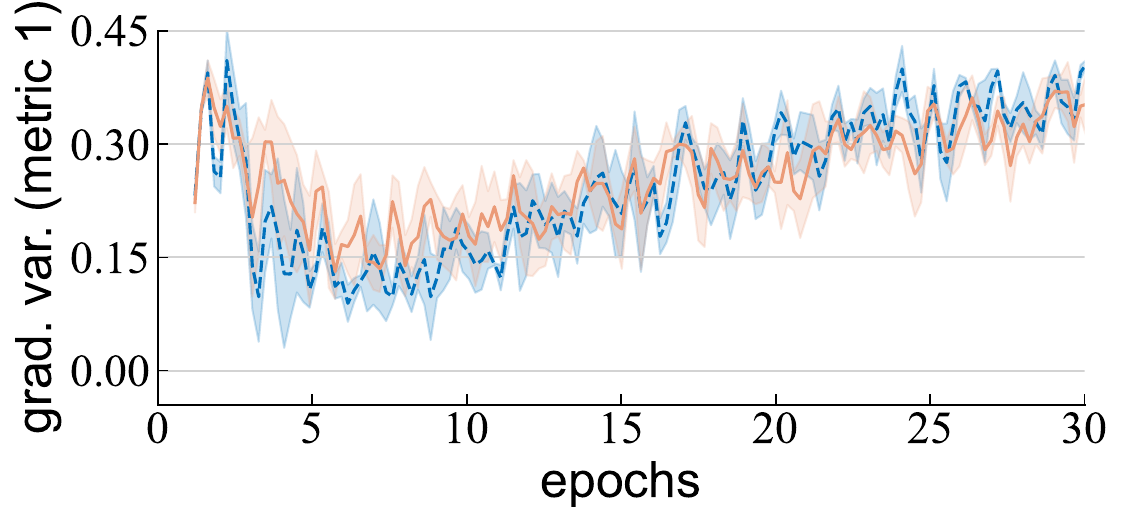}
\end{minipage}%
\begin{minipage}{0.48\textwidth}
    \centering
    \includegraphics[width=\linewidth,keepaspectratio]{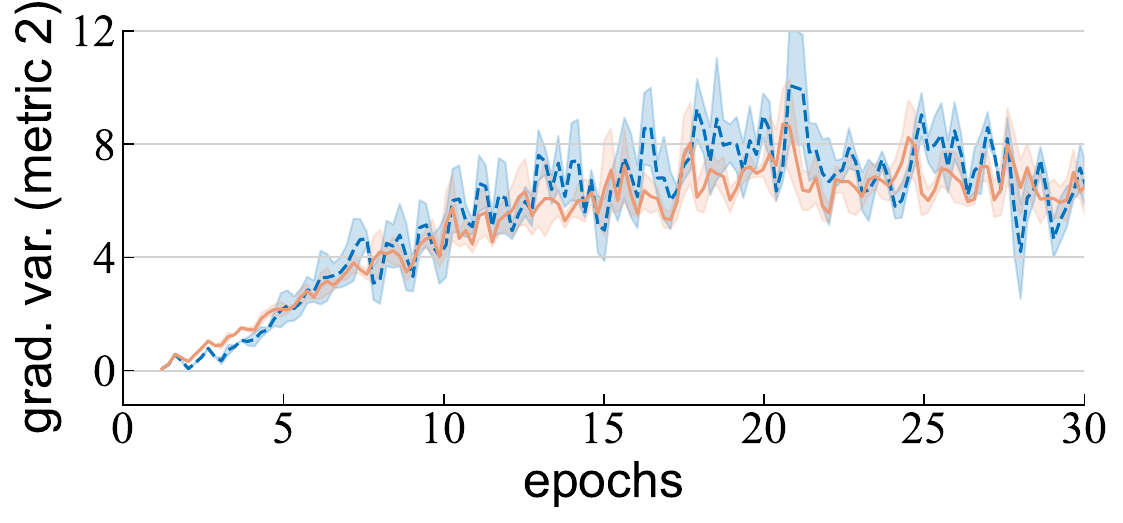}
\end{minipage}
\begin{minipage}{0.48\textwidth}
    \centering
    \includegraphics[width=\linewidth,keepaspectratio]{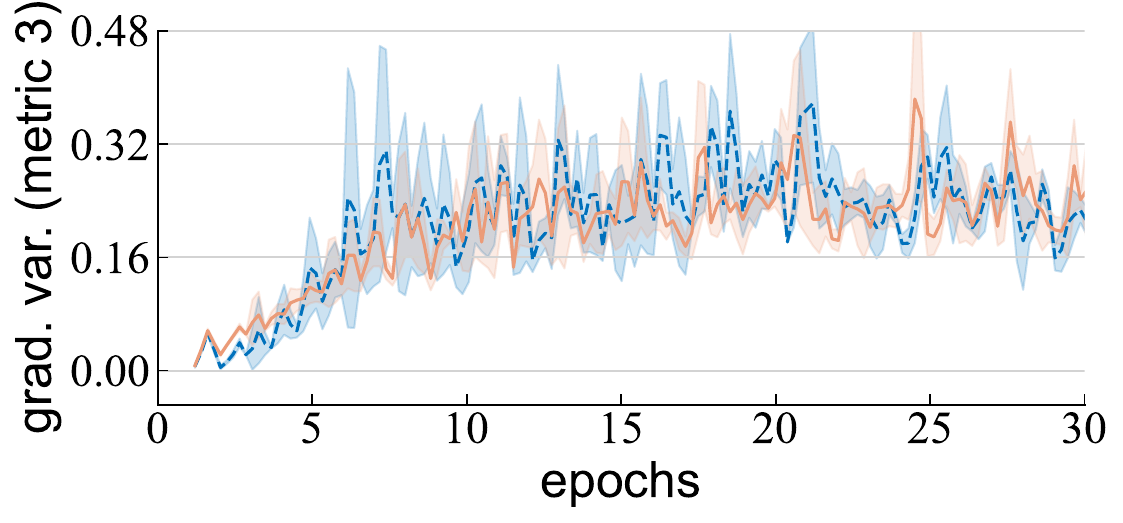}
\end{minipage}%
\begin{minipage}{0.48\textwidth}
    \centering
    \includegraphics[width=\linewidth,keepaspectratio]{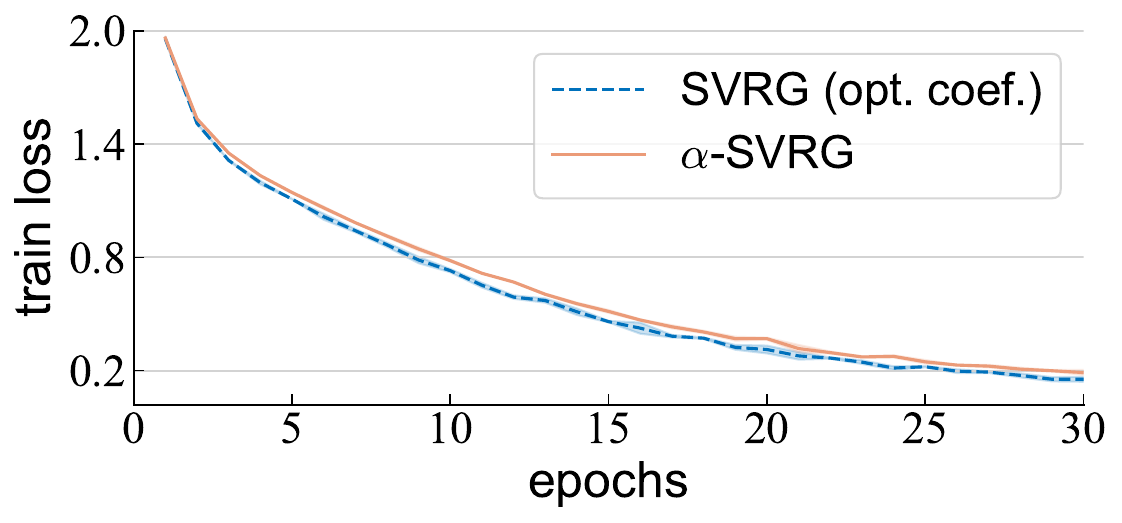}
\end{minipage}
\caption{\textbf{$\boldsymbol{\alpha}$-SVRG on MLP-4.} \approach\ behaves similarly to SVRG with optimal coefficient.}
\label{fig:optimal_svrg_vs_0.5svrg}
\end{figure}
To evaluate how well {\approach} matches SVRG with optimal coefficient, we train an MLP-4 using {\approach} and compare it to SVRG with optimal coefficient. For all experiments in this section, we set \(\alpha_0=0.5\). The results are presented in Figure~\ref{fig:optimal_svrg_vs_0.5svrg}. Interestingly, {\approach} exhibits a gradient variance trend that is not much different from SVRG with optimal coefficient. Similarly, the training loss of {\approach} is only marginally larger than that of SVRG with optimal coefficient.  

\begin{figure}[bh]
\centering
\begin{minipage}{0.48\textwidth}
    \centering
    \includegraphics[width=\linewidth,keepaspectratio]{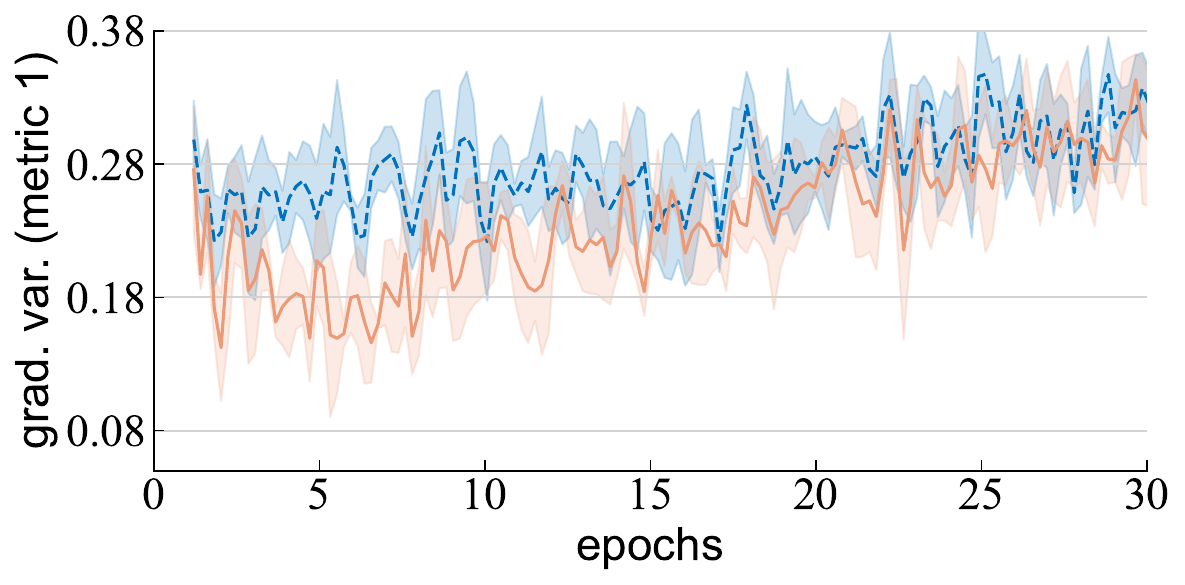}
\end{minipage}%
\begin{minipage}{0.48\textwidth}
    \centering
    \includegraphics[width=\linewidth,keepaspectratio]{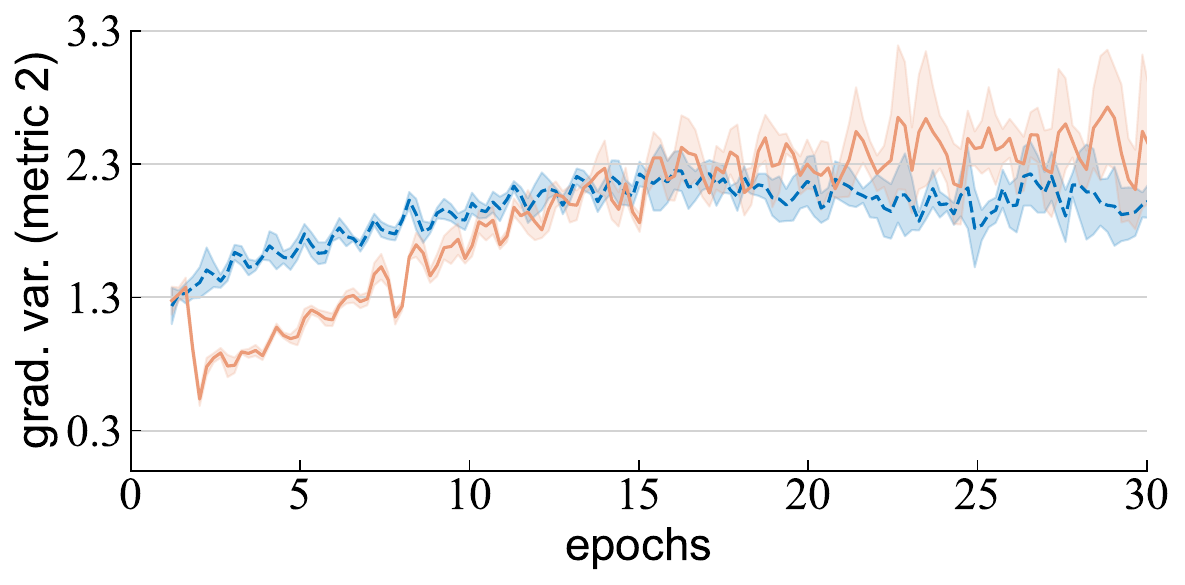}
\end{minipage}
\begin{minipage}{0.48\textwidth}
    \centering
    \includegraphics[width=\linewidth,keepaspectratio]{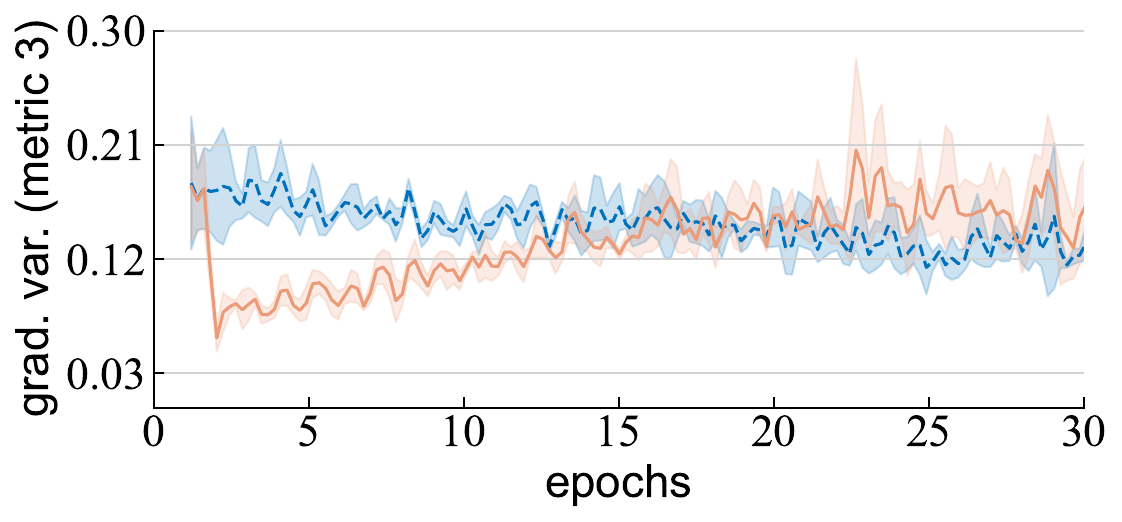}
\end{minipage}%
\begin{minipage}{0.48\textwidth}
    \centering
    \includegraphics[width=\linewidth,keepaspectratio]{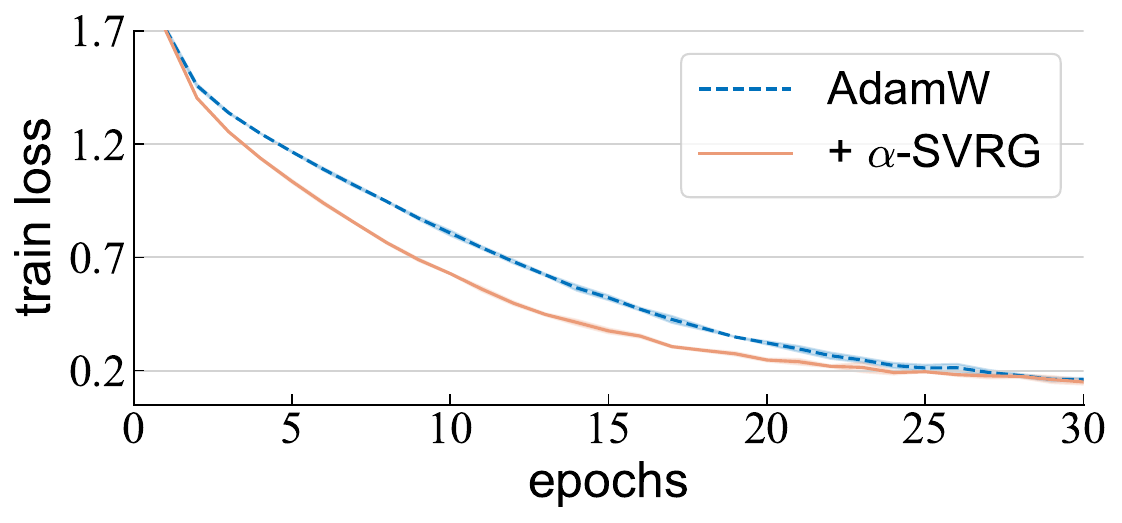}
\end{minipage}
\caption{\textbf{$\boldsymbol{\alpha}$-SVRG with AdamW on MLP-4}. \approach\ can lower the gradient variance at the first 10 epochs, leading to a faster convergence than the baseline AdamW.}
\label{fig:svrg_vs_0.5_svrg_adam}
\end{figure} 

\textbf{\(\boldsymbol{\alpha}\)-SVRG with AdamW.} Since AdamW~\citep{adamw} is a widely used optimizer in modern neural network training, we assess the performance of {\approach} with AdamW. We change the base optimizer in {\approach} to AdamW and use it to train an MLP-4 on CIFAR-10. We compare {\approach} to a baseline using only AdamW. As shown in Figure~\ref{fig:svrg_vs_0.5_svrg_adam}, {\approach} has a noticeable gradient variance reduction initially and achieves a consistent lower training loss for MLP-4 than the baseline.

\begin{figure}[h]
\centering
\begin{minipage}{0.48\textwidth}
    \centering
    \includegraphics[width=\linewidth,keepaspectratio]{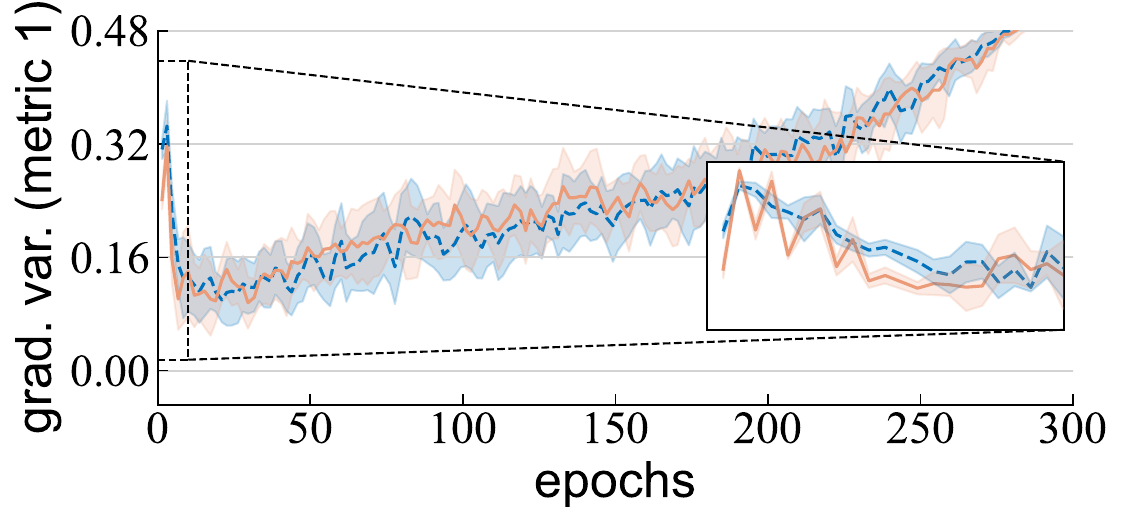}
\end{minipage}%
\begin{minipage}{0.48\textwidth}
    \centering
    \includegraphics[width=\linewidth,keepaspectratio]{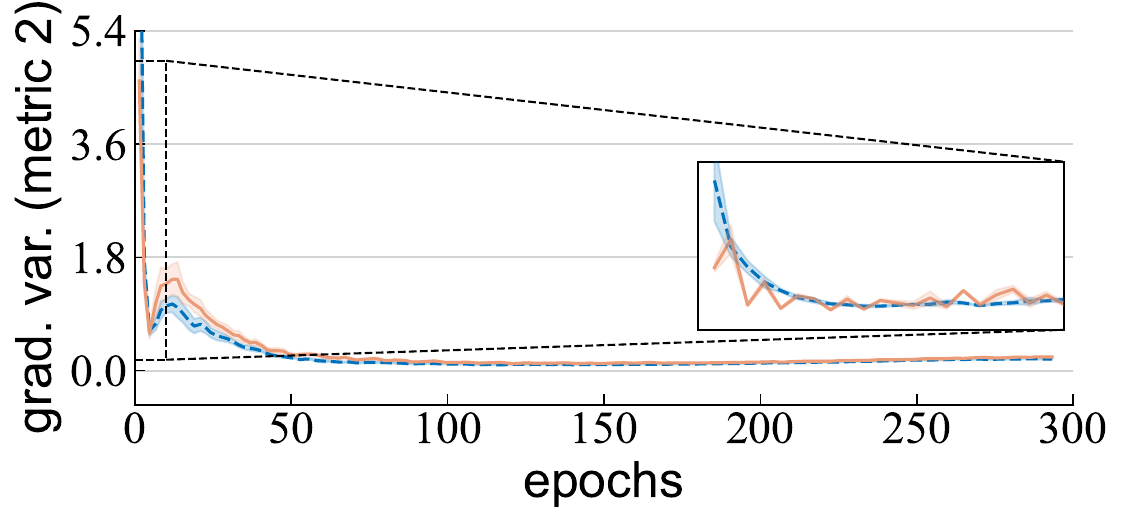}
\end{minipage}
\begin{minipage}{0.48\textwidth}
    \centering
    \includegraphics[width=\linewidth,keepaspectratio]{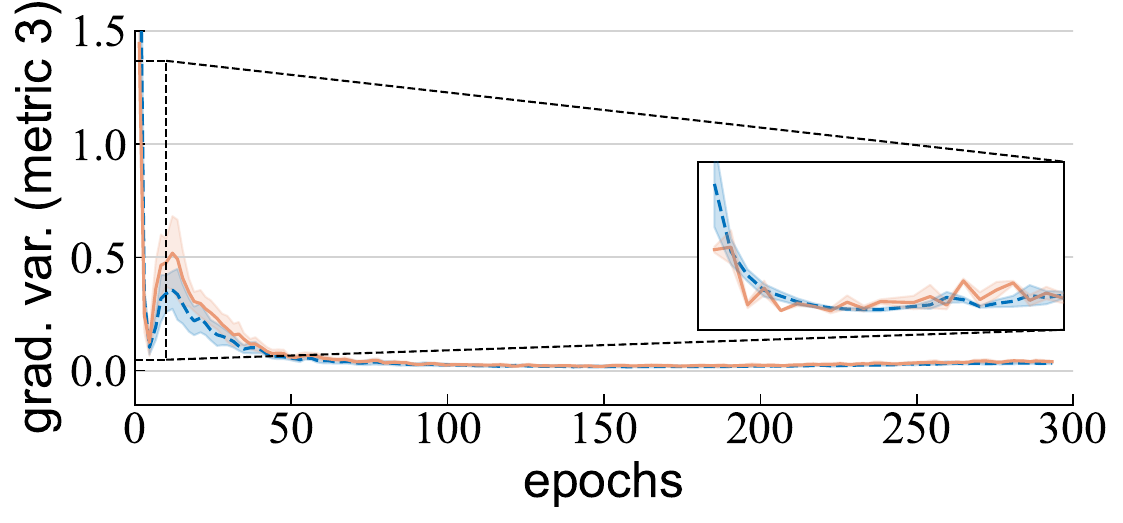}
\end{minipage}%
\begin{minipage}{0.48\textwidth}
    \centering
    \includegraphics[width=\linewidth,keepaspectratio]{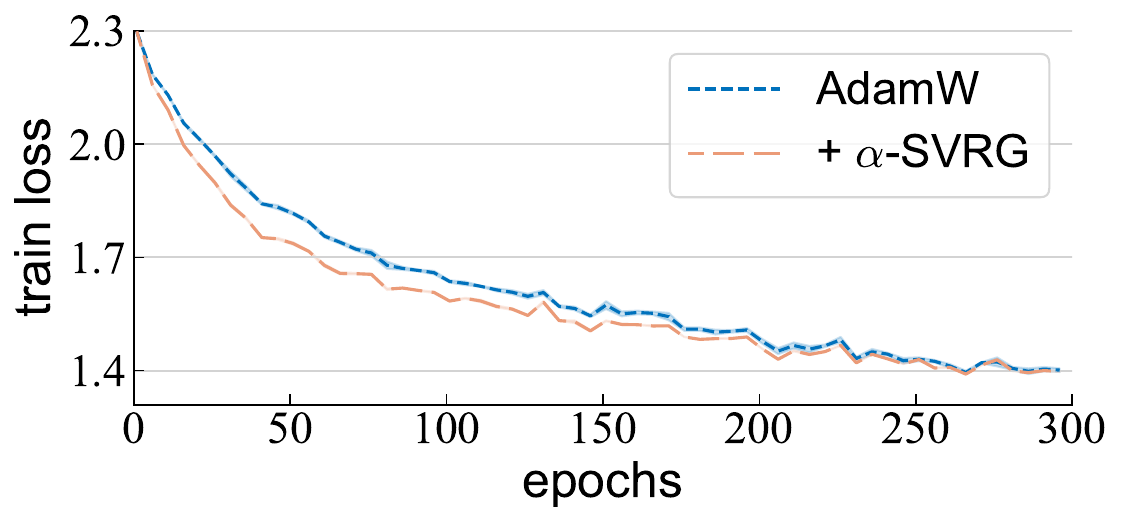}
\end{minipage}
\caption{\textbf{$\boldsymbol{\alpha}$-SVRG on ConvNeXt-Femto}. \approach\ can reduce the gradient variance for ConvNeXt-Femto during the first 10 epochs (zoom-in plot) without increasing it later on. Consequently, $\alpha$-SVRG can decrease the training loss at a faster rate than the baseline AdamW.}
\label{fig:convnext}
\end{figure}

\textbf{\(\boldsymbol{\alpha}\)-SVRG on deeper networks.}
We further study the effectiveness of {\approach} with AdamW on real-world neural architectures, moving beyond simple MLPs. To this end, we train a modern ConvNet architecture, ConvNeXt-Femto~\citep{convnextzhuang,convnext-f}, on CIFAR-10 using the default AdamW optimizer. We compare {\approach} to the baseline using vanilla AdamW in Figure~\ref{fig:convnext}. {\approach} can reduce gradient variance during the first 10 epochs (zoom-in plot of Figure~\ref{fig:convnext}) and then maintain it at the same level as the baseline. As a result, the training loss of {\approach} converges much faster than the baseline. This demonstrates the potential of {\approach} in optimizing more complex models. We further explore this with additional experiments next.

\section{Experiments}
\label{sec:experi}
\subsection{Settings}
\textbf{Datasets.} We evaluate \approach\ using ImageNet-1K classification~\citep{ImageNet1k} as well as smaller image classification datasets: CIFAR-100~\citep{cifar}, Pets~\citep{pets}, Flowers~\citep{flowers}, STL-10~\citep{stl10}, Food-101~\citep{food101}, DTD~\citep{dtd}, SVHN~\citep{svhn}, and EuroSAT~\citep{eurosat1}.

\textbf{Models.}
We use recently proposed vision models on ImageNet-1K, categorized into two groups: (1) smaller models with 5-19M parameters, including ConvNeXt-F~\citep{convnext-f, convnextzhuang}, ViT-T/16~\citep{vit}, Swin-F~\citep{swin}, and Mixer-S/32~\citep{mlp-mixer}; (2) larger models featuring 86M and 89M parameters: ViT-B/16 and ConvNeXt-B. ConvNeXt-F is also evaluated on all smaller image classification datasets.

\textbf{Training.}
We report both final epoch training loss and top-1 validation accuracy. Our basic training setting follows ConvNeXt~\citep{convnextzhuang}, which uses AdamW. Both SVRG and $\alpha$-SVRG also use AdamW as the base optimizer.  On small datasets, we choose the best $\alpha_0$ from \{0.5, 0.75, 1\}. We find the coefficient is robust and does not require extensive tuning. Therefore, for ImageNet-1K, we set $\alpha_0$ to 0.75 for smaller models and 0.5 for larger ones. Other training settings for \approach\ remain the same as the baseline. Further experimental settings can be found in Appendix \ref{appendix:setting}. 

\begin{table}[t]
\def\arraystretch{1.3}
\addtolength{\tabcolsep}{-2pt}
\centering
\small
\begin{tabular}{lcccccccccccc}
& \multicolumn{2}{c}{ConvNeXt-F} & \multicolumn{2}{c}{ViT-T} & \multicolumn{2}{c}{Swin-F} & \multicolumn{2}{c}{Mixer-S} & \multicolumn{2}{c}{ViT-B} & \multicolumn{2}{c}{ConvNeXt-B} \\
\Xhline{0.7pt}
\multicolumn{13}{c}{training loss} \\
AdamW & 3.487 & - & 3.443 & - & 3.427 & - &  3.149 & - & 2.817 & - & 2.644 & - \\
+\ SVRG & 3.505 & \worse{.018} & 3.431 & \better{.012} & \textbf{3.389} & \better{.038} & 3.172 & \worse{.023} & 3.309 & \worse{.492} & 3.113 & \worse{.469} \\
\rowcolor{gray}
+\ \approach & \textbf{3.467} & \better{.020} & \textbf{3.415} & \better{.028} & 3.392 & \better{.035} & \textbf{3.097} & \better{.052} & \textbf{2.806} & \better{.011} & \textbf{2.642} & \better{.002} \\
\Xhline{0.7pt}
\multicolumn{13}{c}{validation accuracy} \\
AdamW & 76.0 & - & 73.9 & - & 74.3 & - & \textbf{76.4} & - & \textbf{81.6} & - & \textbf{83.7} & - \\
+\ SVRG & 75.7 & \worseinv{0.3} & \textbf{74.3} & \betterinv{0.4} & 74.3 & \betterinv{0.0} & 74.5 & \worseinv{1.9} & 78.0 & \worseinv{3.6} & 80.8 & \worseinv{2.9} \\
\rowcolor{gray}
+\ \approach & \textbf{76.3} & \betterinv{0.3} & 74.2 & \betterinv{0.3} & \textbf{74.8} & \betterinv{0.5} & 76.1 & \worseinv{0.3} & \textbf{81.6} & \betterinv{0.0} & 83.1 & \worseinv{0.6} \\
\end{tabular}
\caption{\textbf{Results on ImageNet-1K.} The standard SVRG increases the training loss for most models, whereas \approach\ consistently decreases it for all models.}
\label{tab:pre-trained1}
\end{table}

\begin{table}[ht]
\centering
\small
\addtolength{\tabcolsep}{-4.35pt}
\def\arraystretch{1.3}
\begin{tabular}{lcccccccccccccccc}
&\multicolumn{2}{c}{CIFAR-100} & \multicolumn{2}{c}{Pets} & \multicolumn{2}{c}{Flowers} & \multicolumn{2}{c}{STL-10} & \multicolumn{2}{c}{Food-101} & \multicolumn{2}{c}{DTD} & \multicolumn{2}{c}{SVHN} & \multicolumn{2}{c}{EuroSAT} \\
\Xhline{0.7pt}

\multicolumn{17}{c}{training loss} \\
AdamW & 2.66 & - & 2.20 & - & 2.40 & - & 1.64 & - & 2.45 & - & 1.98 & - & 1.59 & - & 1.25 & - \\
+\ SVRG & 2.94 & \worse{0.28} & 3.42 & \worse{1.22} & 2.26 & \better{0.14} & 1.90 & \worse{0.26} & 3.03 & \worse{0.58} & 2.01 & \worse{0.03} & 1.64 & \worse{0.05} & 1.25 & 0.00 \\
\rowcolor{gray}
+\ \approach & \textbf{2.62} & \better{0.04} & \textbf{1.96} & \better{0.24} & \textbf{2.16} & \better{0.24} & \textbf{1.57} & \better{0.07} & \textbf{2.42} & \better{0.03} & \textbf{1.83} & \better{0.15} & \textbf{1.57} & \better{0.02} & \textbf{1.23} & \better{0.02} \\

\Xhline{0.7pt}
\multicolumn{17}{c}{validation accuracy} \\
AdamW & 81.0 & - & 72.8 & - & 80.8 & - & 82.3 & - & \textbf{85.9} & - & 57.9 & - & 94.9 & - & 98.1 & - \\
+\ SVRG & 78.2 & \worseinv{2.8} & 17.6 & \worseinv{55.2} &  82.6 & \betterinv{1.8} & 65.1 & \worseinv{17.2} & 79.6 & \worseinv{6.3} & 57.8 & \worseinv{0.1} & 95.7 & \betterinv{0.8} & 97.9 & \worseinv{0.2} \\
\rowcolor{gray}
+\ \approach & \textbf{81.4} & \betterinv{0.4} & \textbf{77.8} & \betterinv{5.0} & \textbf{83.3} & \betterinv{2.5} & \textbf{84.0} & \betterinv{1.7} & \textbf{85.9} & \betterinv{0.0} & \textbf{61.8} & \betterinv{3.9} & \textbf{95.8} & \betterinv{0.9} & \textbf{98.2} & \betterinv{0.1} \\
\end{tabular}
\caption{\textbf{Results on smaller classification datasets.} While the standard SVRG mostly hurts the performance, \approach\ decreases the training loss and increases the validation accuracy.}
\label{tab:pre-trained2}
\end{table}
\subsection{Results}
Table~\ref{tab:pre-trained1} presents the results of training various models on ImageNet-1K. The standard SVRG often increases the training loss, especially for larger models. In contrast, \approach\ decreases the training loss for both smaller and larger models. This also supports our earlier finding that deeper models benefit from lower coefficient values, and using a default coefficient of 1 impedes convergence.

Table~\ref{tab:pre-trained2} displays the results of training ConvNeXt-F on various smaller datasets. The standard SVRG generally elevates the training loss and impairs the generalization. On the contrary, \approach\ lowers the training loss and improves the validation accuracy across all small datasets. We have provided additional experiment results to demonstrate $\alpha$-SVRG's effectiveness in Appendix~\ref{appendix:additional_experiments}. 

Note that a lower training loss in \approach\ does not always lead to better generalization. For smaller models, a lower training loss usually directly translates to a higher validation accuracy. In larger models (Mixer-S, ViT-B, and ConvNeXt-B), additional adjustments to regularization strength may be needed for better generalization. This is out of scope for \approach\ as an optimization method but warrants future research on co-adapting optimization and regularization. 

\subsection{Analysis}
\label{sec:analysis}
We analyze various components in \approach. In the following experiments, we use an initial value $\alpha_0=0.5$ and ConvNeXt-F on STL-10 as the default setting. Because the standard SVRG is ineffective as discussed above, we omit it and only compare \approach\ to an AdamW baseline.

\begin{figure}[t]
\centering
\begin{minipage}[b]{0.48\textwidth}
    \includegraphics[width=\textwidth,keepaspectratio]{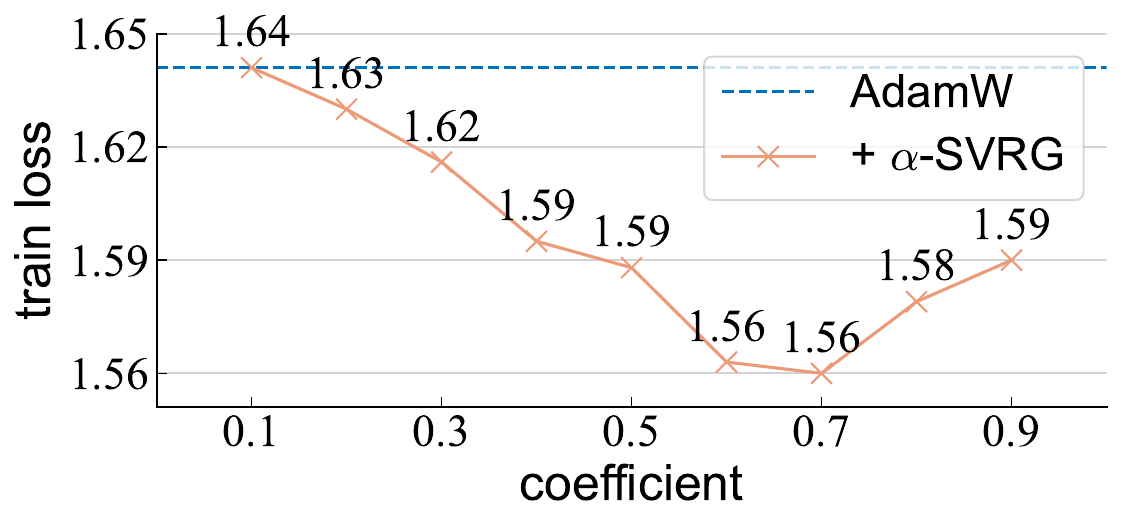}
    \vspace*{-2em}
    \caption{\textbf{Coefficient value.} \approach\ is effective with a wide range of coefficient values.}
    \label{fig:coef}
\end{minipage}%
\hfill
\begin{minipage}[b]{0.48\textwidth}
    \includegraphics[width=\textwidth,keepaspectratio]{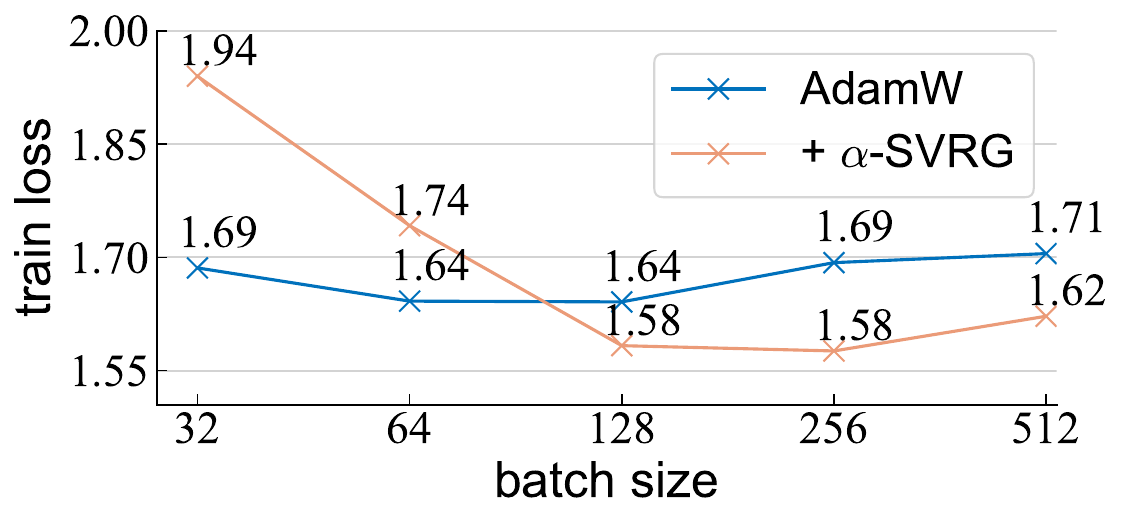}
    \vspace*{-2em}
    \caption{\textbf{Batch size.} \approach's effectiveness is observed for larger batch sizes.}
    \label{fig:batch}
\end{minipage}
\end{figure}

\textbf{Coefficient value.} We investigate the impact of the initial value of the coefficient \(\alpha_0\) for \approach. We vary it between 0 and 1 and observe its effect on the training loss. The results are presented in Figure~\ref{fig:coef}. The favorable range for initial values in \approach\ is quite broad, ranging from 0.2 to 0.9. This robustness indicates \approach\ requires minimal tuning in the practical setting. 

\textbf{Batch size.} Since the batch size controls the variance among mini-batch data, we change the batch size to understand how it affects \approach. We also scale the learning rate linearly~\citep{Goyal2017}. The default batch size is 128. In Figure~\ref{fig:batch}, we can see that \approach\ leads to a lower training loss when the batch size is larger, but it is worse than the baseline when the batch size is smaller. This may stem from the weakening correlation between snapshot gradients and model gradients as the batch size decreases. Therefore, a sufficiently large batch size is essential for \approach.

\begin{figure*}[h]
    \centering
    \begin{subfigure}[t]{0.49\linewidth}
        \centering
        \includegraphics[width=\linewidth,keepaspectratio]{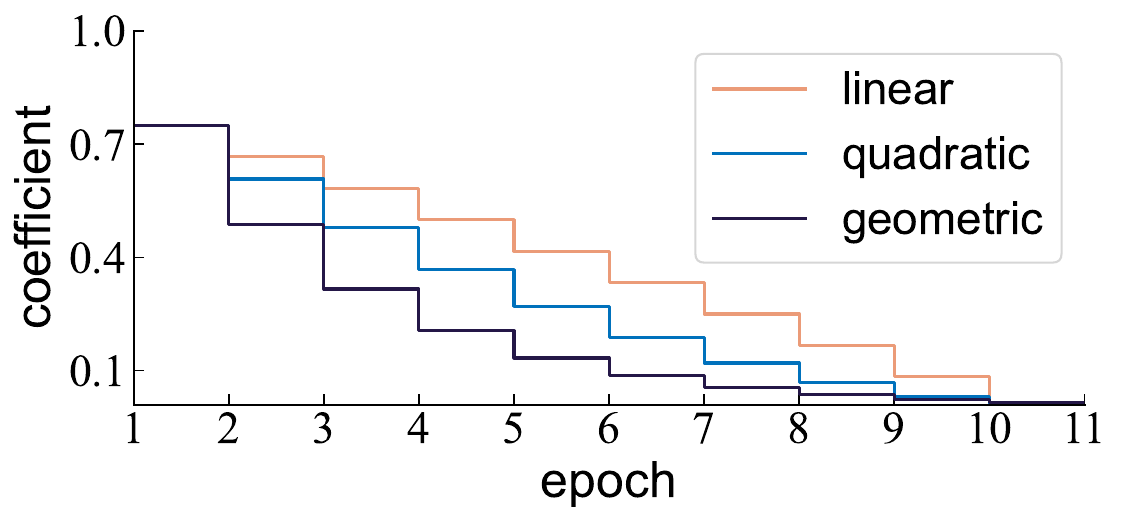}
        \centering
        \vspace{-1.5em}
        \caption{global schedules}
        \label{fig:global_schedules}
    \end{subfigure}%
    \begin{subfigure}[t]{0.49\linewidth}
        \centering
        \includegraphics[width=\linewidth,keepaspectratio]{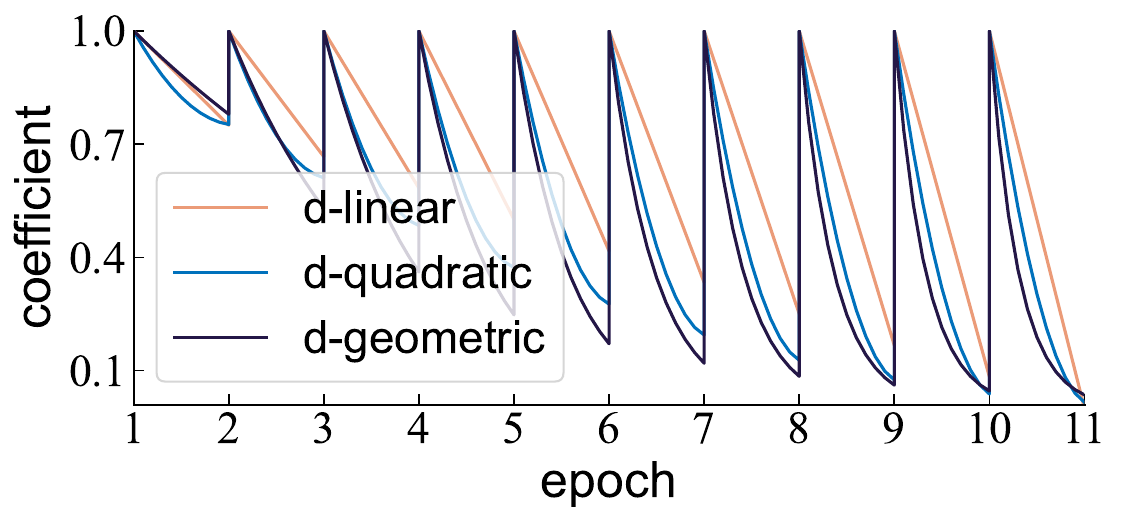}
        \centering
        \vspace{-1.5em}
        \caption{double schedules}
        \label{fig:double_schedules}
    \end{subfigure}
    \vspace*{-0.1cm}
    \caption{\textbf{Different coefficient schedules with $\boldsymbol{\alpha_0=0.75}$.} Each global schedule (left) maintains a static coefficient within an epoch and applies a coefficient decay only at the end of each epoch. In contrast, each double schedule (right) also adjusts the coefficient within an epoch.}
    \label{fig:schedule}
\end{figure*}

\textbf{Coefficient schedule.} By default, our $\alpha$-SVRG uses a linearly decreasing schedule to adjust the coefficient. Below we explore other schedules and illustrate them in Figure~\ref{fig:schedule}. Global schedules only decay the coefficient \emph{across epochs and keep as a constant within an epoch}. In contrast, double schedules also model the local decay in each epoch (Figure~\ref{fig:depth}) by initiating the coefficient at 1 and decreasing to an ending value specified by the global decay. More details on them are in Appendix~\ref{appendix:schedule}.

Table~\ref{tab:schedule} presents the results of {\approach} using each schedule. \approach\ with double schedules surprisingly have a higher training loss than the AdamW baseline (1.64). This is possibly because the coefficient within an epoch sometimes overestimates the optimal coefficient and therefore increases gradient variance. In contrast, \approach\ with global schedules consistently achieves a lower training loss than the baseline (1.64) regardless of the choice of any initial coefficient.  

\begin{table}[h]
\def\arraystretch{1.3}
\centering
\small
\begin{tabular}{lcccccc}
train loss & linear & quadratic & geometric & d-linear & d-quadratic & d-geometric \\
\Xhline{0.7pt}
\( \alpha_0=0.5\) & \textbf{1.59}	& 1.61	& 1.62	& 2.07	& 1.97	& 1.81\\
\( \alpha_0=0.75\) & \textbf{1.57} & 1.58 & 1.58 & 2.07 & 2.00 & 1.93\\
\( \alpha_0=1\) & 1.57 &	\textbf{1.56} &	1.57 &	2.00 &	1.97 & 1.88\\
\end{tabular}
\caption{\textbf{Schedules.} \textbf{Bold} indicates the lowest training loss among different schedules using the same initial coefficient (row). \approach\ with global schedules outperforms that with double schedules.}
\label{tab:schedule}
\end{table}

\begin{figure}[h]
\centering
    \centering
    \includegraphics[width=.48\textwidth,keepaspectratio]{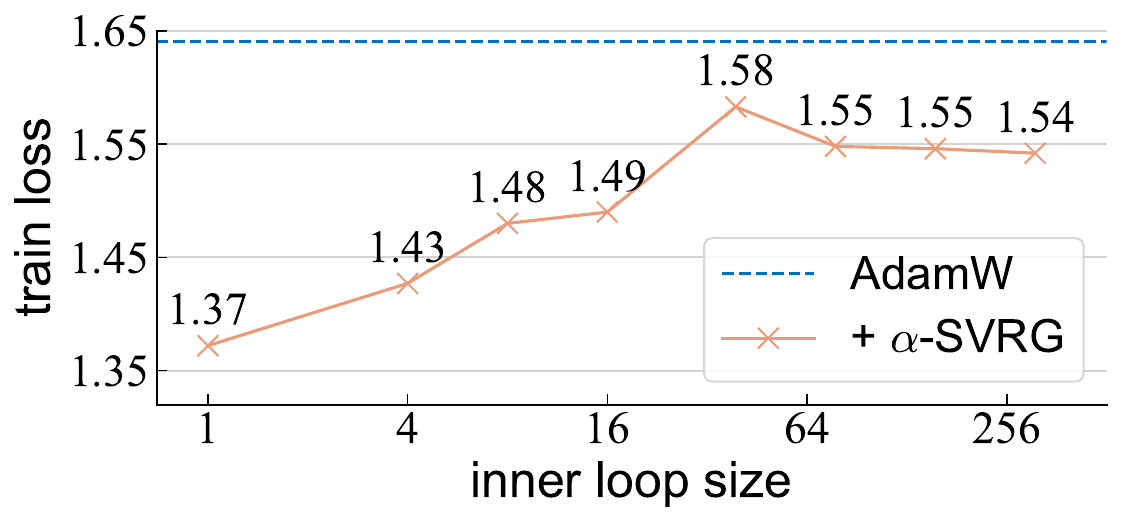}
    \vspace*{-.5em}
    \caption{\textbf{Inner loop size.} Although a greater inner loop size leads to a weakening correlation between the model gradients and the snapshot gradients, \approach\ can still lower training loss.}
    
    \label{fig:snapshot}
\end{figure}

\textbf{Inner loop size.} The inner loop size specifies the number of iterations between two consecutive snapshot captures. We vary it from 1 to 312 iterations to understand its effect on \approach. The default value is 39 iterations (one epoch). Figure~\ref{fig:snapshot} illustrates \approach\ has a lower training loss than the baseline even with a larger inner loop size, where the snapshot is relatively distant from the current model. On the other hand, a smaller inner loop size results in a lower training loss but requires additional training time, as a full gradient must be calculated each time a snapshot is taken.

\section{Related Work}
\textbf{Variance reduction in optimization.} There are a range of methods aiming at reducing gradient variance by directly modifying stochastic gradient. Initial works~\citep{johnson2013accelerating, sag} focus on simple convex settings. Subsequent research enhances these methods~\citep{defazio2014saga, miso, lin2015universal, defazio2016simple, allenzhu2018katyusha, lin2018catalyst} or handles finite sums in non-convex landscapes~\citep{allenzhu2016variance, nguyen2017sarah, fang2018spider, li2018simple, cutkosky2020momentumbasedvariancereductionnonconvex, elibol2020variance, zhou2020stochastic, adaspider}. For these methods, we either need to store all gradient with respect to each individual data point~\citep{defazio2014finito, sdca, li2021zerosarah} or calculate full gradient periodically~\citep{johnson2013accelerating, fang2018spider}. \citet{survey} provide a comprehensive review for variance reduction methods. While these studies focus on theories of SVRG, we primarily explore the practical utility of SVRG for real-world neural networks.

One of the most relevant works to us is MARS~\citep{yuan2024mars}, which also demonstrates that using a coefficient helps the variance reduction methods optimize modern neural networks. In contrast, our work studies how a coefficient makes SVRG effective through step-by-step controlled experiments.

\textbf{Implicit variance reduction.} Apart from methods that explicitly adjust the gradient, there are variance reduction techniques that implicitly reduce gradient variance through other means. A variety of optimizers~\citep{adadelta, adam, nadam, adagrad, adamw, radam, sophia, lion} utilize momentum to mitigate gradient variance. They achieve this by averaging past gradients exponentially, thus stabilizing subsequent updates. Lookahead optimizer~\citep{lookahead} reduces gradient variance by only updating the model once every $k$ iterations. Dropout~\citep{dropout1} is also found to reduce gradient variance and better optimize models when used at early training~\citep{liu2023dropout}.
\section{Conclusion}
Over the past decade, SVRG has been a method with a significant impact on the theory of optimization. In this work, we explore the effectiveness of SVRG in training real-world neural networks. Our key insight is the optimal strength for the variance reduction term in SVRG is not necessarily 1. It should be lower for deeper networks and decrease as training advances. This motivates us to introduce \approach: applying a linearly decreasing coefficient $\alpha$ to SVRG. \approach\ leads to a steady reduction in gradient variance and optimizes models better. Our experiments show that \approach\ consistently achieves a lower training loss compared to both baseline and the standard SVRG. Our results motivate further research of variance reduction methods in neural networks training.

\textbf{Acknowledgement.} We would like to thank Kaiming He, Aaron Defazio, Zeyuan Allen-Zhu, Kirill Vishniakov, Huijin Ou, Jiayi Xu, Shuyi Wang, and Zekai Wang for valuable discussions and feedback.
\clearpage
\newpage
\bibliography{main}

\begin{thebibliography}{65}
\providecommand{\natexlab}[1]{#1}
\providecommand{\url}[1]{\texttt{#1}}
\expandafter\ifx\csname urlstyle\endcsname\relax
  \providecommand{\doi}[1]{doi: #1}\else
  \providecommand{\doi}{doi: \begingroup \urlstyle{rm}\Url}\fi

\bibitem[Allen-Zhu(2017)]{allenzhu2018katyusha}
Zeyuan Allen-Zhu.
\newblock Katyusha: The first direct acceleration of stochastic gradient methods.
\newblock \emph{Symposium on Theory of Computing}, 2017.

\bibitem[Allen-Zhu \& Hazan(2016)Allen-Zhu and Hazan]{allenzhu2016variance}
Zeyuan Allen-Zhu and Elad Hazan.
\newblock Variance reduction for faster non-convex optimization.
\newblock In \emph{ICML}, 2016.

\bibitem[Bossard et~al.(2014)Bossard, Guillaumin, and Gool]{food101}
Lukas Bossard, Matthieu Guillaumin, and Luc~Van Gool.
\newblock Food-101--mining discriminative components with random forests.
\newblock In \emph{ECCV}, 2014.

\bibitem[Chen et~al.(2023)Chen, Liang, Huang, Real, Wang, Liu, Pham, Dong, Luong, Hsieh, Lu, and Le]{lion}
Xiangning Chen, Chen Liang, Da~Huang, Esteban Real, Kaiyuan Wang, Yao Liu, Hieu Pham, Xuanyi Dong, Thang Luong, Cho-Jui Hsieh, Yifeng Lu, and Quoc Le.
\newblock Symbolic discovery of optimization algorithms.
\newblock \emph{arXiv preprint arXiv:2302.06675}, 2023.

\bibitem[Cimpoi et~al.(2014)Cimpoi, Maji, Kokkinos, Mohamed, and Vedaldi]{dtd}
Mircea Cimpoi, Subhransu Maji, Iasonas Kokkinos, Sammy Mohamed, and Andrea Vedaldi.
\newblock Describing textures in the wild.
\newblock In \emph{CVPR}, 2014.

\bibitem[Coates et~al.(2011)Coates, Ng, and Lee]{stl10}
Adam Coates, Andrew Ng, and Honglak Lee.
\newblock An analysis of single-layer networks in unsupervised feature learning.
\newblock In \emph{AISTATS}, 2011.

\bibitem[Cubuk et~al.(2020)Cubuk, Zoph, Shlens, and Le]{randaugment}
Ekin Cubuk, Barret Zoph, Jonathon Shlens, and Quoc Le.
\newblock Randaugment: Practical automated data augmentation with a reduced search space.
\newblock In \emph{CVPR Workshops}, 2020.

\bibitem[Cutkosky \& Orabona(2019)Cutkosky and Orabona]{cutkosky2020momentumbasedvariancereductionnonconvex}
Ashok Cutkosky and Francesco Orabona.
\newblock Momentum-based variance reduction in non-convex sgd.
\newblock In \emph{NeurIPS}, 2019.

\bibitem[Damian et~al.(2021)Damian, Ma, and Lee]{damian2021label}
Alex Damian, Tengyu Ma, and Jason~D Lee.
\newblock Label noise sgd provably prefers flat global minimizers.
\newblock In \emph{NeurIPS}, 2021.

\bibitem[Defazio(2016)]{defazio2016simple}
Aaron Defazio.
\newblock A simple practical accelerated method for finite sums.
\newblock In \emph{NeurIPS}, 2016.

\bibitem[Defazio \& Bottou(2019)Defazio and Bottou]{defazio2019ineffectiveness}
Aaron Defazio and Léon Bottou.
\newblock On the ineffectiveness of variance reduced optimization for deep learning.
\newblock \emph{arXiv preprint arXiv:1812.04529}, 2019.

\bibitem[Defazio et~al.(2014{\natexlab{a}})Defazio, Bach, and Lacoste-Julien]{defazio2014saga}
Aaron Defazio, Francis Bach, and Simon Lacoste-Julien.
\newblock Saga: A fast incremental gradient method with support for non-strongly convex composite objectives.
\newblock In \emph{NeurIPS}, 2014{\natexlab{a}}.

\bibitem[Defazio et~al.(2014{\natexlab{b}})Defazio, Caetano, and Domke]{defazio2014finito}
Aaron Defazio, Tibério Caetano, and Justin Domke.
\newblock Finito: A faster, permutable incremental gradient method for big data problems.
\newblock In \emph{ICML}, 2014{\natexlab{b}}.

\bibitem[Deng et~al.(2009)Deng, Dong, Socher, Li, Li, and Fei-Fei]{ImageNet1k}
Jia Deng, Wei Dong, Richard Socher, Li-Jia Li, Kai Li, and Li~Fei-Fei.
\newblock {ImageNet: A large-scale hierarchical image database}.
\newblock In \emph{CVPR}, 2009.

\bibitem[Dosovitskiy et~al.(2021)Dosovitskiy, Beyer, Kolesnikov, Weissenborn, Zhai, Unterthiner, Dehghani, Minderer, Heigold, Gelly, Uszkoreit, and Houlsby]{vit}
Alexey Dosovitskiy, Lucas Beyer, Alexander Kolesnikov, Dirk Weissenborn, Xiaohua Zhai, Thomas Unterthiner, Mostafa Dehghani, Matthias Minderer, Georg Heigold, Sylvain Gelly, Jakob Uszkoreit, and Neil Houlsby.
\newblock An image is worth 16x16 words: Transformers for image recognition at scale.
\newblock In \emph{ICLR}, 2021.

\bibitem[Dozat(2016)]{nadam}
Timothy Dozat.
\newblock Incorporating {Nesterov Momentum into Adam}.
\newblock In \emph{ICLR}, 2016.

\bibitem[Dubois-Taine et~al.(2021)Dubois-Taine, Vaswani, Babanezhad, Schmidt, and Lacoste-Julien]{adasvrg}
Benjamin Dubois-Taine, Sharan Vaswani, Reza Babanezhad, Mark Schmidt, and Simon Lacoste-Julien.
\newblock Svrg meets adagrad: Painless variance reduction.
\newblock \emph{arXiv preprint arXiv:2102.09645}, 2021.

\bibitem[Elibol et~al.(2020)Elibol, Lei, and Jordan]{elibol2020variance}
Melih Elibol, Lihua Lei, and Michael Jordan.
\newblock Variance reduction with sparse gradients.
\newblock In \emph{ICLR}, 2020.

\bibitem[Fang et~al.(2018)Fang, Li, Lin, and Zhang]{fang2018spider}
Cong Fang, Chris~Junchi Li, Zhouchen Lin, and Tong Zhang.
\newblock Spider: Near-optimal non-convex optimization via stochastic path integrated differential estimator.
\newblock In \emph{NeurIPS}, 2018.

\bibitem[Gower et~al.(2020)Gower, Schmidt, Bach, and Richtárik]{survey}
Robert Gower, Mark Schmidt, Francis Bach, and Peter Richtárik.
\newblock Variance-reduced methods for machine learning.
\newblock In \emph{IEEE}, 2020.

\bibitem[Goyal et~al.(2017)Goyal, Doll{\'a}r, Girshick, Noordhuis, Wesolowski, Kyrola, Tulloch, Jia, and He]{Goyal2017}
Priya Goyal, Piotr Doll{\'a}r, Ross Girshick, Pieter Noordhuis, Lukasz Wesolowski, Aapo Kyrola, Andrew Tulloch, Yangqing Jia, and Kaiming He.
\newblock Accurate, large minibatch {SGD}: Training {ImageNet} in 1 hour.
\newblock \emph{arXiv preprint arXiv:1706.02677}, 2017.

\bibitem[He et~al.(2016)He, Zhang, Ren, and Sun]{He2016}
Kaiming He, Xiangyu Zhang, Shaoqing Ren, and Jian Sun.
\newblock Deep residual learning for image recognition.
\newblock In \emph{CVPR}, 2016.

\bibitem[Helber et~al.(2019)Helber, Bischke, Dengel, and Borth]{eurosat1}
Patrick Helber, Benjamin Bischke, Andreas Dengel, and Damian Borth.
\newblock Eurosat: A novel dataset and deep learning benchmark for land use and land cover classification.
\newblock \emph{IEEE Journal of Selected Topics in Applied Earth Observations and Remote Sensing}, 2019.

\bibitem[Hinton et~al.(2012)Hinton, Srivastava, Krizhevsky, Sutskever, and Salakhutdinov]{dropout1}
Geoffrey Hinton, Nitish Srivastava, Alex Krizhevsky, Ilya Sutskever, and Ruslan~R Salakhutdinov.
\newblock Improving neural networks by preventing co-adaptation of feature detectors.
\newblock \emph{arXiv preprint arXiv:1207.0580}, 2012.

\bibitem[Huang et~al.(2016)Huang, Sun, Liu, Sedra, and Weinberger]{stochasticdepth}
Gao Huang, Yu~Sun, Zhuang Liu, Daniel Sedra, and Kilian~Q Weinberger.
\newblock Deep networks with stochastic depth.
\newblock In \emph{ECCV}, 2016.

\bibitem[Jastrzebski et~al.(2020)Jastrzebski, Szymczak, Fort, Arpit, Tabor, Cho, and Geras]{jastrzebski2020breakeven}
Stanislaw Jastrzebski, Maciej Szymczak, Stanislav Fort, Devansh Arpit, Jacek Tabor, Kyunghyun Cho, and Krzysztof Geras.
\newblock The break-even point on optimization trajectories of deep neural networks.
\newblock In \emph{ICLR}, 2020.

\bibitem[Johnson \& Zhang(2013)Johnson and Zhang]{johnson2013accelerating}
Rie Johnson and Tong Zhang.
\newblock Accelerating stochastic gradient descent using predictive variance reduction.
\newblock In \emph{NeurIPS}, 2013.

\bibitem[Kavis et~al.(2022)Kavis, Skoulakis, Antonakopoulos, Dadi, and Cevher]{adaspider}
Ali Kavis, Stratis Skoulakis, Kimon Antonakopoulos, Leello~Tadesse Dadi, and Volkan Cevher.
\newblock Adaptive stochastic variance reduction for non-convex finite-sum minimization.
\newblock In \emph{NeurIPS}, 2022.

\bibitem[Kingma \& Ba(2015)Kingma and Ba]{adam}
Diederik Kingma and Jimmy Ba.
\newblock Adam: A method for stochastic optimization.
\newblock In \emph{ICLR}, 2015.

\bibitem[Krizhevsky(2009)]{cifar}
Alex Krizhevsky.
\newblock Learning multiple layers of features from tiny images.
\newblock \emph{Tech Report}, 2009.

\bibitem[Lavenberg et~al.(1977)Lavenberg, Moeller, and Welch]{lavenberg1977application}
Stephen Lavenberg, Thomas~L Moeller, and Peter~D Welch.
\newblock The application of control variables to the simulation of closed queueing networks.
\newblock In \emph{Proceedings of the 9th conference on Winter simulation-Volume 1}, 1977.

\bibitem[Lei et~al.(2017)Lei, Ju, Chen, and Jordan]{lei2017nonconvex}
Lihua Lei, Cheng Ju, Jianbo Chen, and Michael Jordan.
\newblock Non-convex finite-sum optimization via scsg methods.
\newblock In \emph{NeurIPS}, 2017.

\bibitem[Li et~al.(2022)Li, Wang, and Arora]{li2022happenssgdreacheszero}
Zhiyuan Li, Tianhao Wang, and Sanjeev Arora.
\newblock What happens after sgd reaches zero loss? --a mathematical framework.
\newblock In \emph{ICLR}, 2022.

\bibitem[Li \& Li(2018)Li and Li]{li2018simple}
Zhize Li and Jian Li.
\newblock A simple proximal stochastic gradient method for nonsmooth nonconvex optimization.
\newblock In \emph{NeurIPS}, 2018.

\bibitem[Li et~al.(2021)Li, Hanzely, and Richt{\'a}rik]{li2021zerosarah}
Zhize Li, Slavom{\'\i}r Hanzely, and Peter Richt{\'a}rik.
\newblock Zerosarah: Efficient nonconvex finite-sum optimization with zero full gradient computation.
\newblock \emph{arXiv preprint arXiv:2103.01447}, 2021.

\bibitem[Lin et~al.(2015)Lin, Mairal, and Harchaoui]{lin2015universal}
Hongzhou Lin, Julien Mairal, and Zaid Harchaoui.
\newblock A universal catalyst for first-order optimization.
\newblock In \emph{NeurIPS}, 2015.

\bibitem[Lin et~al.(2018)Lin, Mairal, and Harchaoui]{lin2018catalyst}
Hongzhou Lin, Julien Mairal, and Zaid Harchaoui.
\newblock Catalyst acceleration for first-order convex optimization: from theory to practice.
\newblock \emph{JMLR}, 2018.

\bibitem[Liu et~al.(2024)Liu, Li, Hall, Liang, and Ma]{sophia}
Hong Liu, Zhiyuan Li, David Hall, Percy Liang, and Tengyu Ma.
\newblock Sophia: A scalable stochastic second-order optimizer for language model pre-training.
\newblock In \emph{ICLR}, 2024.

\bibitem[Liu et~al.(2021{\natexlab{a}})Liu, Jiang, He, Chen, Liu, Gao, and Han]{radam}
Liyuan Liu, Haoming Jiang, Pengcheng He, Weizhu Chen, Xiaodong Liu, Jianfeng Gao, and Jiawei Han.
\newblock On the variance of the adaptive learning rate and beyond.
\newblock In \emph{ICLR}, 2021{\natexlab{a}}.

\bibitem[Liu et~al.(2021{\natexlab{b}})Liu, Lin, Cao, Hu, Wei, Zhang, Lin, and Guo]{swin}
Ze~Liu, Yutong Lin, Yue Cao, Han Hu, Yixuan Wei, Zheng Zhang, Stephen Lin, and Baining Guo.
\newblock Swin transformer: Hierarchical vision transformer using shifted windows.
\newblock In \emph{ICCV}, 2021{\natexlab{b}}.

\bibitem[Liu et~al.(2022)Liu, Mao, Wu, Feichtenhofer, Darrell, and Xie]{convnextzhuang}
Zhuang Liu, Hanzi Mao, Chao-Yuan Wu, Christoph Feichtenhofer, Trevor Darrell, and Saining Xie.
\newblock A convnet for the 2020s.
\newblock In \emph{CVPR}, 2022.

\bibitem[Liu et~al.(2023)Liu, Xu, Jin, Shen, and Darrell]{liu2023dropout}
Zhuang Liu, Zhiqiu Xu, Joseph Jin, Zhiqiang Shen, and Trevor Darrell.
\newblock Dropout reduces underfitting.
\newblock In \emph{ICML}, 2023.

\bibitem[Loshchilov \& Hutter(2019)Loshchilov and Hutter]{adamw}
Ilya Loshchilov and Frank Hutter.
\newblock Decoupled weight decay regularization.
\newblock In \emph{ICLR}, 2019.

\bibitem[Lydia \& Francis(2019)Lydia and Francis]{adagrad}
Agnes Lydia and Sagayaraj Francis.
\newblock Adagrad—an optimizer for stochastic gradient descent.
\newblock \emph{JMLR}, 2019.

\bibitem[Mairal(2015)]{miso}
Julien Mairal.
\newblock Incremental majorization-minimization optimization with application to large-scale machine learning.
\newblock \emph{SIAM Journal on Optimization}, 2015.

\bibitem[Netzer et~al.(2011)Netzer, Wang, Coates, Bissacco, Wu, and Ng]{svhn}
Yuval Netzer, Tao Wang, Adam Coates, Alessandro Bissacco, Bo~Wu, and Andrew Ng.
\newblock Reading digits in natural images with unsupervised feature learning.
\newblock In \emph{NIPS Workshop}, 2011.

\bibitem[Nguyen et~al.(2017)Nguyen, Liu, Scheinberg, and Takáč]{nguyen2017sarah}
Lam Nguyen, Jie Liu, Katya Scheinberg, and Martin Takáč.
\newblock Sarah: A novel method for machine learning problems using stochastic recursive gradient.
\newblock In \emph{ICML}, 2017.

\bibitem[Nilsback \& Zisserman(2008)Nilsback and Zisserman]{flowers}
Maria-Elena Nilsback and Andrew Zisserman.
\newblock Automated flower classification over a large number of classes.
\newblock In \emph{Indian Conference on Computer Vision, Graphics \& Image Processing}, 2008.

\bibitem[Parkhi et~al.(2012)Parkhi, Vedaldi, Zisserman, and Jawahar]{pets}
Omkar Parkhi, Andrea Vedaldi, Andrew Zisserman, and CV~Jawahar.
\newblock Cats and dogs.
\newblock In \emph{CVPR}, 2012.

\bibitem[Reddi et~al.(2016)Reddi, Hefny, Sra, Poczos, and Smola]{reddi2016stochastic}
Sashank Reddi, Ahmed Hefny, Suvrit Sra, Barnabas Poczos, and Alex Smola.
\newblock Stochastic variance reduction for nonconvex optimization.
\newblock In \emph{ICML}, 2016.

\bibitem[Schmidt et~al.(2016)Schmidt, Roux, and Bach]{sag}
Mark Schmidt, Nicolas~Le Roux, and Francis Bach.
\newblock Minimizing finite sums with the stochastic average gradient.
\newblock \emph{arXiv preprint arXiv:1309.2388}, 2016.

\bibitem[Shalev-Shwartz \& Zhang(2013)Shalev-Shwartz and Zhang]{sdca}
Shai Shalev-Shwartz and Tong Zhang.
\newblock Stochastic dual coordinate ascent methods for regularized loss minimization.
\newblock \emph{JMLR}, 2013.

\bibitem[Smith et~al.(2020)Smith, Elsen, and De]{smith2020generalization}
Samuel Smith, Erich Elsen, and Soham De.
\newblock On the generalization benefit of noise in stochastic gradient descent.
\newblock In \emph{ICML}, 2020.

\bibitem[Szegedy et~al.(2016)Szegedy, Vanhoucke, Ioffe, Shlens, and Wojna]{label-smoothing}
Christian Szegedy, Vincent Vanhoucke, Sergey Ioffe, Jonathon Shlens, and Zbigniew Wojna.
\newblock Rethinking the inception architecture for computer vision.
\newblock In \emph{CVPR}, 2016.

\bibitem[Tolstikhin et~al.(2021)Tolstikhin, Houlsby, Kolesnikov, Beyer, Zhai, Unterthiner, Yung, Steiner, Keysers, Uszkoreit, Lucic, and Dosovitskiy]{mlp-mixer}
Ilya Tolstikhin, Neil Houlsby, Alexander Kolesnikov, Lucas Beyer, Xiaohua Zhai, Thomas Unterthiner, Jessica Yung, Andreas Steiner, Daniel Keysers, Jakob Uszkoreit, Mario Lucic, and Alexey Dosovitskiy.
\newblock Mlp-mixer: An all-mlp architecture for vision.
\newblock In \emph{NeurIPS}, 2021.

\bibitem[Wang \& Klabjan(2022)Wang and Klabjan]{adamsvrg}
Ruiqi Wang and Diego Klabjan.
\newblock Divergence results and convergence of a variance reduced version of adam.
\newblock \emph{arXiv preprint arXiv:2210.05607}, 2022.

\bibitem[Wightman(2019)]{convnext-f}
Ross Wightman.
\newblock {GitHub} repository: Pytorch image models.
\newblock \emph{GitHub repository}, 2019.

\bibitem[Xiao \& Zhang(2014)Xiao and Zhang]{xiao2014proximal}
Lin Xiao and Tong Zhang.
\newblock A proximal stochastic gradient method with progressive variance reduction.
\newblock \emph{SIAM Journal on Optimization}, 2014.

\bibitem[Yuan et~al.(2024)Yuan, Liu, Wu, Zhou, and Gu]{yuan2024mars}
Huizhuo Yuan, Yifeng Liu, Shuang Wu, Xun Zhou, and Quanquan Gu.
\newblock Mars: Unleashing the power of variance reduction for training large models.
\newblock \emph{arXiv preprint arXiv:2411.10438}, 2024.

\bibitem[Yun et~al.(2019)Yun, Han, Oh, Chun, Choe, and Yoo]{yun2019cutmix}
Sangdoo Yun, Dongyoon Han, Seong~Joon Oh, Sanghyuk Chun, Junsuk Choe, and Youngjoon Yoo.
\newblock Cutmix: Regularization strategy to train strong classifiers with localizable features.
\newblock In \emph{ICCV}, 2019.

\bibitem[Zeiler(2012)]{adadelta}
Matthew Zeiler.
\newblock Adadelta: An adaptive learning rate method.
\newblock \emph{arXiv preprint arXiv:1212.5701}, 2012.

\bibitem[Zhang et~al.(2018)Zhang, Cisse, Dauphin, and Lopez-Paz]{zhang2018mixup}
Hongyi Zhang, Moustapha Cisse, Yann Dauphin, and David Lopez-Paz.
\newblock mixup: Beyond empirical risk minimization.
\newblock In \emph{ICLR}, 2018.

\bibitem[Zhang et~al.(2019)Zhang, Lucas, Hinton, and Ba]{lookahead}
Michael Zhang, James Lucas, Geoffrey Hinton, and Jimmy Ba.
\newblock Lookahead optimizer: k steps forward, 1 step back.
\newblock In \emph{NeurIPS}, 2019.

\bibitem[Zhong et~al.(2020)Zhong, Zheng, Kang, Li, and Yang]{randomerasing}
Zhun Zhong, Liang Zheng, Guoliang Kang, Shaozi Li, and Yi~Yang.
\newblock Random erasing data augmentation.
\newblock In \emph{AAAI}, 2020.

\bibitem[Zhou et~al.(2020)Zhou, Xu, and Gu]{zhou2020stochastic}
Dongruo Zhou, Pan Xu, and Quanquan Gu.
\newblock Stochastic nested variance reduction for nonconvex optimization.
\newblock In \emph{NeurIPS}, 2020.

\end{thebibliography}
\bibliographystyle{iclr2025_conference}

\clearpage
\appendix

\section*{\Large{Appendix}}

\section{Derivation of the optimal coefficient}
\label{appendix:derivation}

We present the full derivation of the optimal coefficient for control variates:
\begin{gather}
    \min_{\alpha}\Var(\ermX^*)= \min_{\alpha}\Var(\ermX - \alpha \ermY)\\
    =\min_{\alpha}\Var(\ermX)-2\alpha \Cov(\ermX, \ermY)+\alpha^2\Var(\ermY).
\end{gather}
Differentiating the objective with respect to \(\alpha\), we can determine the optimal coefficient $\alpha^*$:
\begin{gather}
    2\alpha\Var(\ermY) -2 \Cov(\ermX, \ermY) = 0,\\
    \implies \alpha^* = \frac{\Cov(\ermX, \ermY)}{\Var(\ermY)}.\label{eq:10}
\end{gather}
Lastly, we can plug the definition of the correlation coefficient:
\begin{gather}
   \rho(\ermX, \ermY) = \frac{\Cov(\ermX, \ermY)}{\sigma(\ermX)\sigma(\ermY)} 
\end{gather}
into the optimal coefficient and rewrite Equation~\ref{eq:10} as:
\begin{gather}
    \alpha^*=\rho(\ermX, \ermY)\frac{\sigma(\ermX)}{\sigma(\ermY)}.
\end{gather}
\section{Experimental Settings}
\label{appendix:setting}
\textbf{Training recipe.}
Table~\ref{tab:recipe} outlines our training recipe. It is based on the setting in ConvNeXt~\citep{convnextzhuang}. For all experiments, the base learning rate is set at 4e-3, except for training ConvNeXt-F on ImageNet-1K using \approach, where increasing it to 8e-3 reduces training loss very much.
\begin{table}[h]
\centering
\small
\addtolength{\tabcolsep}{-.3pt}
\def\arraystretch{1.2}
\begin{tabular}{y{146}|x{98}}
config & value \\
\Xhline{0.7pt}
weight init  & trunc. normal (0.2) \\
optimizer & AdamW \\
base learning rate & 4e-3 \\
weight decay & 0.05 \\
optimizer momentum & $\beta_1, \beta_2=0.9, 0.999$ \\
learning rate schedule & cosine decay \\
warmup schedule & linear \\
randaugment \citep{randaugment} & (9, 0.5) \\
mixup \citep{zhang2018mixup} & 0.8 \\
cutmix \citep{yun2019cutmix} & 1.0 \\
random erasing \citep{randomerasing} & 0.25 \\
label smoothing \citep{label-smoothing} & 0.1
\end{tabular}
\caption{\textbf{Our basic training recipe,} adapted from ConvNeXt \citep{convnextzhuang}.}
\label{tab:recipe}
\end{table}

\begin{table}[bp]
    \centering
    \small
    \def\arraystretch{1.2}
    \addtolength{\tabcolsep}{-2.5pt}
    \begin{tabular}{lcccccccccc}
    &CIFAR-100 & Pets & Flowers & STL-10 & Food101 &DTD & SVHN & EuroSAT & ImageNet-1K \\
    \Xhline{0.7pt}
   batch size & 1024 & 128 & 128 & 128 & 1024 & 128 & 1024 & 512 & 4096\\
   warmup epochs & 50 & 100 & 100 & 50 & 50 & 100 & 20 & 40 & 50\\
   training epochs & 300 & 600 & 600 & 300 & 300 & 600 & 100 & 200 & 300
    \end{tabular}
    
    \caption{\textbf{Hyperparameter setting.}}
    \label{tab:hyper}
\end{table}
\textbf{Hyperparameters.}
Table~\ref{tab:hyper} lists the batch size, warmup epochs, and training epochs for each dataset. Note that these hyperparameters selections are done on the AdamW baseline. For each dataset, we set the batch size proportional to its total size and tune the training epochs to achieve reasonable performance. The warmup epochs are set to one-fifth or one-sixth of the total training epochs.

We do not use stochastic depth~\citep{stochasticdepth} on small models. For larger models, we adhere to the original work~\citep{vit, convnextzhuang}, using a stochastic depth rate of 0.4 for ViT-B and 0.5 for ConvNeXt-B. In these models, we maintain a consistent stochastic pattern between the current model and the snapshot at each iteration~\citep{defazio2019ineffectiveness}.

\section{Additional Results of $\alpha$-SVRG}
\label{appendix:additional_experiments}
In this section, we provide additional experiment results to demonstrate the effectiveness of {\approach}. This includes full results of {\approach} with different initial coefficients $\alpha_0$ on small image classification datasets (Appendix~\ref{appendix:full_coef_results}), training with three random seeds (Appendix~\ref{appendix:seed}), a full learning rate search (Appendix~\ref{appendix:lr}), and applying {\approach} only in the early training stage (Appendix~\ref{appendix:early_svrg}).
\subsection{Different Initial Coefficients}
\label{appendix:full_coef_results}
Table~\ref{tab:last} presents the performance of ConvNeXt-F trained with \approach\ using different initial coefficients $\alpha_0$ on various image classification datasets. \approach\ consistently reduces the training loss of ConvNeXt-F and enhances the validation accuracy on most datasets, regardless of the choice of initial coefficient $\alpha_0$. This demonstrates the robustness of \approach\ to the initial coefficient.
\begin{table}[h]
\centering
\small
\def\arraystretch{1.3}
\vspace{-.5em}
\begin{tabular}{lcccccccc}
& \multicolumn{2}{c}{CIFAR-100} & \multicolumn{2}{c}{Pets} & \multicolumn{2}{c}{Flowers} & \multicolumn{2}{c}{STL-10} \\
\Xhline{0.7pt}
AdamW & 2.659 & - & 2.203 & - & 2.400 & - & 1.641 & - \\
+\ SVRG & 2.937 & \worse{0.278} & 3.424 & \worse{1.221} & 2.256 & \better{0.144} & 1.899 & \worse{0.258} \\
\rowcolor{gray}
+\ \approach\ $(\alpha_0=0.5)$ &  \textbf{2.622} & \better{0.037} & 1.960 & \better{0.243} & 2.265 & \better{0.135} & 1.583 & \better{0.058}\\
\rowcolor{gray}
+\ \approach\ $(\alpha_0=0.75)$ & 2.646 & \better{0.013} & 2.004 & \better{0.199} & \textbf{2.162} & \better{0.238} & \textbf{1.568} & \better{0.073} \\
\rowcolor{gray}
+\ \approach\ $(\alpha_0=1)$ & 2.712 & \worse{0.053} & \textbf{1.994} & \better{0.209} & 2.259 & \better{0.141} & 1.573 & \better{0.068} \\
\end{tabular}

\centering
\small
\begin{tabular}{lcccccccccc}
& \multicolumn{2}{c}{Food-101} & \multicolumn{2}{c}{DTD} & \multicolumn{2}{c}{SVHN} & \multicolumn{2}{c}{EuroSAT} \\
\Xhline{0.7pt}
AdamW & 2.451 & - & 1.980 & - & 1.588 & - & 1.247 & - \\
+\ SVRG & 3.026 & \worse{0.575} & 2.009 & \worse{0.029} & 1.639 & \worse{0.051} & 1.249 & \worse{0.002} \\
\rowcolor{gray}
+\ \approach\ $(\alpha_0=0.5)$ & \textbf{2.423} & \better{0.028} & 1.865 & \better{0.115} & \textbf{1.572} & \better{0.016} & 1.243 & \better{0.004} \\
\rowcolor{gray}
+\ \approach\ $(\alpha_0=0.75)$ & 2.461 & \worse{0.010} & 1.829 & \better{0.151} & 1.573 & \better{0.015} & 1.237 & \better{0.010} \\
\rowcolor{gray}
+\ \approach\ $(\alpha_0=1)$ & 2.649 & \worse{0.198} & \textbf{1.790} & \better{0.190} & 1.585 & \better{0.003} & \textbf{1.230} & \better{0.017} \\
\end{tabular}
\caption*{(a) \textbf{training loss}}
\vspace{.5em}
\small
\begin{tabular}{lcccccccc}
& \multicolumn{2}{c}{CIFAR-100} & \multicolumn{2}{c}{Pets} & \multicolumn{2}{c}{Flowers} & \multicolumn{2}{c}{STL-10} \\
\Xhline{0.7pt}
AdamW & 81.0 & - & 72.8 & - & 80.8 & - & 82.3 & - \\
+\ SVRG & 78.2 & \worseinv{2.8} & 17.6 & \worseinv{55.2} &  82.6 & \betterinv{1.8} & 65.1 & \worseinv{17.2} \\
\rowcolor{gray}
+\ \approach\ $(\alpha_0=0.5)$ & \textbf{81.4} & \betterinv{0.4} & \textbf{77.8} & \betterinv{5.0} & \textbf{83.3} & \betterinv{2.5} & \textbf{83.5} & \betterinv{1.2} \\
\rowcolor{gray}
+\ \approach\ $(\alpha_0=0.75)$ & 80.6 & \worseinv{0.4} & 76.7 & \betterinv{3.9} & 82.6 & \betterinv{1.8}  & 84.0 & \betterinv{1.7} \\
\rowcolor{gray}
+\ \approach\ $(\alpha_0=1)$ & 80.0 & \worseinv{1.0} & 77.3 & \betterinv{4.5} & 81.9 & \betterinv{1.1} & 84.0 & \betterinv{1.7} \\
\end{tabular}

\centering
\small
\begin{tabular}{lcccccccccc}
& \multicolumn{2}{c}{Food-101} & \multicolumn{2}{c}{DTD} & \multicolumn{2}{c}{SVHN} & \multicolumn{2}{c}{Euro} \\
\Xhline{0.7pt}
AdamW & \textbf{85.9} & - & 57.9 & - & 94.9 & - & 98.1 & - \\
+\ SVRG & 79.6 & \worseinv{6.3} & 57.8 & \worseinv{0.1} & 95.7 & \betterinv{0.8} & 97.9 & \worseinv{0.2} \\
\rowcolor{gray}
+\ \approach\ $(\alpha_0=0.5)$ & \textbf{85.9} & \betterinv{0.0} & 57.0 & \worseinv{0.9} & 95.4 & \betterinv{0.5} & \textbf{98.2} & \betterinv{0.1} \\
\rowcolor{gray}
+\ \approach\ $(\alpha_0=0.75)$& 85.0 & \worseinv{0.9} & 60.3 & \betterinv{2.4} & 95.7 & \betterinv{0.8} & \textbf{98.2} & \betterinv{0.1} \\
\rowcolor{gray}
+\ \approach\ $(\alpha_0=1)$ & 83.8 & \worseinv{2.1} & \textbf{61.8} & \betterinv{3.9} & \textbf{95.8} & \betterinv{0.9} & \textbf{98.2} & \betterinv{0.1} \\
\end{tabular}
\caption*{(b) \textbf{validation accuracy}}
\caption{\textbf{Results on smaller classification datasets with different initial coefficients.} While SVRG negatively affects performance on most of these datasets, \approach\ consistently reduces the training loss and improves the validation accuracy for almost any initial coefficient on each dataset.}
\label{tab:last}
\vspace{-.5em}
\end{table}

\clearpage
\newpage

\subsection{Standard Deviation Results}
\label{appendix:seed}
Here we run the AdamW baseline and \approach\ in Table~\ref{tab:pre-trained2} with 3 random seeds. Table~\ref{tab:seed} presents the results. \approach\ decreases the mean training loss and improves the mean validation accuracy. The mean difference is usually larger than one standard deviation, indicating the reliability of \approach.
\begin{table}[h]
\centering
\small
\def\arraystretch{1.3}
\addtolength{\tabcolsep}{5pt}
\begin{tabular}{lcccccccc}
& \multicolumn{2}{c}{CIFAR-100} & \multicolumn{2}{c}{Pets} & \multicolumn{2}{c}{Flowers} & \multicolumn{2}{c}{STL-10} \\
\Xhline{0.7pt}
AdamW & \multicolumn{2}{c}{2.645\ \(\pm\)\ 0.013} & \multicolumn{2}{c}{2.326\ \(\pm\)\ 0.088} & \multicolumn{2}{c}{2.436\ \(\pm\)\ 0.038} & \multicolumn{2}{c}{1.660\ \(\pm\)\ 0.017} \\
\rowcolor{gray}
+\ \approach\ & \multicolumn{2}{c}{\textbf{2.606}\ \(\pm\)\ 0.017} & \multicolumn{2}{c}{\textbf{2.060}\ \(\pm\)\ 0.071} & \multicolumn{2}{c}{\textbf{2.221}\ \(\pm\)\ 0.042} & \multicolumn{2}{c}{\textbf{1.577}\ \(\pm\)\ 0.022}\\
\end{tabular}

\centering
\small
\begin{tabular}{lcccccccc}
& \multicolumn{2}{c}{Food-101} & \multicolumn{2}{c}{DTD} & \multicolumn{2}{c}{SVHN} & \multicolumn{2}{c}{EuroSAT} \\
\Xhline{0.7pt}
AdamW & \multicolumn{2}{c}{2.478\ \(\pm\)\ 0.021} & \multicolumn{2}{c}{2.072\ \(\pm\)\ 0.066} & \multicolumn{2}{c}{1.583\ \(\pm\)\ 0.005} & \multicolumn{2}{c}{1.259\ \(\pm\)\ 0.017}\\
\rowcolor{gray}
+\ \approach\ & \multicolumn{2}{c}{\textbf{2.426}\ \(\pm\)\ 0.007} & \multicolumn{2}{c}{\textbf{1.896}\ \(\pm\)\ 0.075} & \multicolumn{2}{c}{\textbf{1.572}\ \(\pm\)\ 0.011} & \multicolumn{2}{c}{\textbf{1.239}\ \(\pm\)\ 0.016}\\
\end{tabular}
\caption*{(a) \textbf{training loss}}
\vspace*{0.2cm}

\begin{tabular}{lcccccccc}
& \multicolumn{2}{c}{CIFAR-100} & \multicolumn{2}{c}{Pets} & \multicolumn{2}{c}{Flowers} & \multicolumn{2}{c}{STL-10} \\
\Xhline{0.7pt}
AdamW & \multicolumn{2}{c}{81.02\ \(\pm\)\ 0.07} & \multicolumn{2}{c}{70.61\ \(\pm\)\ 1.55} & \multicolumn{2}{c}{80.33\ \(\pm\)\ 1.01} & \multicolumn{2}{c}{80.80\ \(\pm\)\ 1.46} \\
\rowcolor{gray}
+\ \approach\ & \multicolumn{2}{c}{\textbf{81.07}\ \(\pm\)\ 0.22} & \multicolumn{2}{c}{\textbf{76.37}\ \(\pm\)\ 1.06} & \multicolumn{2}{c}{\textbf{84.15}\ \(\pm\)\ 1.15} & \multicolumn{2}{c}{\textbf{83.65}\ \(\pm\)\ 0.92} \\
\end{tabular}

\centering
\small
\begin{tabular}{lcccccccccc}
& \multicolumn{2}{c}{Food-101} & \multicolumn{2}{c}{DTD} & \multicolumn{2}{c}{SVHN} & \multicolumn{2}{c}{EuroSAT} \\
\Xhline{0.7pt}
AdamW & \multicolumn{2}{c}{85.29\ \(\pm\)\ 0.47} & \multicolumn{2}{c}{56.21\ \(\pm\)\ 1.19} & \multicolumn{2}{c}{94.29\ \(\pm\)\ 0.67} & \multicolumn{2}{c}{97.91\ \(\pm\)\ 0.12} \\
\rowcolor{gray}
+\ \approach\ & \multicolumn{2}{c}{\textbf{85.45}\ \(\pm\)\ 0.43} & \multicolumn{2}{c}{\textbf{61.44}\ \(\pm\)\ 0.35} & \multicolumn{2}{c}{\textbf{94.94}\ \(\pm\)\ 0.60} & \multicolumn{2}{c}{\textbf{98.13}\ \(\pm\)\ 0.07}
\end{tabular}
\caption*{(b) \textbf{validation accuracy}}
\caption{\textbf{Results of {$\boldsymbol{\alpha}$-SVRG} with AdamW with standard deviation.}}
\label{tab:seed}
\end{table}

In Section~\ref{sec:experi}, we primarily use AdamW as the base optimizer to study {\approach}. Here we switch the base optimizer in {\approach} from AdamW to SGD. Specifically, we train a ResNet-18~\citep{He2016}, which by default uses SGD to train, on the small image classification datasets. Following~\citet{He2016}, we use an initial learning rate of 0.1, which is divided by 10 when the error plateaus, a momentum of 0.9,  and a weight decay of 1e-4. Other settings in training recipe remain the same as Table~\ref{tab:recipe}. Table~\ref{tab:hyper2} details other hyperparameters, such as batch size and training epochs.

\begin{table}[th]
\centering
\small
\def\arraystretch{1.3}
\addtolength{\tabcolsep}{5pt}
\begin{tabular}{lcccccccc}
	
& \multicolumn{2}{c}{CIFAR-100} & \multicolumn{2}{c}{Pets} & \multicolumn{2}{c}{Flowers} & \multicolumn{2}{c}{STL-10} \\
\Xhline{0.7pt}
SGD & \multicolumn{2}{c}{3.118\ \(\pm\)\ 0.035} & \multicolumn{2}{c}{2.706\ \(\pm\)\ 0.095} & \multicolumn{2}{c}{2.822\ \(\pm\)\ 0.058} & \multicolumn{2}{c}{1.763\ \(\pm\)\ 0.032} \\
\rowcolor{gray}
+\ \approach\ & \multicolumn{2}{c}{\textbf{3.087}\ \(\pm\)\ 0.011} & \multicolumn{2}{c}{\textbf{2.655}\ \(\pm\)\ 0.134} & \multicolumn{2}{c}{\textbf{2.699}\ \(\pm\)\ 	0.049} & \multicolumn{2}{c}{\textbf{1.725}\ \(\pm\)\ 0.043}\\
\end{tabular}
	
\centering
\small
\begin{tabular}{lcccccccc}
& \multicolumn{2}{c}{Food-101} & \multicolumn{2}{c}{DTD} & \multicolumn{2}{c}{SVHN} & \multicolumn{2}{c}{EuroSAT} \\
\Xhline{0.7pt}
SGD & \multicolumn{2}{c}{3.424\ \(\pm\)\ 0.015} & \multicolumn{2}{c}{2.589\ \(\pm\)\ 0.032} & \multicolumn{2}{c}{1.789\ \(\pm\)\ 0.029} & \multicolumn{2}{c}{1.449\ \(\pm\)\ 0.029}\\
\rowcolor{gray}
+\ \approach\ & \multicolumn{2}{c}{\textbf{3.397}\ \(\pm\)\ 0.006} & \multicolumn{2}{c}{\textbf{2.543}\ \(\pm\)\ 0.039} & \multicolumn{2}{c}{\textbf{1.764}\ \(\pm\)\ 0.014} & \multicolumn{2}{c}{\textbf{1.412}\ \(\pm\)\ 0.011}\\
\end{tabular}
\caption*{(a) \textbf{training loss}}
\vspace*{0.2cm}

\begin{tabular}{lcccccccc}
& \multicolumn{2}{c}{CIFAR-100} & \multicolumn{2}{c}{Pets} & \multicolumn{2}{c}{Flowers} & \multicolumn{2}{c}{STL-10} \\
\Xhline{0.7pt}
SGD & \multicolumn{2}{c}{75.43\ \(\pm\)\ 0.88} & \multicolumn{2}{c}{71.25\ \(\pm\)\ 4.74} & \multicolumn{2}{c}{65.92\ \(\pm\)\ 4.24} & \multicolumn{2}{c}{76.09\ \(\pm\)\ 1.09} \\
\rowcolor{gray}
+\ \approach\ & \multicolumn{2}{c}{\textbf{75.93}\ \(\pm\)\ 0.83} & \multicolumn{2}{c}{\textbf{71.89}\ \(\pm\)\ 4.84} & \multicolumn{2}{c}{\textbf{74.98}\ \(\pm\)\ 3.14} & \multicolumn{2}{c}{\textbf{78.55}\ \(\pm\)\ 2.54} \\
\end{tabular}

\centering
\small
\begin{tabular}{lcccccccccc}
& \multicolumn{2}{c}{Food-101} & \multicolumn{2}{c}{DTD} & \multicolumn{2}{c}{SVHN} & \multicolumn{2}{c}{EuroSAT} \\
\Xhline{0.7pt}
SGD & \multicolumn{2}{c}{60.58\ \(\pm\)\ 2.00} & \multicolumn{2}{c}{57.53\ \(\pm\)\ 1.25} & \multicolumn{2}{c}{91.10\ \(\pm\)\ 1.31} & \multicolumn{2}{c}{
95.31\ \(\pm\)\ 0.46} \\
\rowcolor{gray}

+\ \approach\ & \multicolumn{2}{c}{\textbf{62.89}\ \(\pm\)\ 0.87} & \multicolumn{2}{c}{\textbf{58.44}\ \(\pm\)\ 1.05} & \multicolumn{2}{c}{\textbf{91.60}\ \(\pm\)\ 0.59} & \multicolumn{2}{c}{\textbf{96.17}\ \(\pm\)\ 0.17}
\end{tabular}
\caption*{(b) \textbf{validation accuracy}}
\caption{\textbf{Results of {$\boldsymbol{\alpha}$-SVRG} with SGD on smaller datasets with standard deviation.}}
\label{tab:seed_2}
\end{table}
We compare {\approach} ($\alpha_0$ = 0.5) equipped by a linear decreasing schedule to the baseline using only SGD. The results, shown in Table~\ref{tab:seed_2}, are based on the average of 3 runs with different random seeds. As we can see, {\approach} consistently outperforms the SGD baseline across all datasets.

\begin{table}[th]
    \centering
    \small
    \def\arraystretch{1.2}
    \addtolength{\tabcolsep}{-1.4pt}
    \begin{tabular}{lcccccccccc}
    &CIFAR-100 & Pets & Flowers & STL-10 & Food101 &DTD & SVHN & EuroSAT \\
    \Xhline{0.7pt}
   batch size & 1024 & 128 & 128 & 128 & 1024 & 128 & 1024 & 512 \\
   training epochs & 150 & 200 & 150 & 200 & 50 & 200 & 50  & 100 
    \end{tabular}
    \caption{\textbf{Hyperparameter setting for {$\boldsymbol{\alpha}$-SVRG} with SGD on ResNet-18.}}
    \label{tab:hyper2}
\end{table}

\subsection{Different Learning Rates}
\label{appendix:lr}
\begin{figure}[th]
\centering
\def\arraystretch{1}
  \setlength{\tabcolsep}{2pt} 
 \begin{tabular}{@{}cc@{}} 
 \centering
\includegraphics[width=.48\linewidth]{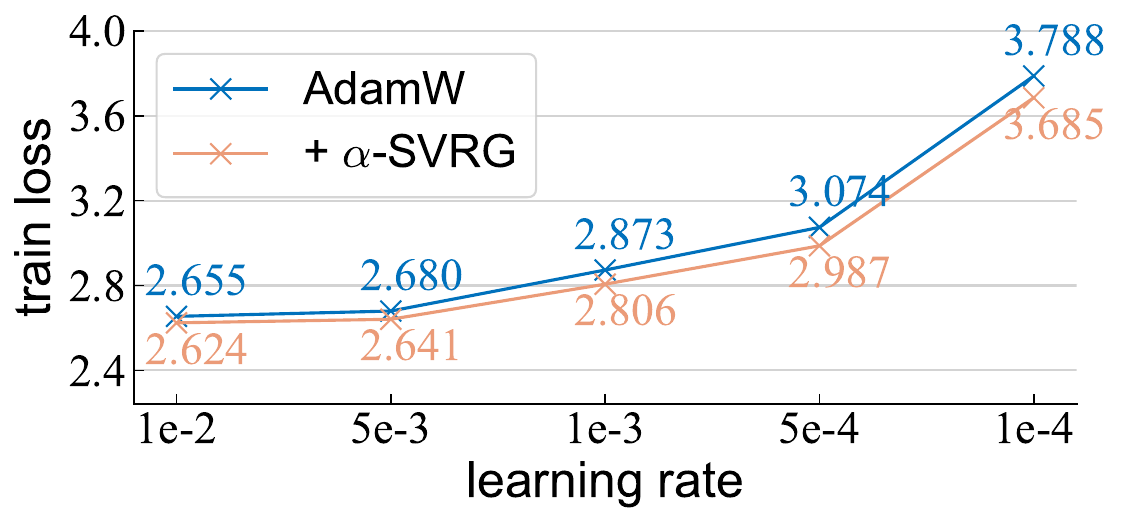} &
    \includegraphics[width=.48\linewidth]{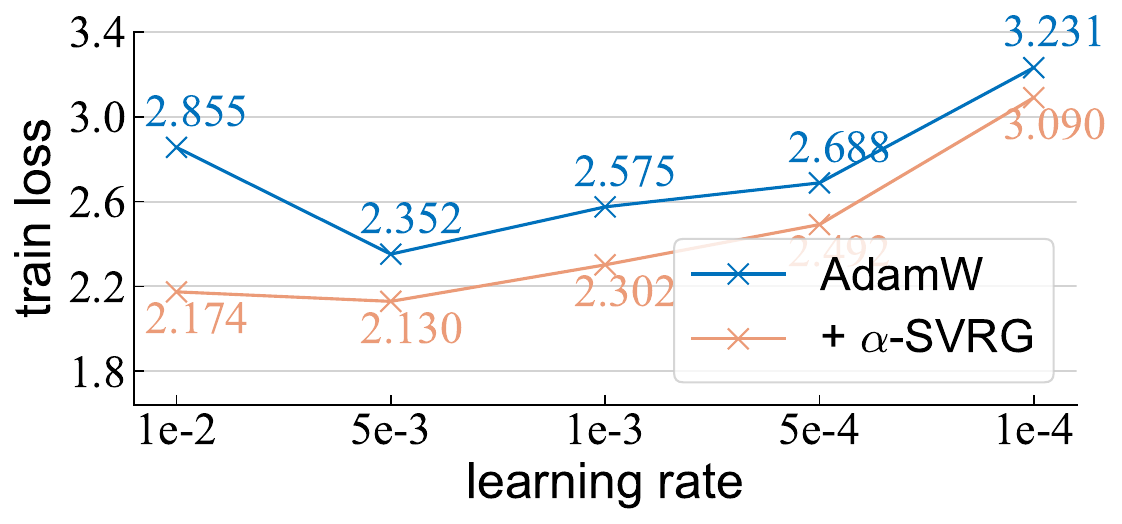} \\
    (a) CIFAR-100 & (b) Pets \\
    
    \includegraphics[width=.48\linewidth]{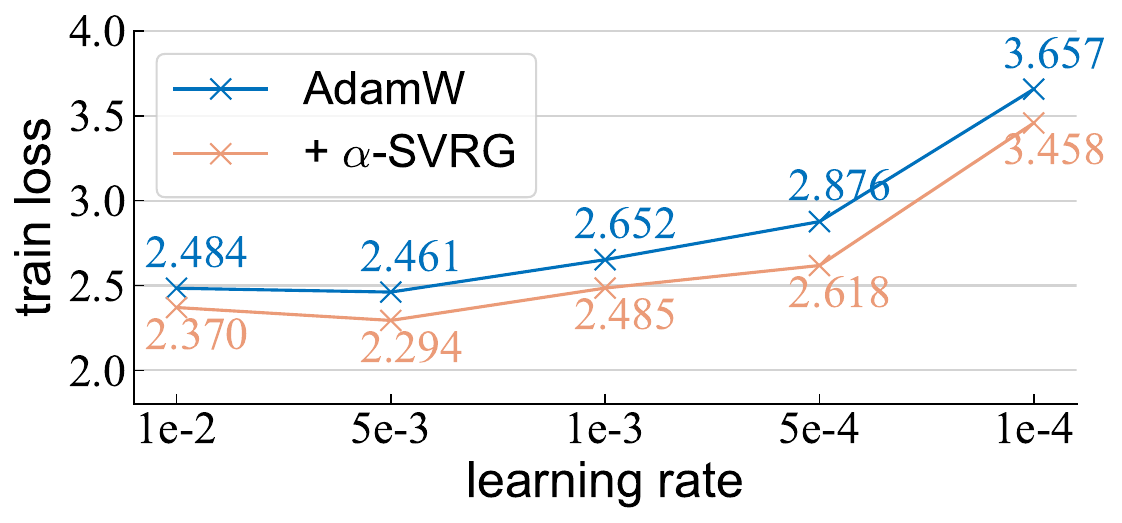} &
    \includegraphics[width=.48\linewidth]{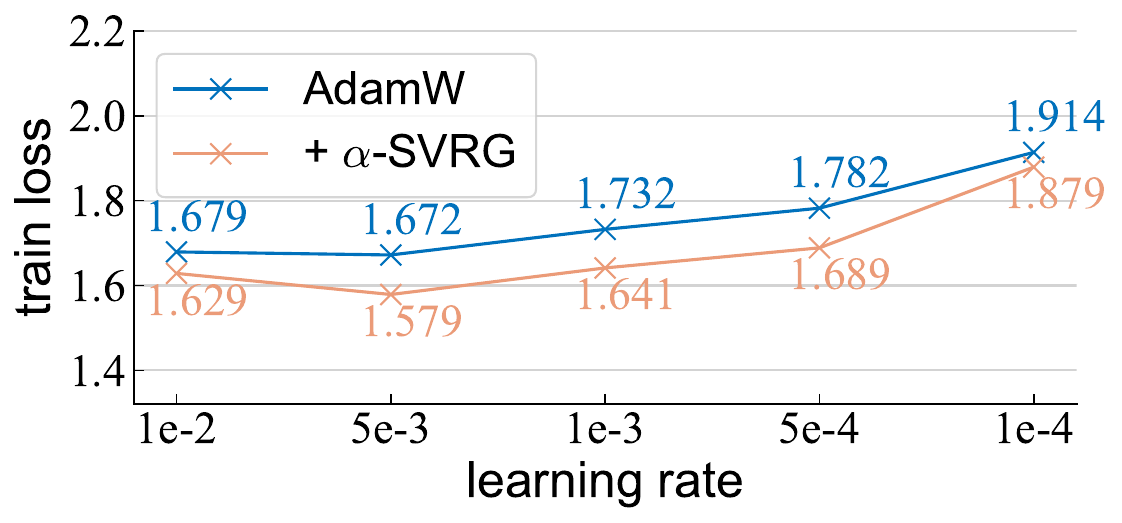} \\
    (c) Flowers & (d) STL-10 \\
    \includegraphics[width=.48\linewidth]{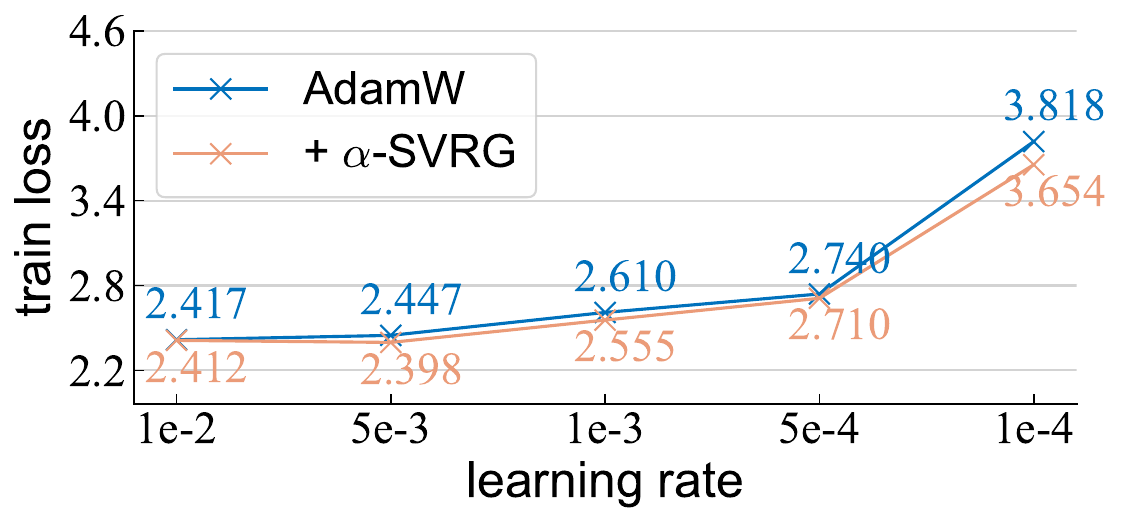} &
    \includegraphics[width=.48\linewidth]{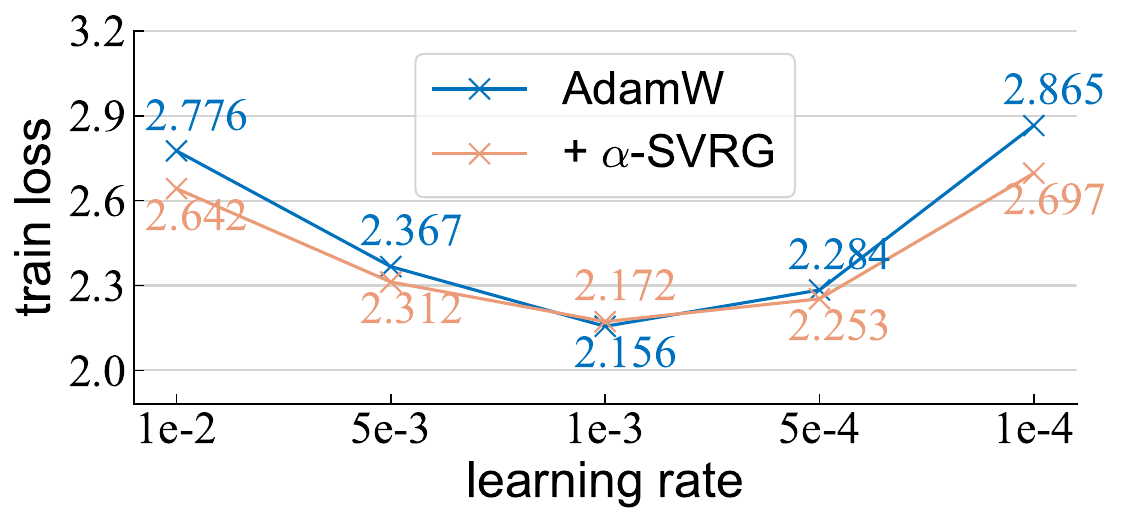} \\
    (e) Food-101 & (f) DTD \\
    \includegraphics[width=.48\linewidth]{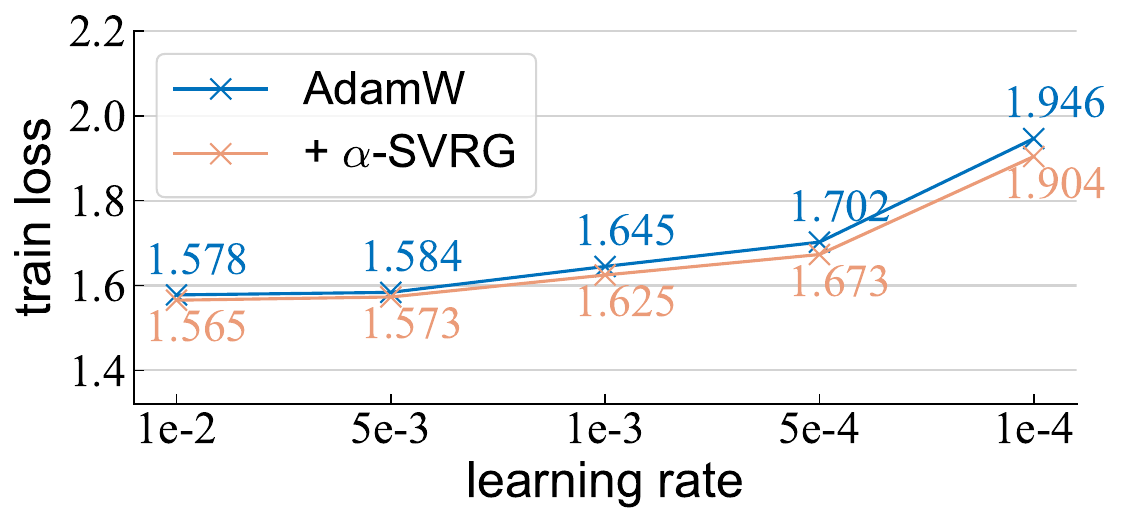} &
    \includegraphics[width=.48\linewidth]{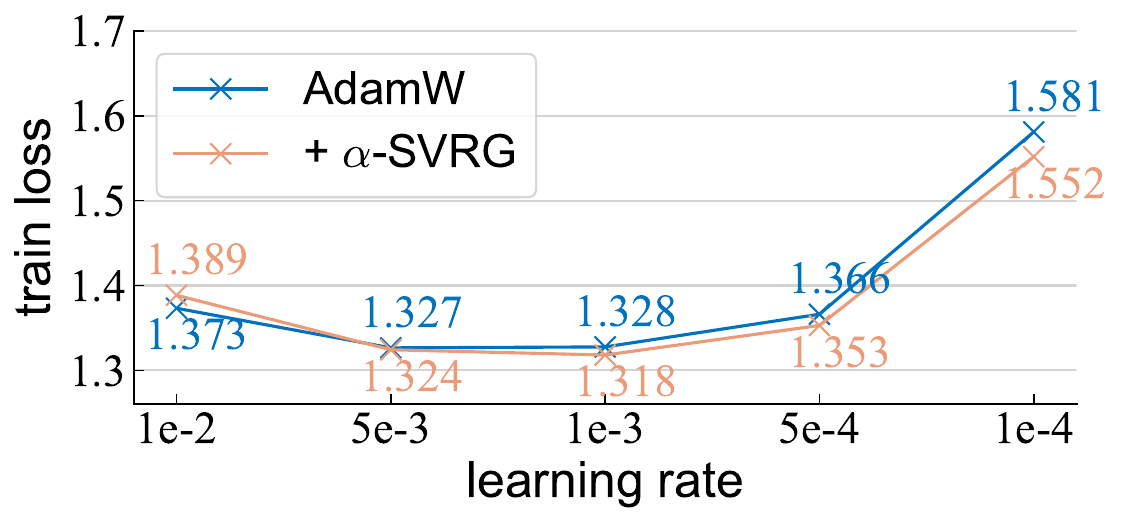} \\
    (g) SVHN & (h) EuroSAT 
    \end{tabular}
    
  \caption{\textbf{$\boldsymbol{\alpha}$-SVRG with AdamW at different learning rates}.}
  \label{fig:learning_rate}
  \end{figure}

\textbf{$\alpha$-SVRG with AdamW.} In Section~\ref{sec:experi}, we use the same base learning rate of 4e-3 for both AdamW and $\alpha$-SVRG. However, each method's optimal learning rate might be different. Here we sweep the base learning rate over the range $\{\text{1e-2}, \text{5e-3}, \text{1e-3}, \text{5e-4}, \text{1e-4}\}$ for both methods. As shown in Figure~\ref{fig:learning_rate}, $\alpha$-SVRG ($\alpha_0=0.5$) can decrease training loss better than vanilla AdamW in most cases.  

\begin{figure}[th]
\centering
\def\arraystretch{1}
  \setlength{\tabcolsep}{2pt} 
 \begin{tabular}{@{}cc@{}} 
 \centering
\includegraphics[width=.48\linewidth]{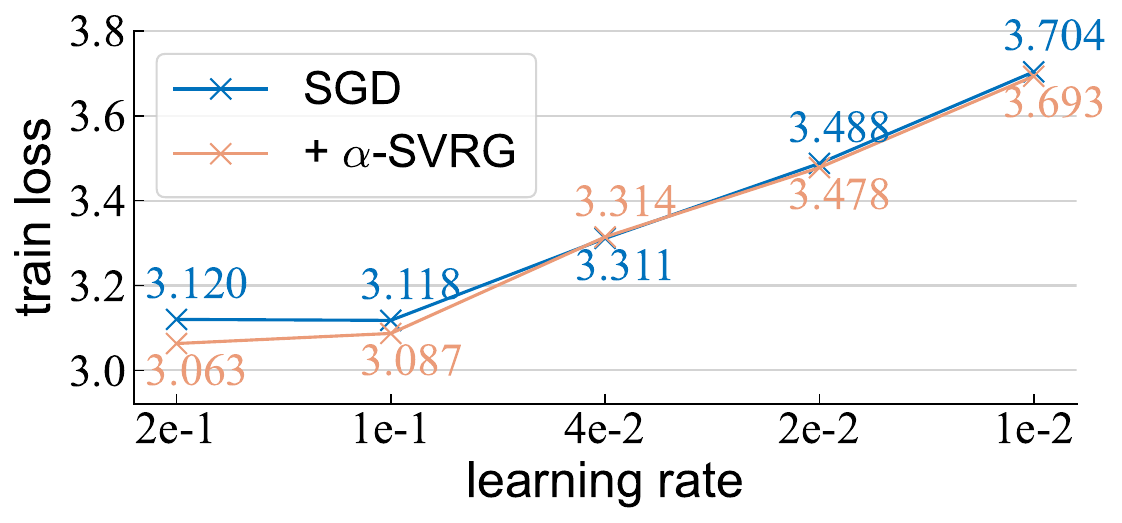} &
    \includegraphics[width=.48\linewidth]{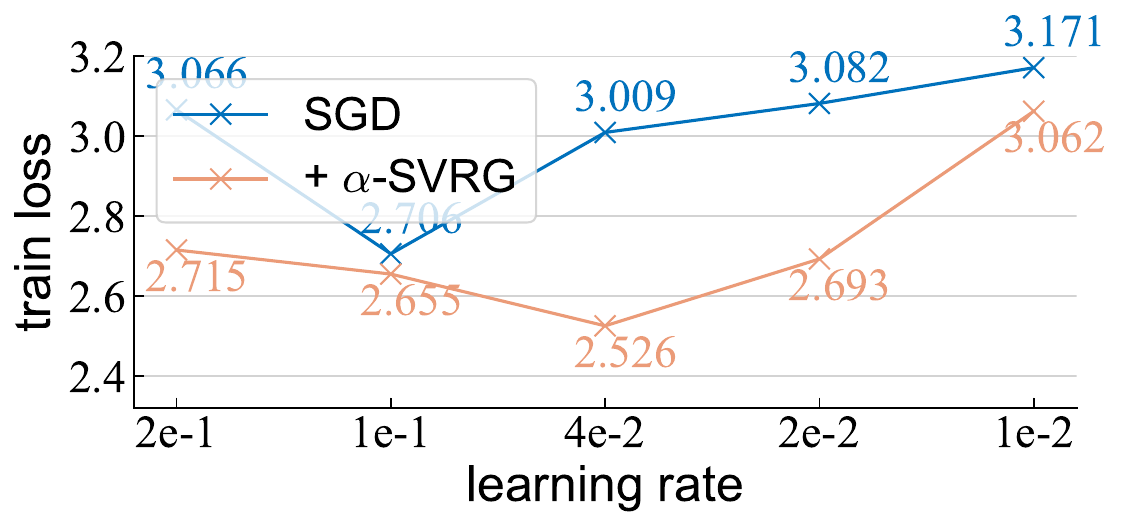} \\
    (a) CIFAR-100 & (b) Pets \\
    
    \includegraphics[width=.48\linewidth]{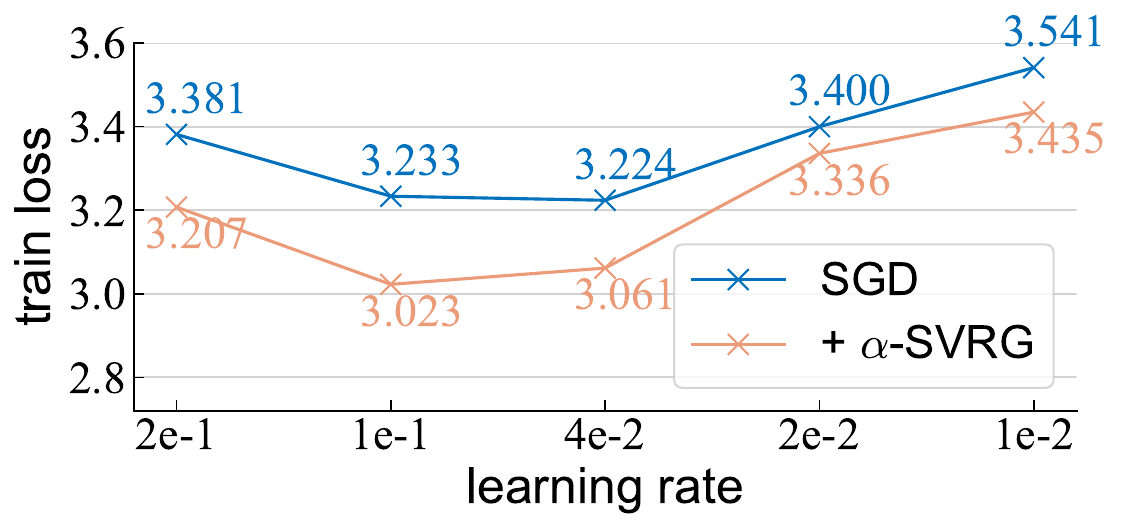} &
    \includegraphics[width=.48\linewidth]{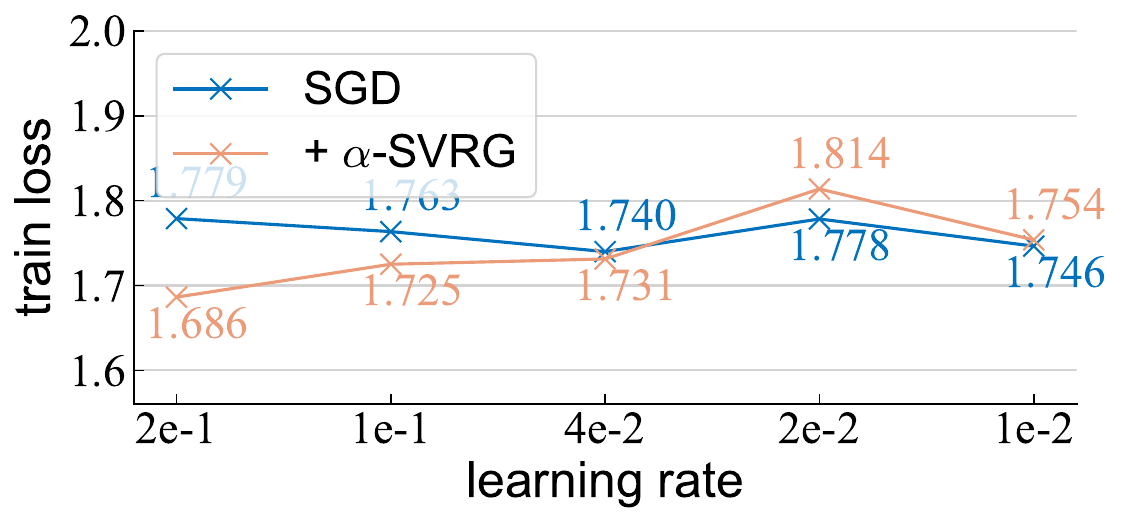} \\
    (c) Flowers & (d) STL-10 \\
    \includegraphics[width=.48\linewidth]{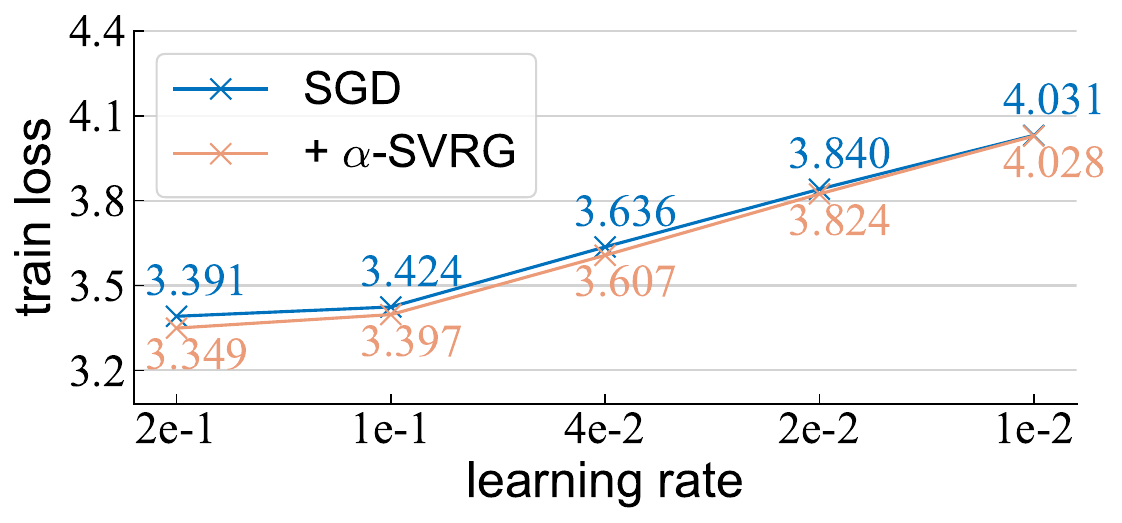} &
    \includegraphics[width=.48\linewidth]{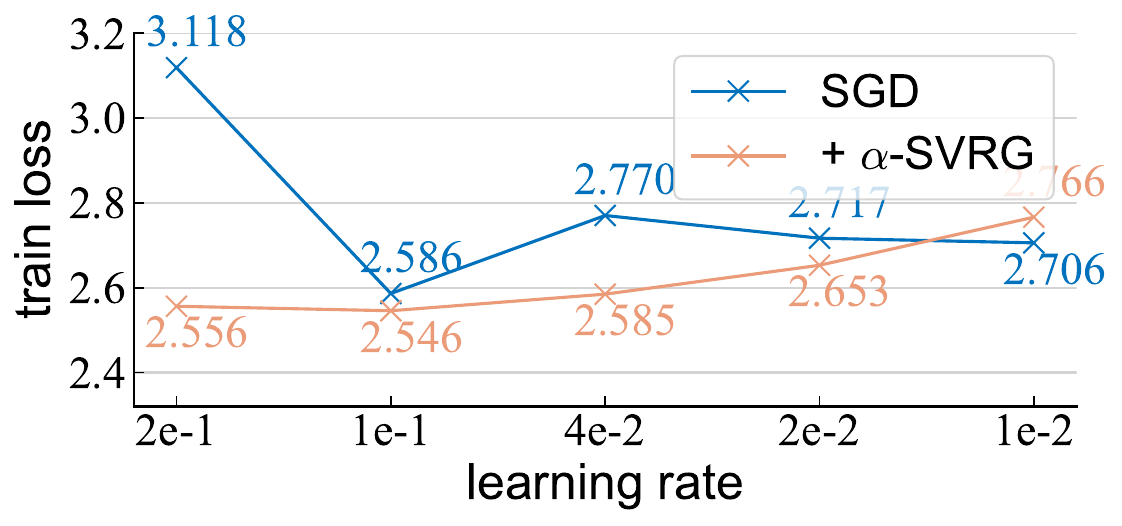} \\
    (e) Food-101 & (f) DTD \\
    \includegraphics[width=.48\linewidth]{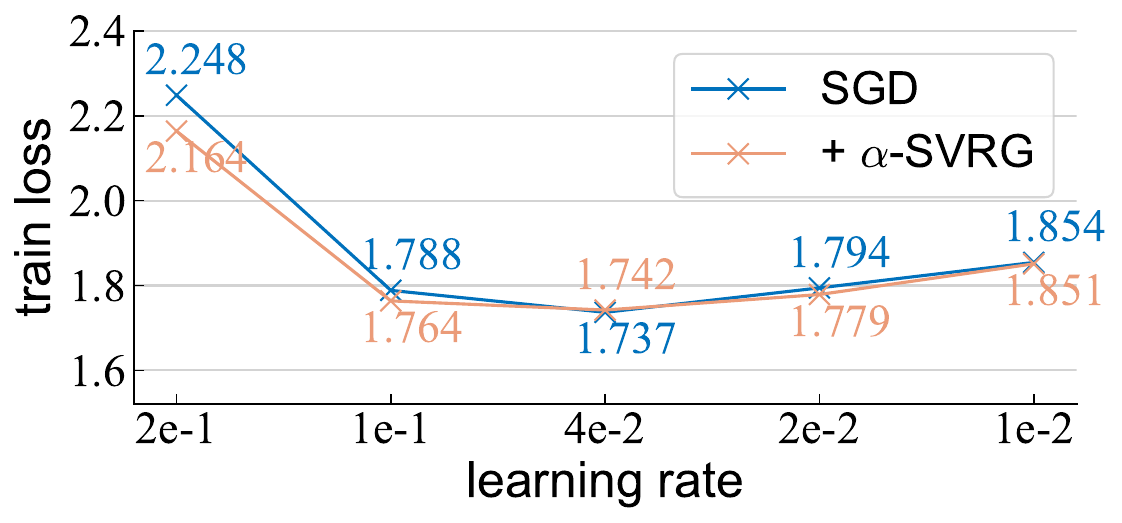} &
    \includegraphics[width=.48\linewidth]{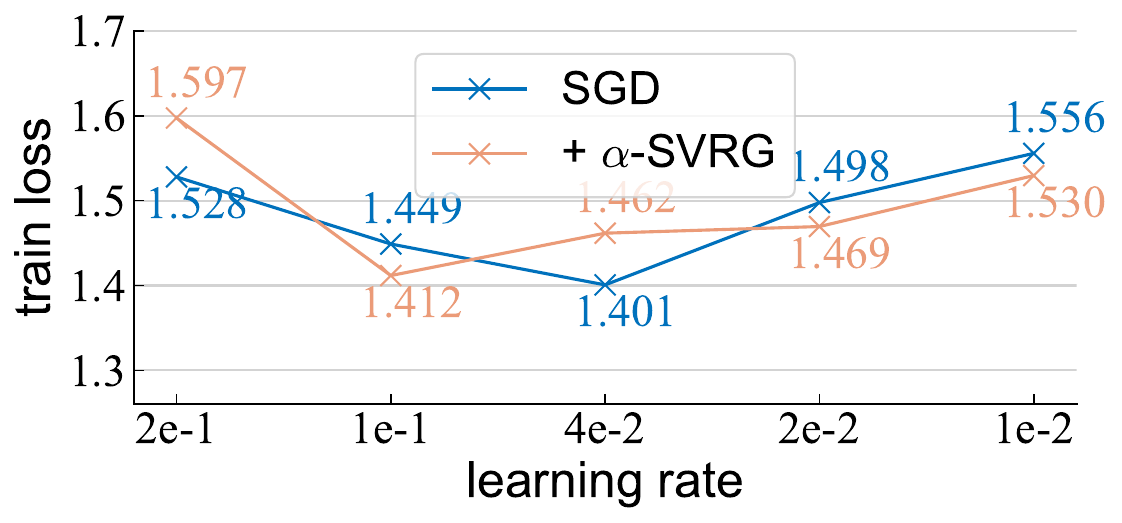} \\
    (g) SVHN & (h) EuroSAT 
    \end{tabular}
    
  \caption{\textbf{$\boldsymbol{\alpha}$-SVRG with SGD at different learning rates}.}
  \label{fig:sgd_learning_rate}
  \end{figure}

\clearpage
\newpage
\textbf{$\alpha$-SVRG with SGD.} We also sweep the base learning rate for the results of Table~\ref{tab:seed_2} using SGD as the base optimizer. We compare vanilla SGD to $\alpha$-SVRG ($\alpha_0=0.5$). The results are shown in Figure~\ref{fig:sgd_learning_rate}. In most learning rates, $\alpha$-SVRG can achieve a lower training loss than SGD.

\clearpage
\newpage

\subsection{Using $\alpha$-SVRG during the Initial Phases of Training}
\label{appendix:early_svrg}

Compared with SGD and AdamW, both standard SVRG and $\alpha$-SVRG require computing the snapshot stochastic gradient $\nabla f_i (\snapshot)$ and snapshot full gradient $\nabla f(\snapshot)$. This leads to about twice the computational cost of the baseline methods. Nevertheless, in Section~\ref{sec:approach}, we find that $\alpha$-SVRG can only reduce gradient variance during the initial training epochs and then maintains at a similar level as the baseline methods. This motivates us to apply $\alpha$-SVRG \emph{only at the early phases of training}, and we refer this approach as early $\alpha$-SVRG. 

To evaluate this approach, we use early $\alpha$-SVRG to train ConvNeXt-Femto on various image classification datasets. We use the default linear decay with an initial coefficient $\alpha_0=0.75$ to schedule early $\alpha$-SVRG, but $\alpha$-SVRG is only applied during the first 10\% of training and is disabled thereafter. Moreover, we add a transition epoch where the coefficient decreases from its original final value to zero. We find this crucial for maintaining the stability of momentum in the base optimizer.

\begin{table}[th]
\centering
\small
\def\arraystretch{1.3}
\begin{tabular}{lcccccccc}
& \multicolumn{2}{c}{CIFAR-100} & \multicolumn{2}{c}{Pets} & \multicolumn{2}{c}{Flowers} & \multicolumn{2}{c}{STL-10} \\
\Xhline{0.7pt}
AdamW & 2.659 & - & 2.203 & - & 2.400 & - & 1.641 & - \\
\rowcolor{gray} 
+\ \approach\ & 2.646 & \better{0.013} & 2.004 & \better{0.199} & 2.162 & \better{0.238} & 1.568 & \better{0.073} \\
\rowcolor{gray}
+\ early \approach\ & 2.644 & \better{0.015} & 2.190 & \better{0.013} & 2.328 & \better{0.072} & 1.616 & \better{0.025} \\
\end{tabular}

\centering
\small
\begin{tabular}{lcccccccccc}
& \multicolumn{2}{c}{Food-101} & \multicolumn{2}{c}{DTD} & \multicolumn{2}{c}{SVHN} & \multicolumn{2}{c}{EuroSAT} \\
\Xhline{0.7pt}
AdamW & 2.451 & - & 1.980 & - & 1.588 & - & 1.247 & - \\
\rowcolor{gray}
+\ \approach\ & 2.461 & \worse{0.010} & 1.829 & \better{0.151} & 1.573 & \better{0.015} & 1.237 & \better{0.010} \\
\rowcolor{gray}
+\ early \approach\ & 2.444 & \better{0.007} & 1.918 & \better{0.062} & 1.583 & \better{0.005} & 1.240 & \better{0.007} \\
\end{tabular}
\caption*{(a) \textbf{training loss}}
\vspace{.5em}
\small
\begin{tabular}{lcccccccc}
& \multicolumn{2}{c}{CIFAR-100} & \multicolumn{2}{c}{Pets} & \multicolumn{2}{c}{Flowers} & \multicolumn{2}{c}{STL-10} \\
\Xhline{0.7pt}
baseline & 81.0 & - & 72.8 & - & 80.8 & - & 82.3 & - \\
\rowcolor{gray}
+\ \approach\ & 80.6 & \worseinv{0.4} & 76.7 & \betterinv{3.9} & 82.6 & \betterinv{1.8}  & 84.0 & \betterinv{1.7} \\
\rowcolor{gray}
+\ early \approach\ & 81.0 & \betterinv{0.0} & 74.6 & \betterinv{1.8} & 83.6 & \betterinv{2.8} & 82.8 & \betterinv{0.5}\\
\end{tabular}

\centering
\small
\begin{tabular}{lcccccccccc}
& \multicolumn{2}{c}{Food-101} & \multicolumn{2}{c}{DTD} & \multicolumn{2}{c}{SVHN} & \multicolumn{2}{c}{Euro} \\
\Xhline{0.7pt}
baseline & 85.9 & - & 57.9 & - & 94.9 & - & 98.1 & - \\
\rowcolor{gray}
+\ \approach\ & 85.0 & \worseinv{0.9} & 60.3 & \betterinv{2.4} & 95.7 & \betterinv{0.8} & 98.2 & \betterinv{0.1} \\
\rowcolor{gray}
+\ early \approach\ & 85.9 & \betterinv{0.0} & 62.3 & \betterinv{4.4} & 95.8 & \betterinv{0.9} & 98.0 & \worseinv{0.1}\\
\end{tabular}
\caption*{(b) \textbf{validation accuracy}}
\caption{\textbf{Early $\boldsymbol{\alpha}$-SVRG on smaller classification datasets.} }
\label{tab:early}
\end{table}

Figure~\ref{tab:early} shows the results. We can see early $\alpha$-SVRG consistently reduces training loss across different datasets. Furthermore, we observe that the validation accuracy achieved by early $\alpha$-SVRG is sometimes higher than that of standard $\alpha$-SVRG. This phenomenon is likely because disabling $\alpha$-SVRG in the later training stages allows the presence of benign gradient noise during optimization. Such noise may drive the model toward local minima that exhibit better generalization properties~\citep{smith2020generalization,damian2021label,li2022happenssgdreacheszero}.

\clearpage
\newpage

\section{Additional Results of Optimal Coefficient}
We provide further results on optimal coefficient in SVRG (Equation~\ref{eq:optimal}) below: gradient variance reduction on other datasets with SVRG using optimal coefficient (Appendix~\ref{appendix:other_dataset}), the impact of data on optimal coefficient (Appendix~\ref{appendix:data_optimal_coefficient}), and the evolution of the correlation between model gradients and snapshot gradients during training (Appendix~\ref{appendix:corr}).
\subsection{SVRG with Optimal Coefficient on Other Datasets}
\label{appendix:other_dataset}
\begin{figure}[h]
\centering
\begin{minipage}{0.48\textwidth}
    \centering
    \includegraphics[width=\linewidth,keepaspectratio]{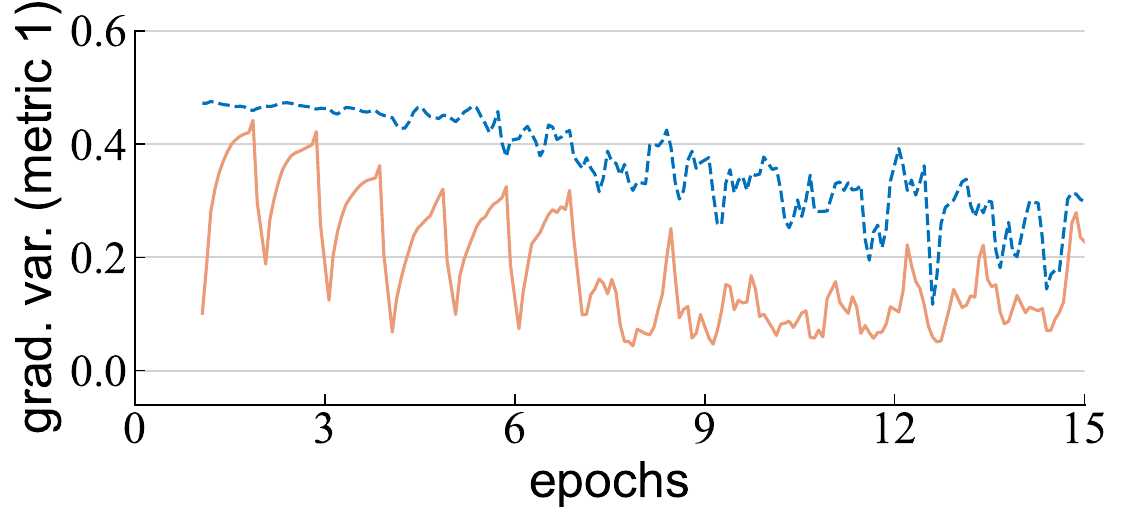}
\end{minipage}%
\begin{minipage}{0.48\textwidth}
    \centering
    \includegraphics[width=\linewidth,keepaspectratio]{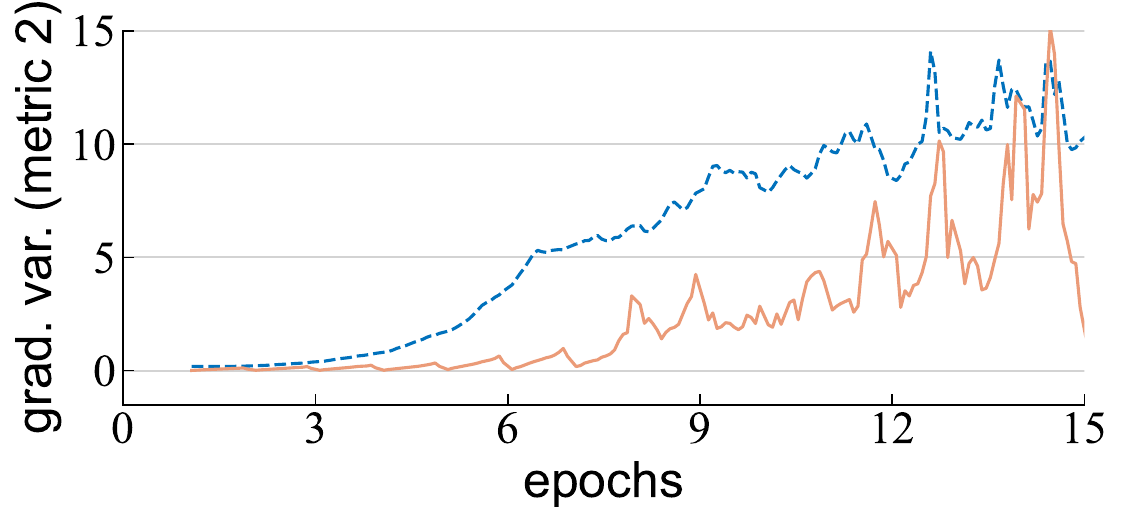}
\end{minipage}
\begin{minipage}{0.48\textwidth}
    \centering
    \includegraphics[width=\linewidth,keepaspectratio]{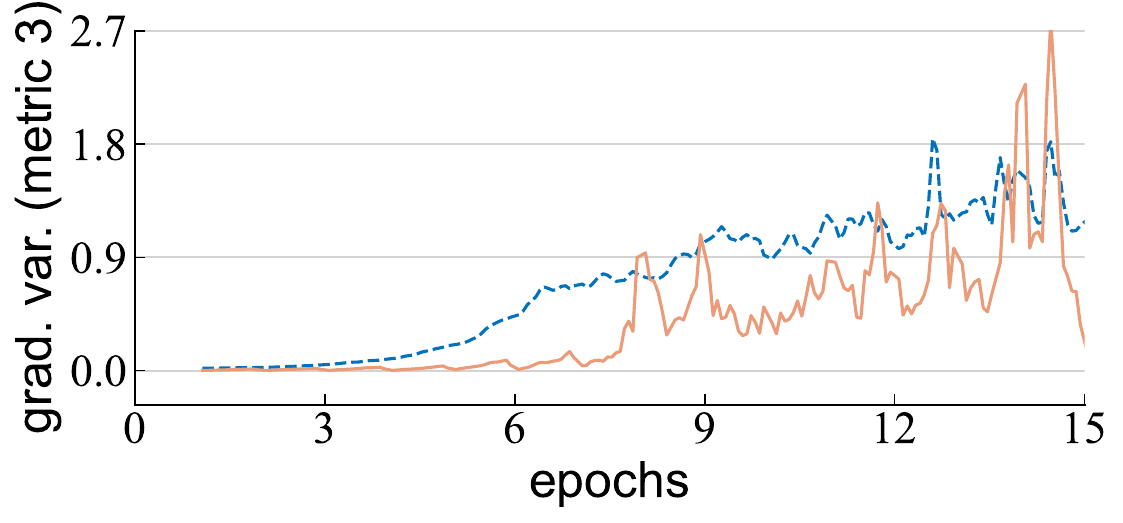}
\end{minipage}%
\begin{minipage}{0.48\textwidth}
    \centering
    \includegraphics[width=\linewidth,keepaspectratio]{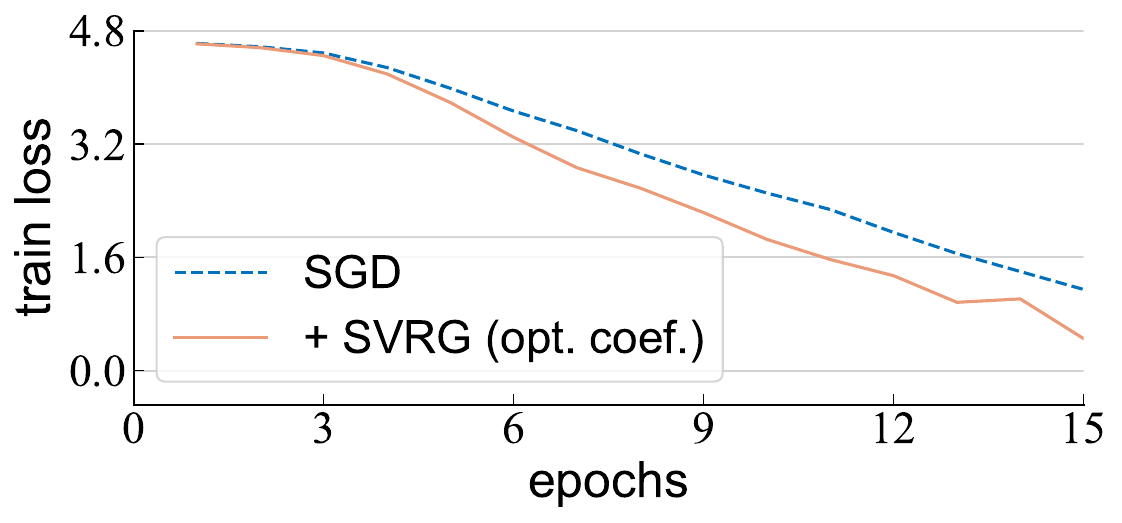}
\end{minipage}
\vspace*{-1em}
\caption*{(a) Flowers}
\vspace{1em}
\begin{minipage}{0.48\textwidth}
    \centering
    \includegraphics[width=\linewidth,keepaspectratio]{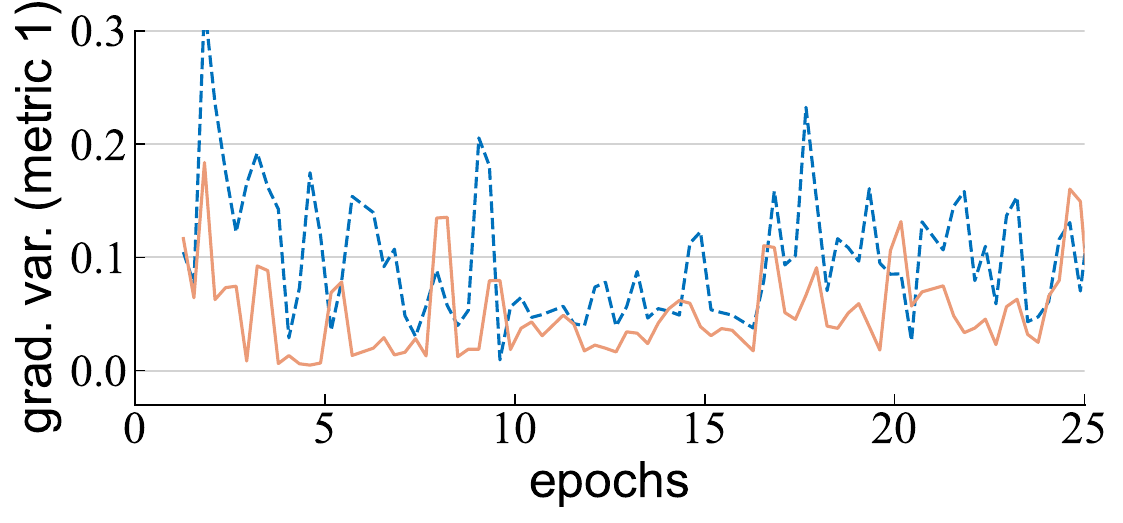}
\end{minipage}%
\begin{minipage}{0.48\textwidth}
    \centering
    \includegraphics[width=\linewidth,keepaspectratio]{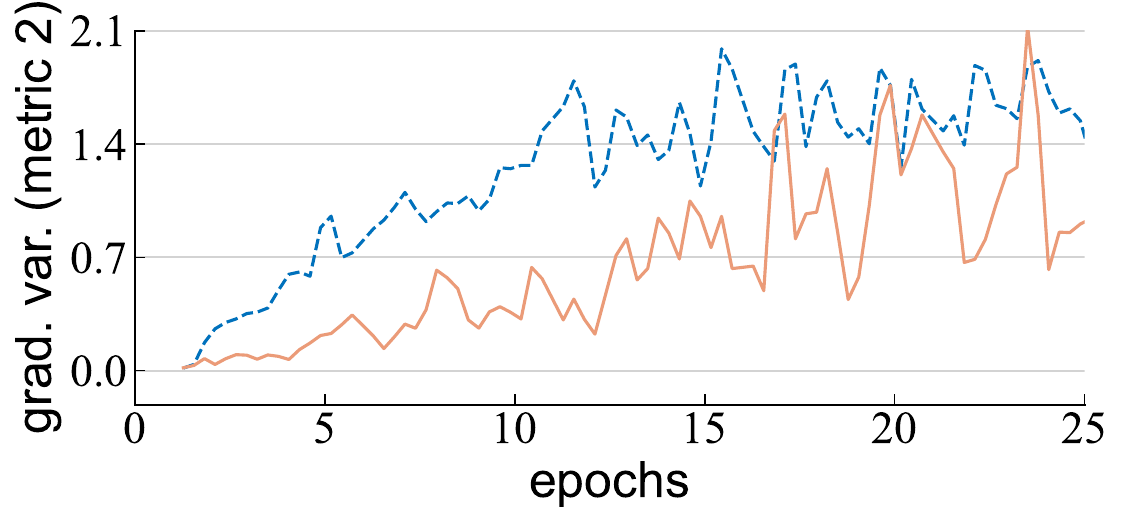}
\end{minipage}
\begin{minipage}{0.48\textwidth}
    \centering
    \includegraphics[width=\linewidth,keepaspectratio]{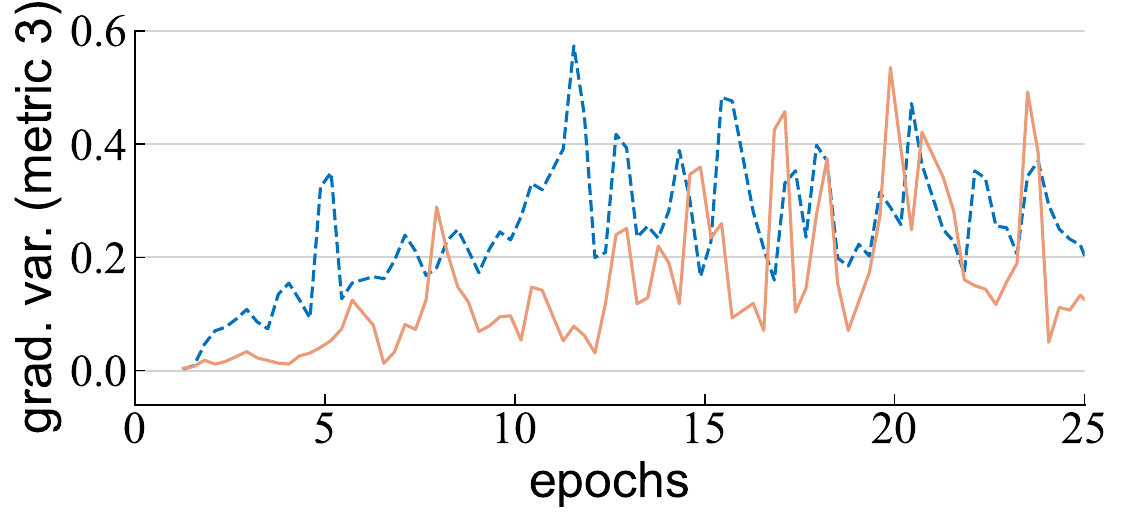}
\end{minipage}%
\begin{minipage}{0.48\textwidth}
    \centering
    \includegraphics[width=\linewidth,keepaspectratio]{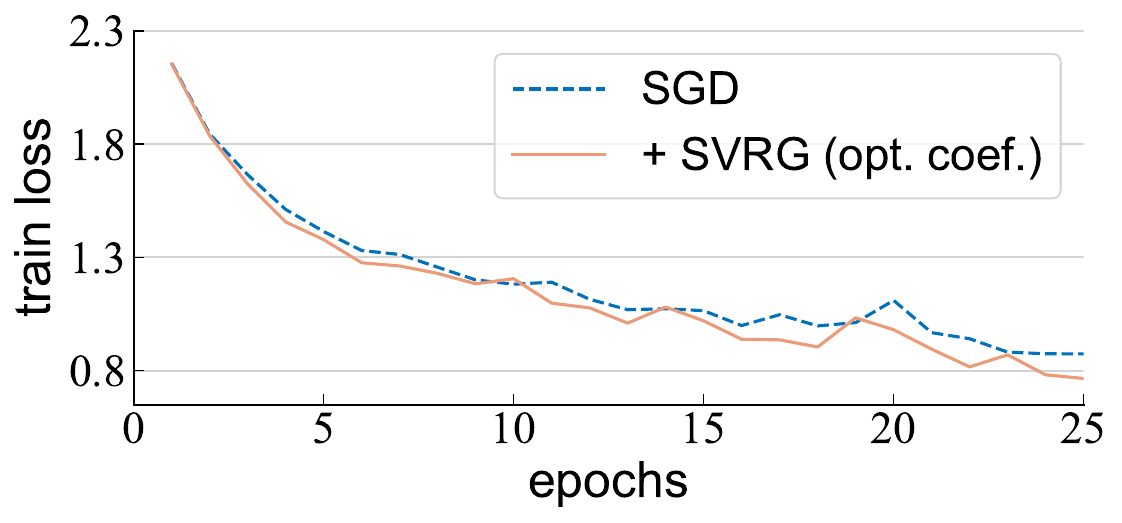}
\end{minipage}
\vspace*{-1em}
\caption*{(b) EuroSAT}
\caption{\textbf{SVRG with optimal coefficient on other datasets.}}
\label{fig:optimal_svrg_others}
\end{figure}

Throughout the experiments in Section~\ref{sec:3}, we mainly study the behavior of SVRG with optimal coefficient on CIFAR-10 dataset. To show the generality of our approach for SVRG, we below monitor the optimal coefficient on other image classification datasets. We train a MLP-4 on Flowers / EuroSAT with SGD and SVRG using optimal coefficient for 15 / 25 epochs. As shown in Figure~\ref{fig:optimal_svrg_others}, SVRG with optimal coefficient can consistently reduce gradient variance and achieve a lower training loss than the baseline SGD.  

\subsection{Effect of Data on Optimal Coefficient}
\label{appendix:data_optimal_coefficient}
\begin{figure}[h]
    \centering
    \begin{minipage}{0.49\linewidth}
        \centering
        \includegraphics[width=\linewidth,keepaspectratio]{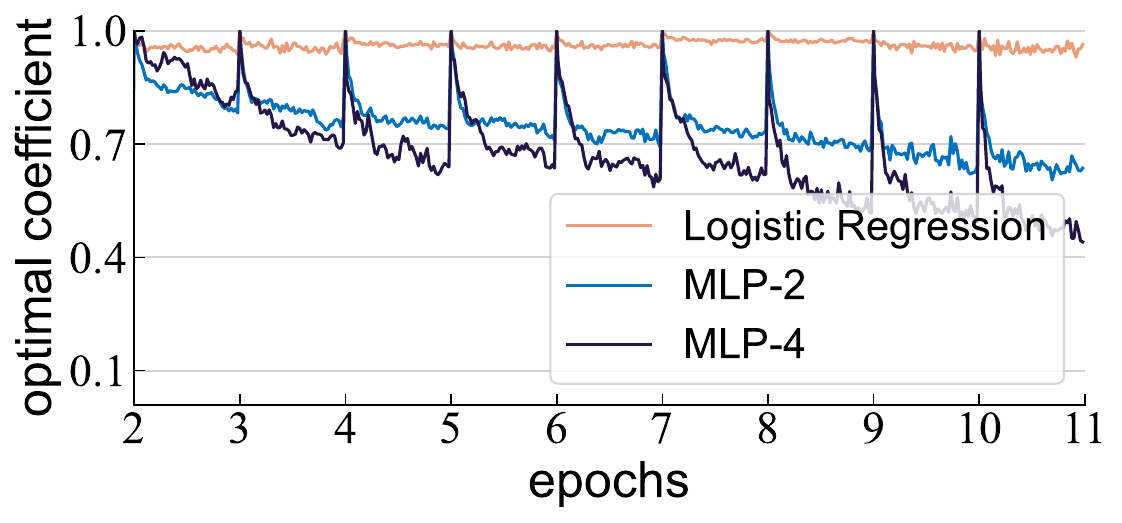}
        \vspace*{-0.7cm}
        \centering
        \caption*{\hspace{0.05\linewidth} (a) CIFAR-10}
    \end{minipage}~
    \begin{minipage}{0.49\linewidth}
        \centering
        \includegraphics[width=\linewidth,keepaspectratio]{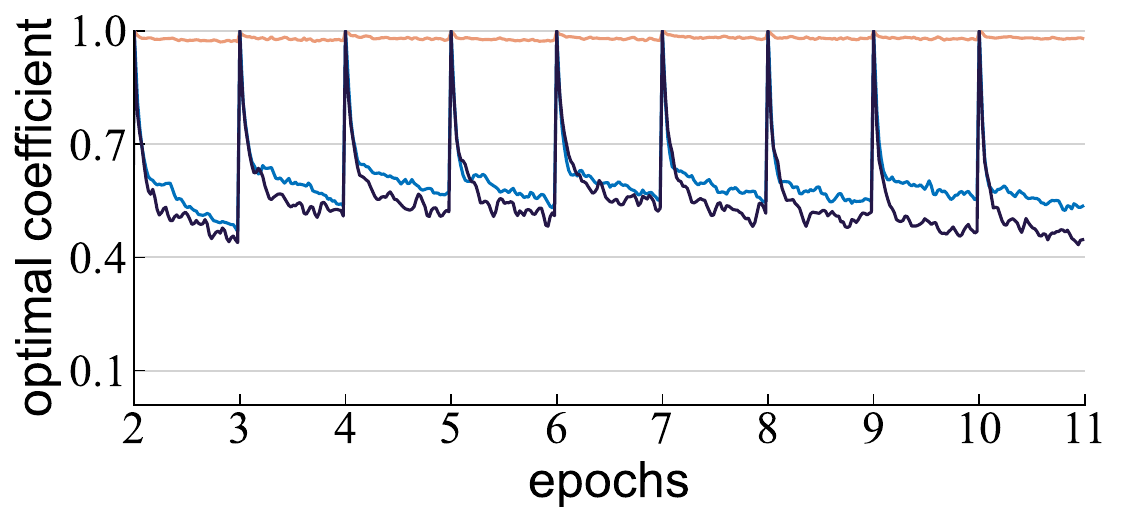}
        \vspace*{-0.7cm}
        \centering
        \caption*{\hspace{0.05\linewidth} (b) CIFAR-100}
    \end{minipage}
    \vspace*{-0.1cm}
    \caption{\textbf{Effects of Data on Optimal Coefficient.}}
    \vspace*{-0.1cm}
    \label{fig:optimal_data}
\end{figure}

Below we conduct experiments on CIFAR-10 and CIFAR-100 to understand how the number of object classes in datasets affect optimal coefficient. Specifically, we train 1, 2, and 4-layer MLPs (Logistic Regression, MLP-2, and MLP-4) on each of the two datasets using SGD (without SVRG) and compute the optimal coefficient (Equation~\ref{eq:optimal}) during the training. Figure~\ref{fig:optimal_data} shows the results. 

The optimal coefficient of each model trained on CIFAR-100 is lower than that on CIFAR-10. This is possibly because CIFAR-100 has 10$\times$ more object classes than CIFAR-10 and therefore the correlation between the snapshot model gradient and the current model gradient is weaker. Thus, we recommend using a smaller coefficient when the training dataset includes a larger number of classes.

\subsection{Correlation between Model Gradients and Snapshot Gradients}
\label{appendix:corr}
\begin{figure}[h]
    \centering
    \begin{minipage}{0.49\linewidth}
        \centering
        \includegraphics[width=\linewidth,keepaspectratio]{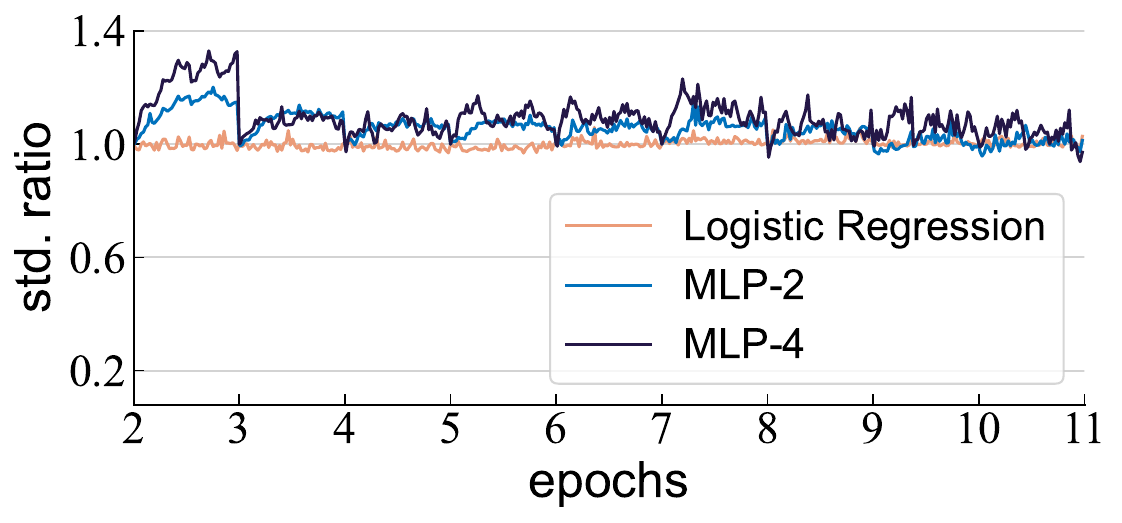}
        \vspace*{-0.7cm}
        \centering
        \caption*{\hspace{0.05\linewidth} (a) SGD}
    \end{minipage}~
    \begin{minipage}{0.49\linewidth}
        \centering
        \includegraphics[width=\linewidth,keepaspectratio]{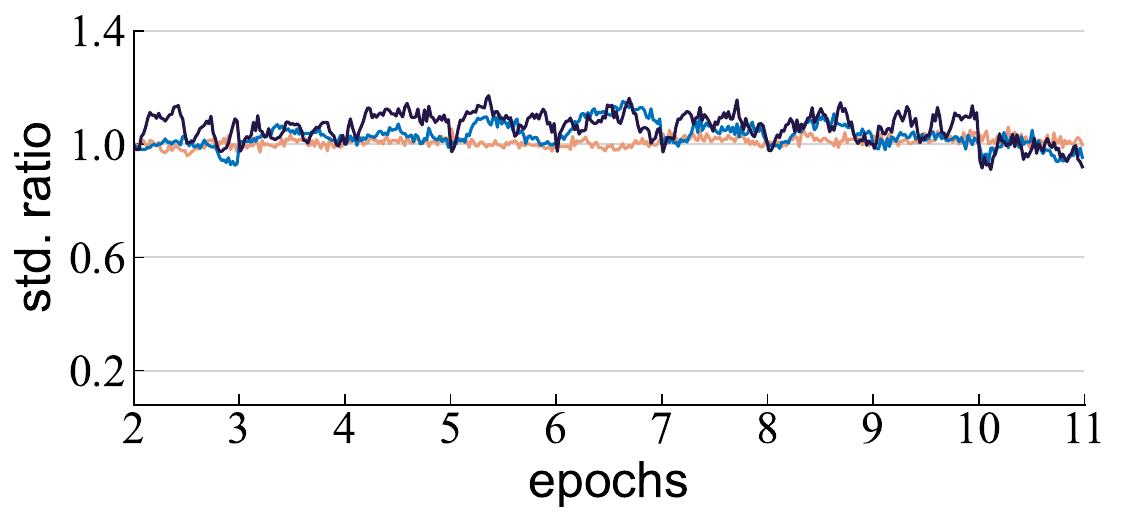}
        \vspace*{-0.7cm}
        \centering
        \caption*{\hspace{0.05\linewidth} (b) AdamW}
    \end{minipage}
    \vspace*{-0.1cm}
    \caption{\textbf{Standard deviation ratio.} The ratio between the standard deviations of the model gradients and the snapshot gradients oscillates around 1 but is relatively stable overall.}
    \label{fig:std}
    \end{figure}

\begin{figure}[h]
    \begin{minipage}{0.49\linewidth}
        \centering
        \includegraphics[width=\linewidth,keepaspectratio]{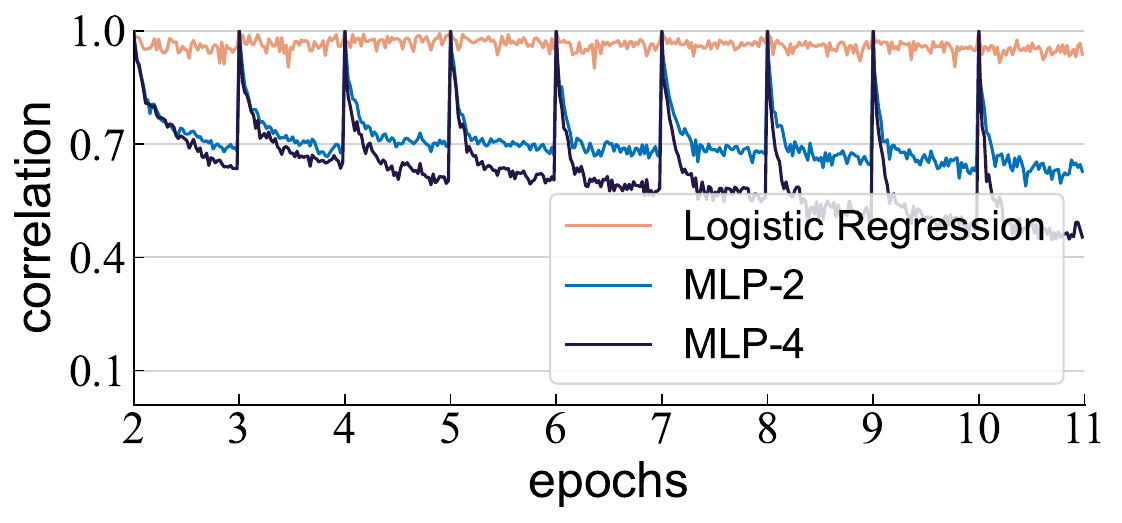}
        \vspace*{-0.7cm}
        \centering
        \caption*{\hspace{0.05\linewidth} (a) SGD}
    \end{minipage}~
    \begin{minipage}{0.49\linewidth}
        \centering
        \includegraphics[width=\linewidth,keepaspectratio]{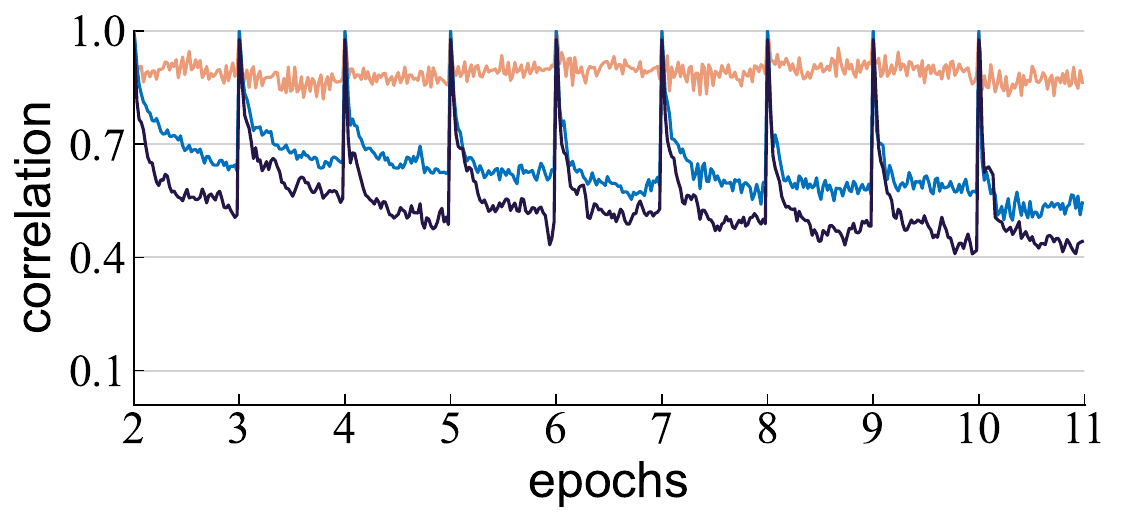}
        \vspace*{-0.7cm}
        \centering
        \caption*{\hspace{0.05\linewidth} (b) AdamW}
    \end{minipage}
    \caption{\textbf{Correlation.} The correlation between the snapshot gradients and the model gradients decreases as the training progresses.}
    \label{fig:corr}
\end{figure}

In Section~\ref{sec:3}, we find each epoch's average optimal coefficient decreases as training progresses. Here, we seek to understand whether the standard deviation ratio $\sigma(\nabla f_{\cdot, k}(\vtheta^t))/ \sigma(\nabla f_{\cdot, k}(\snapshot))$ or the correlation $\rho\left(\nabla f_{\cdot, k}(\snapshot), \nabla f_{\cdot, k}(\vtheta^t)\right)$ in Equation~\ref{eq:optimal} contributes more to this observation. We plot these two values separately in Figures~\ref{fig:std} and~\ref{fig:corr}. We can see the standard deviation ratio is relatively stable during the training, whereas the correlation decreases very much during the training. 

\section{Alternative Schedules for \approach}
\label{appendix:schedule}
In this part, we provide the exact mathematical formulation of each schedule studied in Section~\ref{sec:analysis}.

\textbf{Notations.} We decompose the global iteration index \(t\) into an epoch-wise index \(s\) and an iteration index \(i\) within an epoch. We also denote the total number of training epochs as \(T\) and represent the number of iterations in one epoch as \(M\). 

\textbf{Linear schedule.} This is our default scheduling approach. The coefficient decreases linearly \emph{across epochs and keeps as a constant within an epoch}:
\begin{gather}
    \alpha_\text{linear}^t = \alpha_0(1-\frac{s(t)}{T}),
\end{gather}
\textbf{Other global schedules.} We also consider quadratic decay and geometric decay:
\begin{gather}
    \alpha_\text{quadratic}^t = \frac{\alpha_0}{T^2}(T-s(t))^2,\\
    \alpha_\text{geometric}^t = \alpha_0(\frac{\alpha_{\text{final}}\footnotemark}{\alpha_0})^{\frac{s(t)}{T}},\label{eq:geometric}
\end{gather}%
\footnotetext{We set $\alpha_{\text{final}}=0.01$ to ensure that the geometric schedule eventually decreases to a sufficiently small value. Note that $\alpha_{\text{final}}$ can not be zero, as it serves as the base of the exponent. The same rule applies to Equation~\ref{eq:double_geometric}.}

\textbf{Double schedules.}
In Figure~\ref{fig:depth} of Section~\ref{sec:3}, within each epoch, the coefficient starts from 1 and decreases over time. Motivated by this local behavior, we introduce three additional schedules that combine both the local and global decay: d(ouble)-linear, d-quadratic, and d-geometric. In addition to the global decay, each double schedule has another local decay for each epoch that initiates at 1 and decreases to an ending value specified by the global decay.
\begin{gather}
    \alpha_\text{d-linear}^t = \alpha_\text{linear}^t \underbrace{(1-\frac{i}{M})}_\text{local decay} + \alpha_\text{linear}^t \\
    \hspace*{-0.1cm}\alpha_\text{d-quadratic}^t =  (1-\alpha_\text{quadratic}^t) \underbrace{\frac{1}{M^2}(M-i)^2}_\text{local decay}+\alpha_\text{quadratic}^t \\
    \alpha_\text{d-geometric}^t = \underbrace{(\alpha_\text{geometric}^t+\alpha_{\text{d-final}})^{\frac{i}{M}}}_\text{local decay}\label{eq:double_geometric}
\end{gather}

\section{Convergence Comparison}
\label{appendix:vis}

\begin{figure}[h]
\centering
\begin{subfigure}[h]{.33\linewidth}
    \centering
    \includegraphics[width=\linewidth]{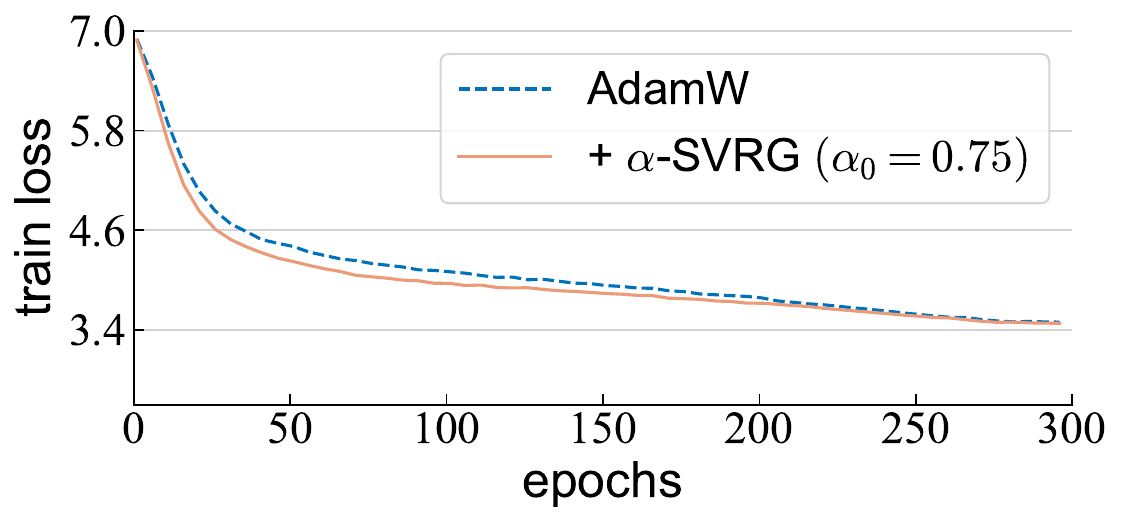}
    \vspace{-1.5em}
    \caption{ConvNeXt-F}
\end{subfigure}~
\begin{subfigure}[h]{.33\linewidth}
    \centering
    \includegraphics[width=\linewidth]{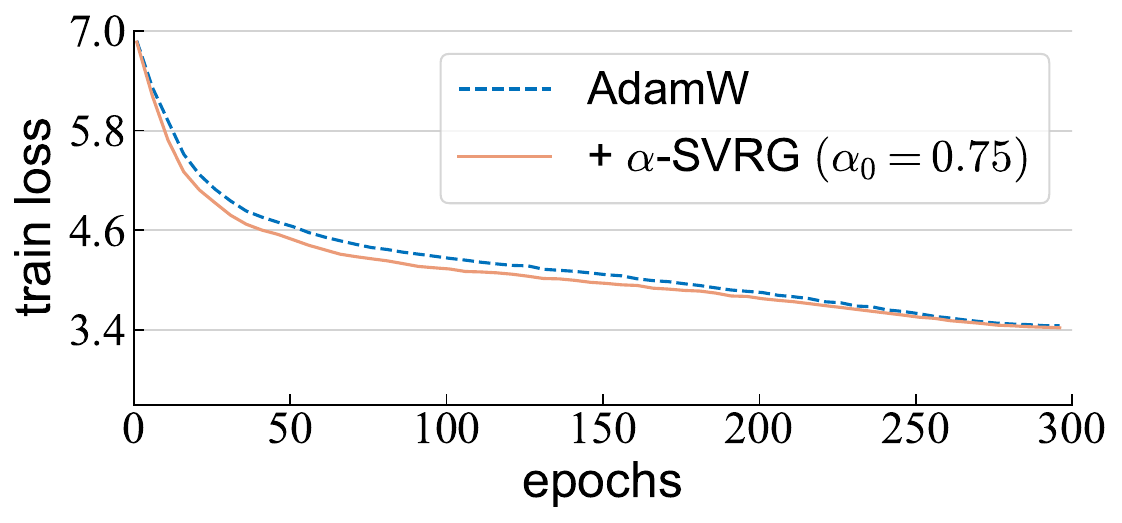}
    \vspace{-1.5em}
    \caption{ViT-T}
\end{subfigure}~
\begin{subfigure}[h]{.33\linewidth}
    \centering
    \includegraphics[width=\linewidth]{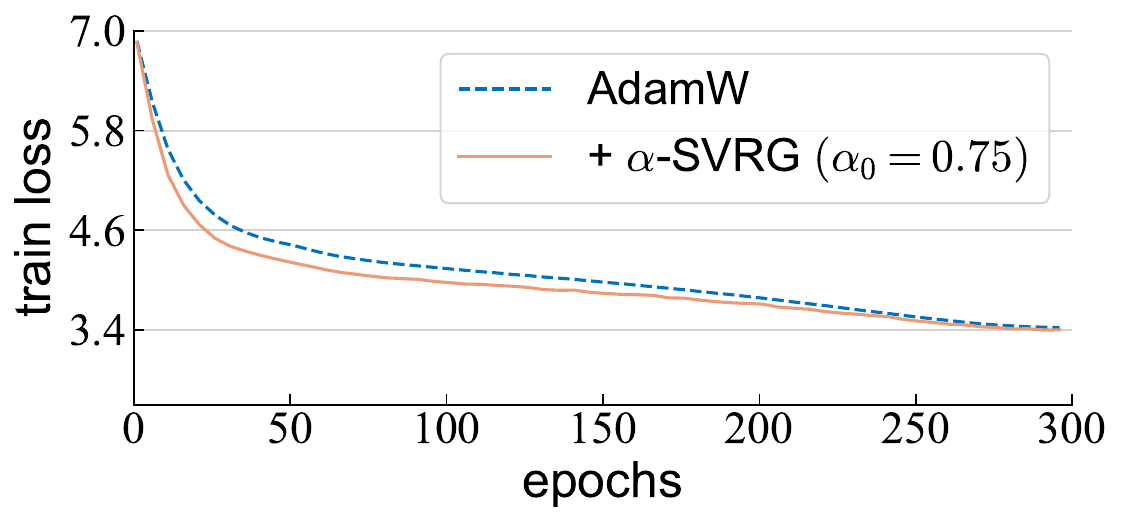}
    \vspace{-1.5em}
    \caption{Swin-F}
\end{subfigure}
\begin{subfigure}[h]{.33\linewidth}
    \centering
    \includegraphics[width=\linewidth]{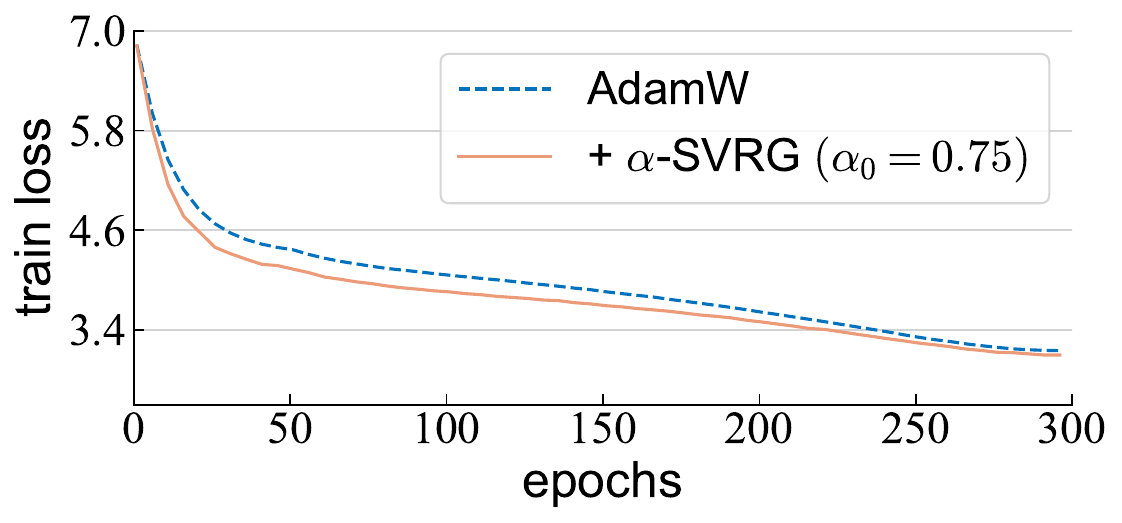}
    \vspace{-1.5em}
    \caption{Mixer-S}
\end{subfigure}~
\begin{subfigure}[h]{.33\linewidth}
    \centering
    \includegraphics[width=\linewidth]{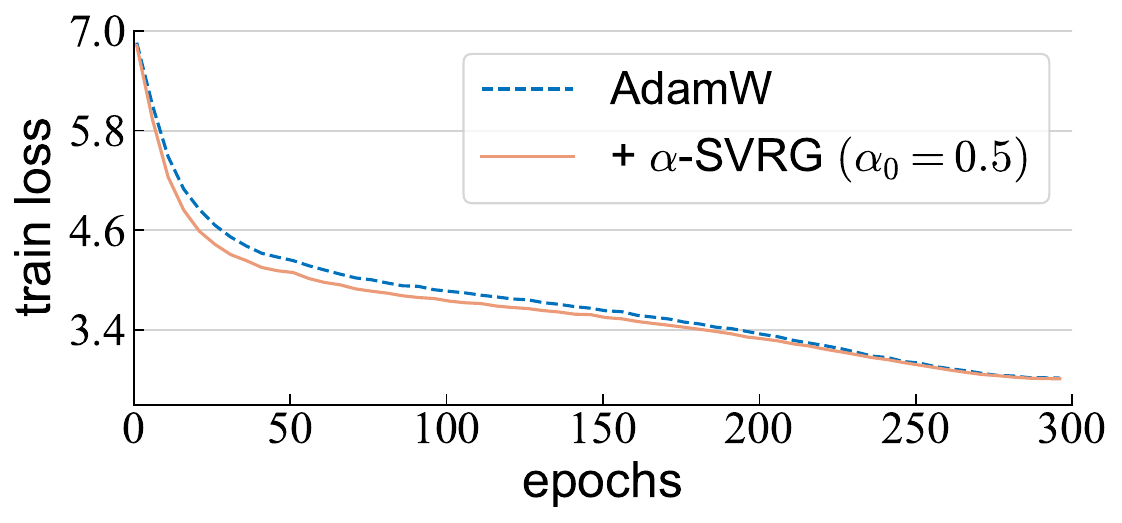}
    \vspace{-1.5em}
    \caption{ViT-B}
\end{subfigure}~
\begin{subfigure}[h]{.33\linewidth}
    \centering
    \includegraphics[width=\linewidth]{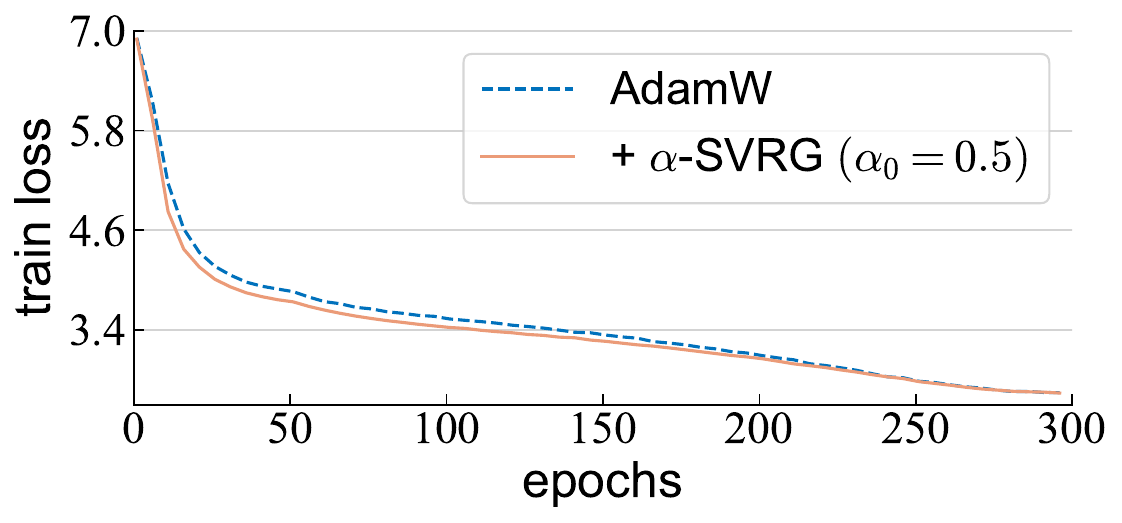}
    \vspace{-1.5em}
    \caption{ConvNeXt-B}
\end{subfigure}
\caption{\textbf{Training loss with AdamW and {$\boldsymbol{\alpha}$-SVRG} on ImageNet-1K (Table~\ref{tab:pre-trained1}).}}
\label{fig:convergence2}
\end{figure}

\begin{figure}[h]
\centering
\begin{subfigure}[h]{\linewidth}
    \centering
    \includegraphics[width=.33\linewidth]{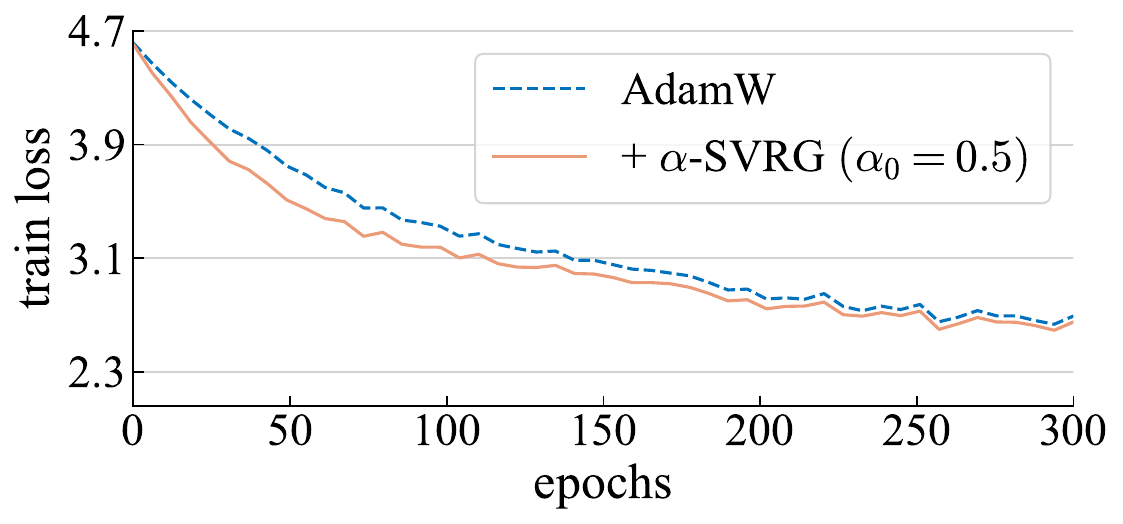}~
    \includegraphics[width=.33\linewidth]{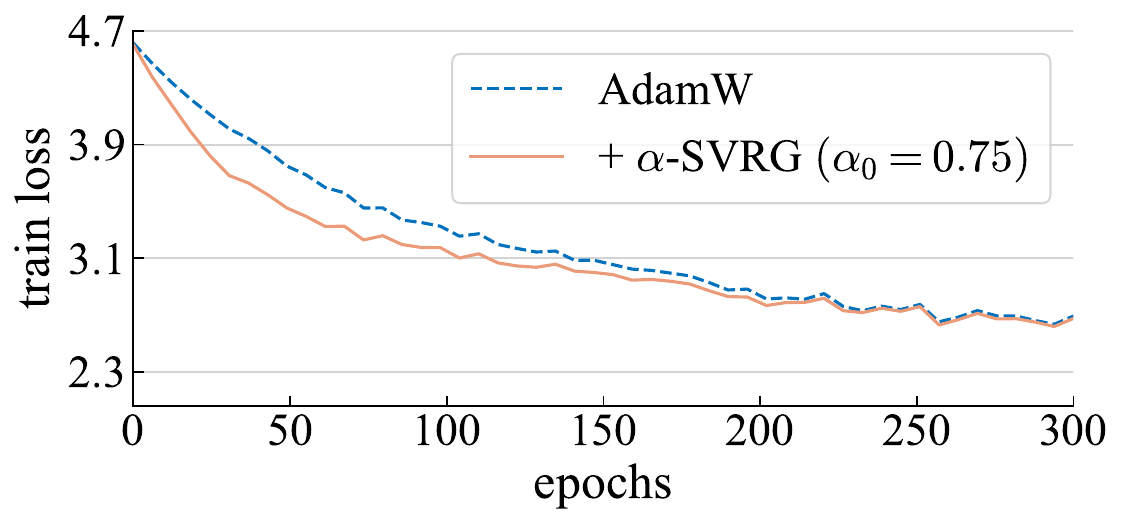}~
    \includegraphics[width=.33\linewidth]{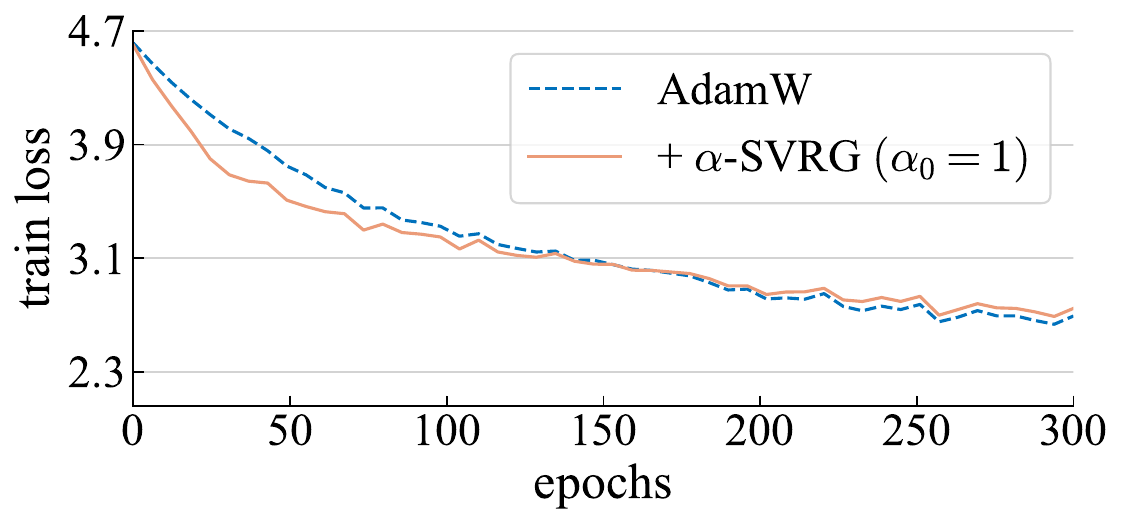}
    \vspace{-1.5em}
    \caption{CIFAR-100}
\end{subfigure}
\begin{subfigure}[h]{\linewidth}
    \centering
    \includegraphics[width=.33\linewidth]{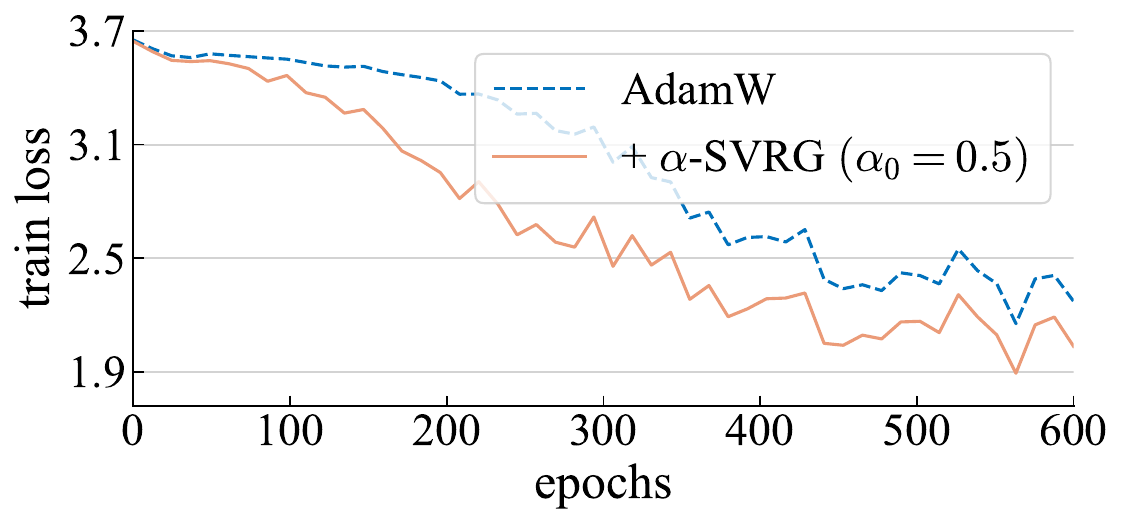}~
    \includegraphics[width=.33\linewidth]{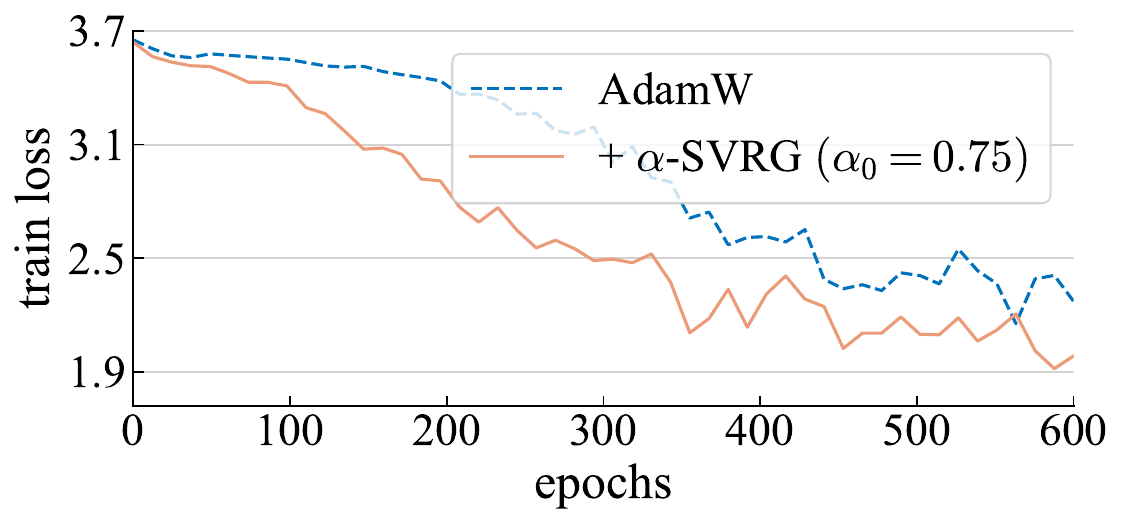}~
    \includegraphics[width=.33\linewidth]{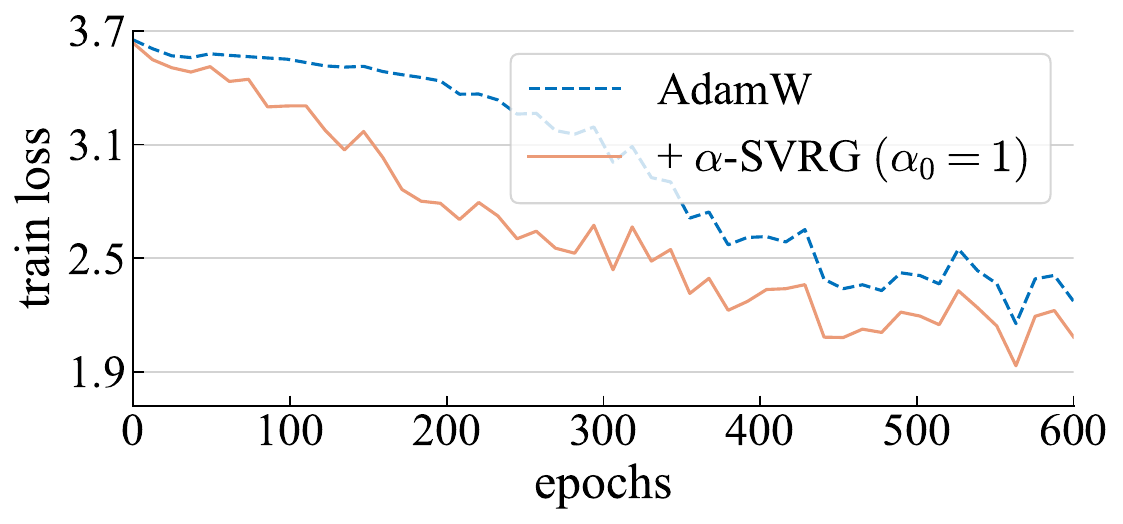}
    \vspace{-1.5em}
    \caption{Pets}
\end{subfigure}
\begin{subfigure}[h]{\linewidth}
    \centering
    \includegraphics[width=.33\linewidth]{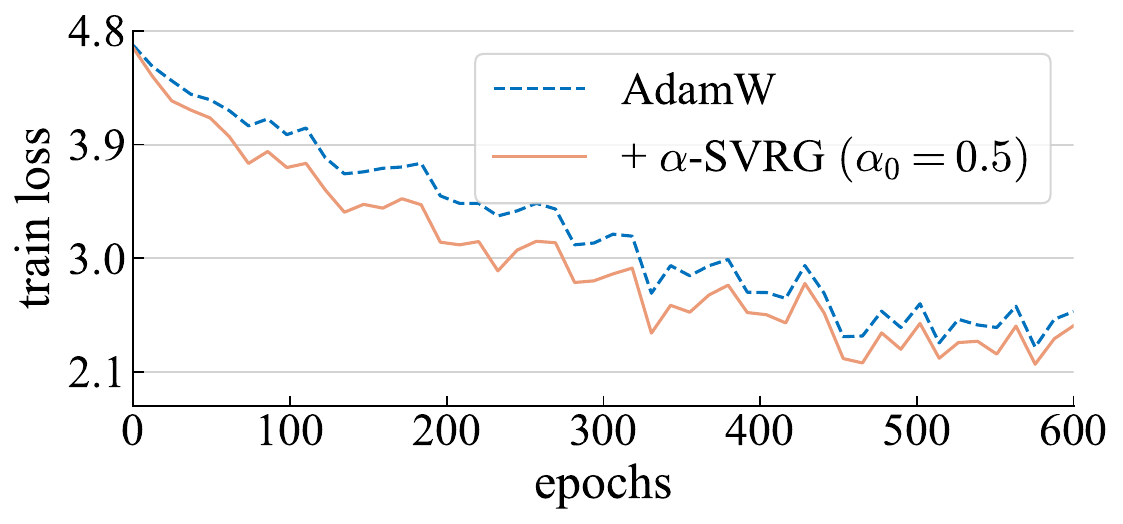}~
    \includegraphics[width=.33\linewidth]{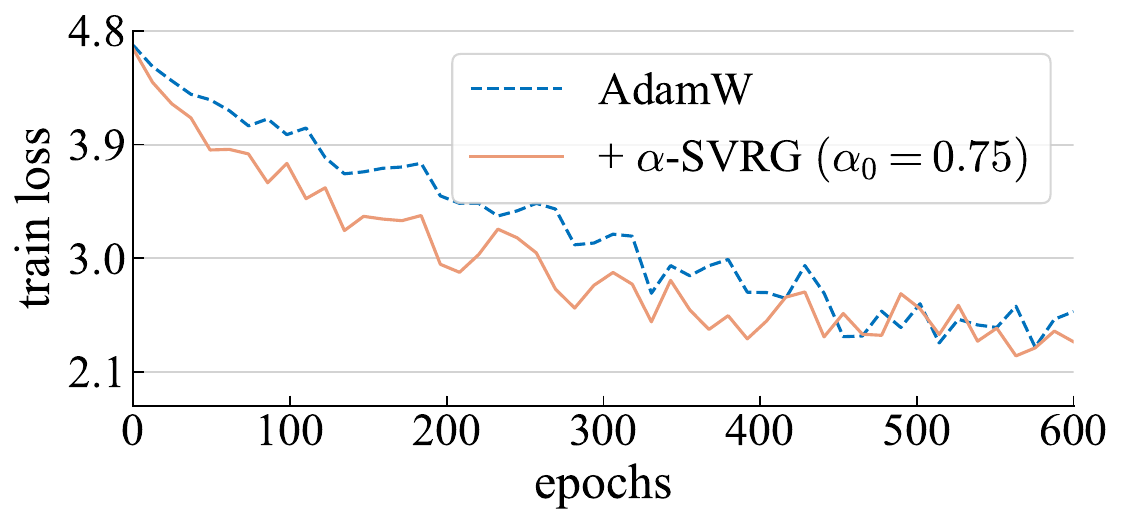}~
    \includegraphics[width=.33\linewidth]{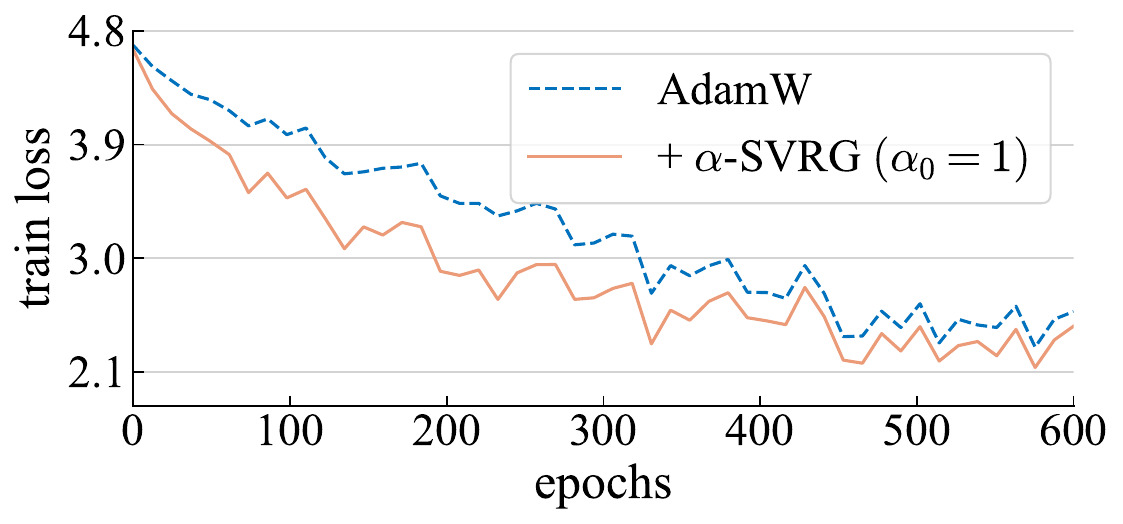} 
    \vspace{-1.5em}
    \caption{Flowers}
\end{subfigure}
\begin{subfigure}[h]{\linewidth}
    \centering
    \includegraphics[width=.33\linewidth]{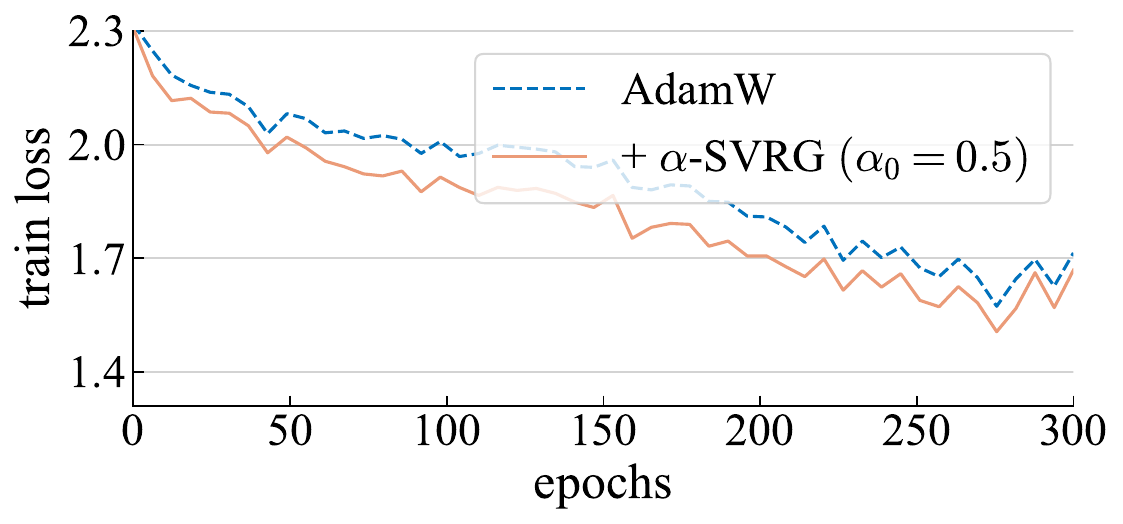}~
    \includegraphics[width=.33\linewidth]{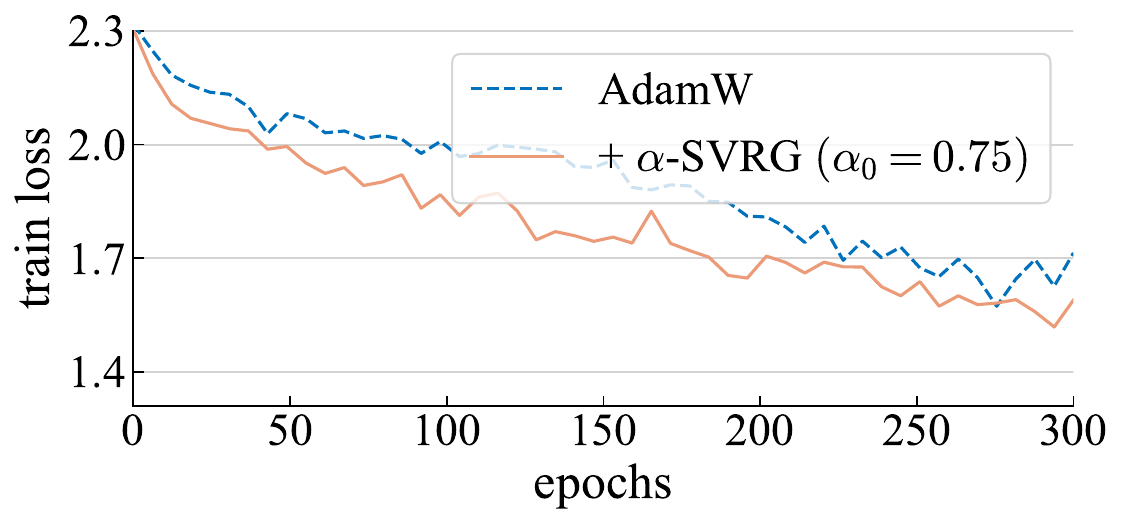}~
    \includegraphics[width=.33\linewidth]{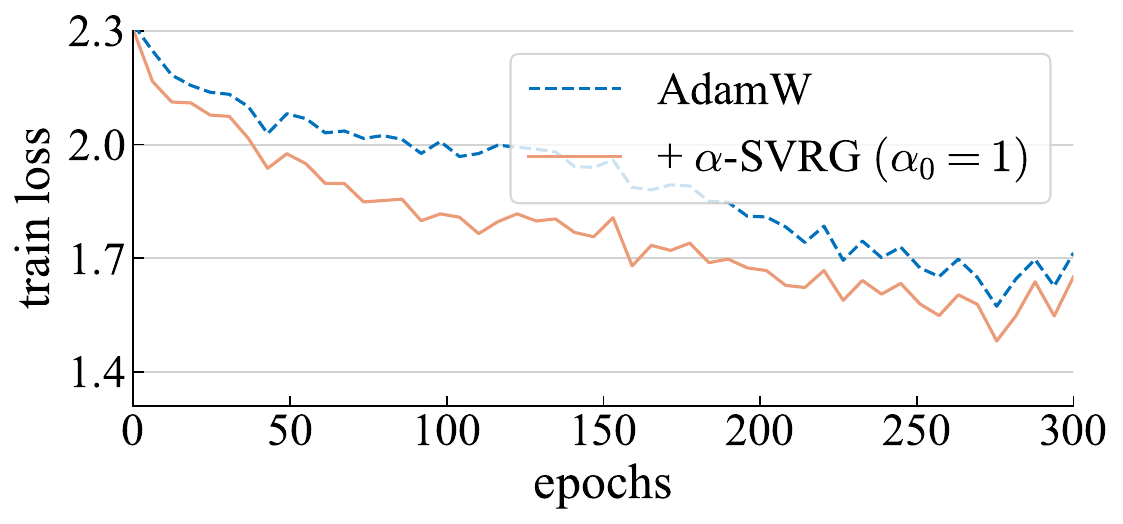}
    \vspace{-1.5em}
    \caption{STL-10}
\end{subfigure}
\begin{subfigure}[h]{\linewidth}
    \centering
    \includegraphics[width=.33\linewidth]{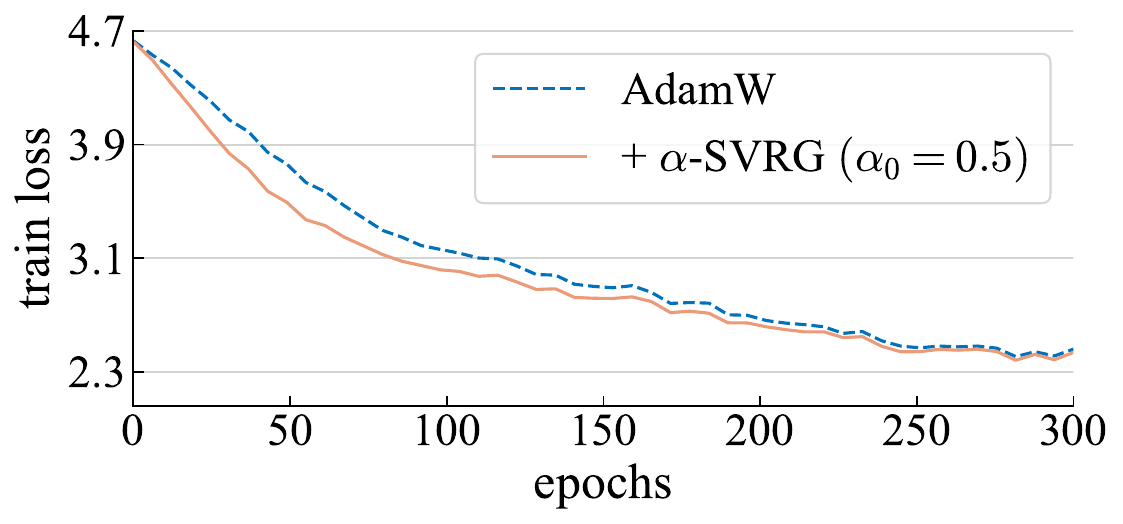}~
    \includegraphics[width=.33\linewidth]{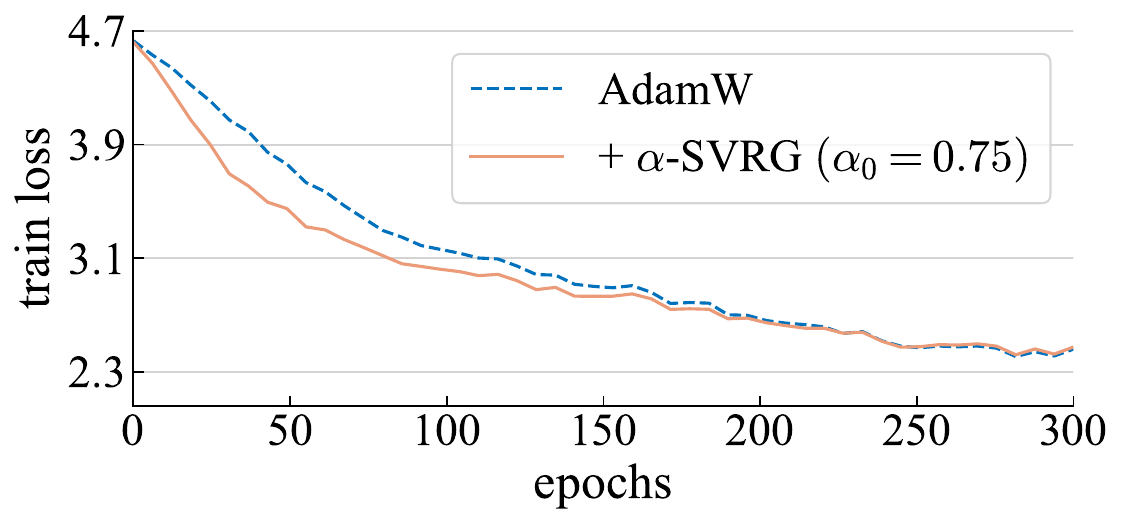}~
    \includegraphics[width=.33\linewidth]{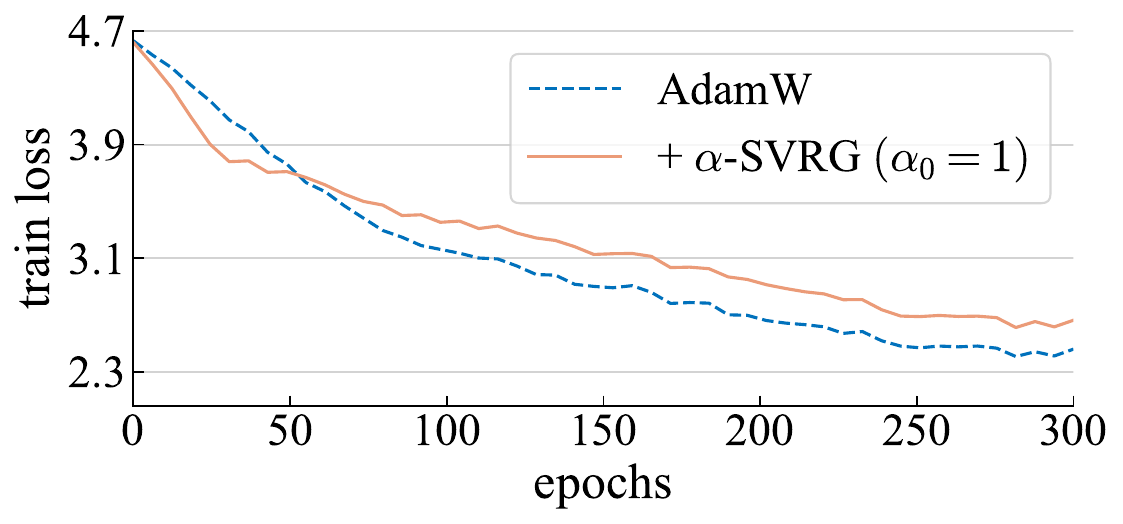}
    \vspace{-1.5em}
    \caption{Food-101}
\end{subfigure}
\begin{subfigure}[h]{\linewidth}
    \centering
    \includegraphics[width=.33\linewidth]{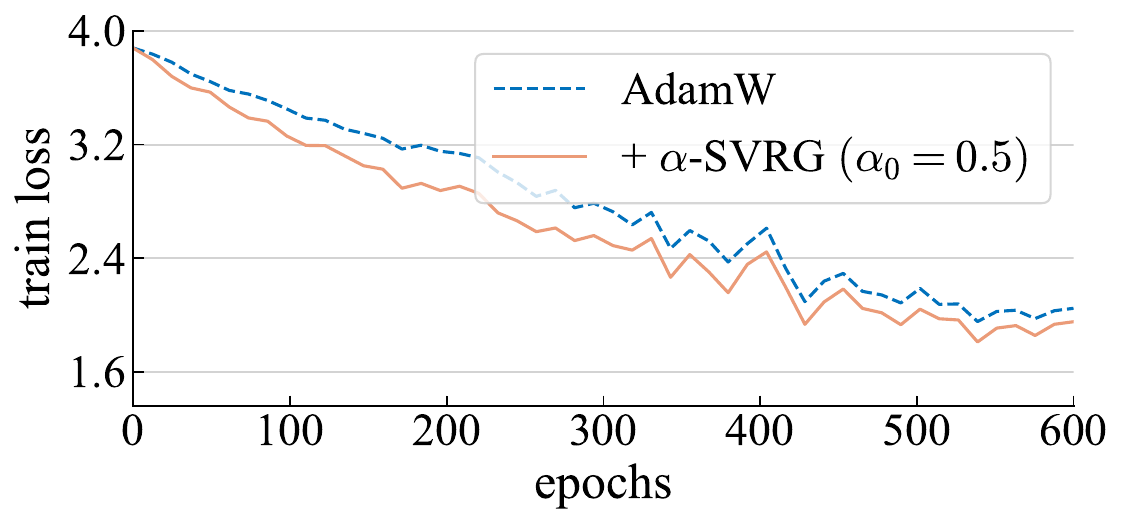}~
    \includegraphics[width=.33\linewidth]{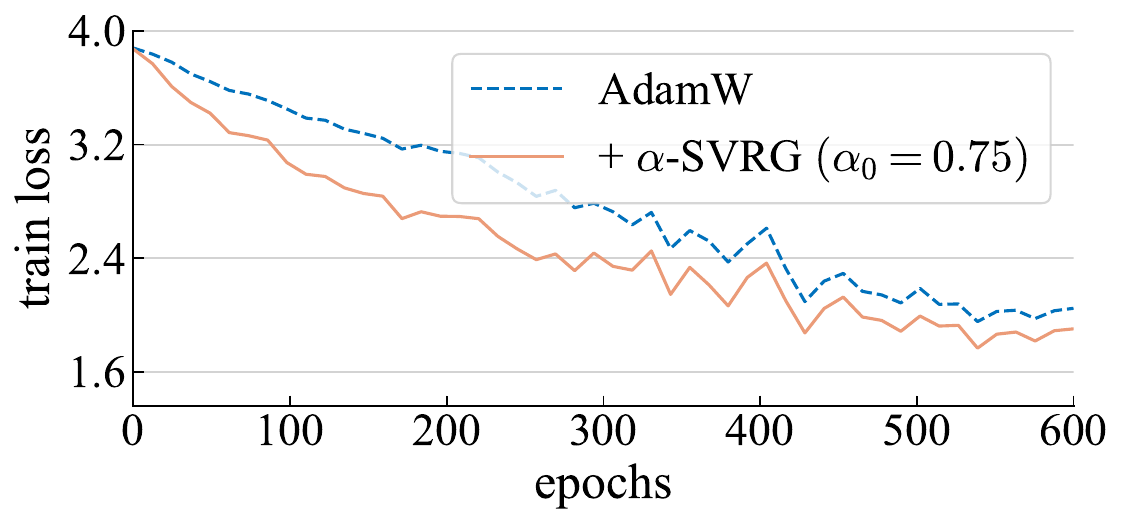}~
    \includegraphics[width=.33\linewidth]{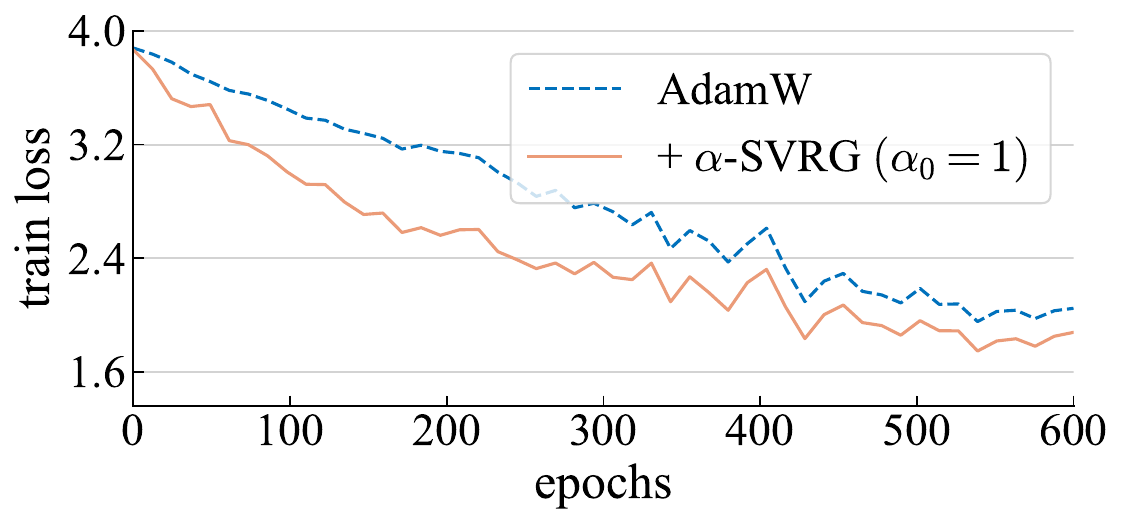}
    \vspace{-1.5em}
    \caption{DTD}
\end{subfigure}
\begin{subfigure}[h]{\linewidth}
    \centering
    \includegraphics[width=.33\linewidth]{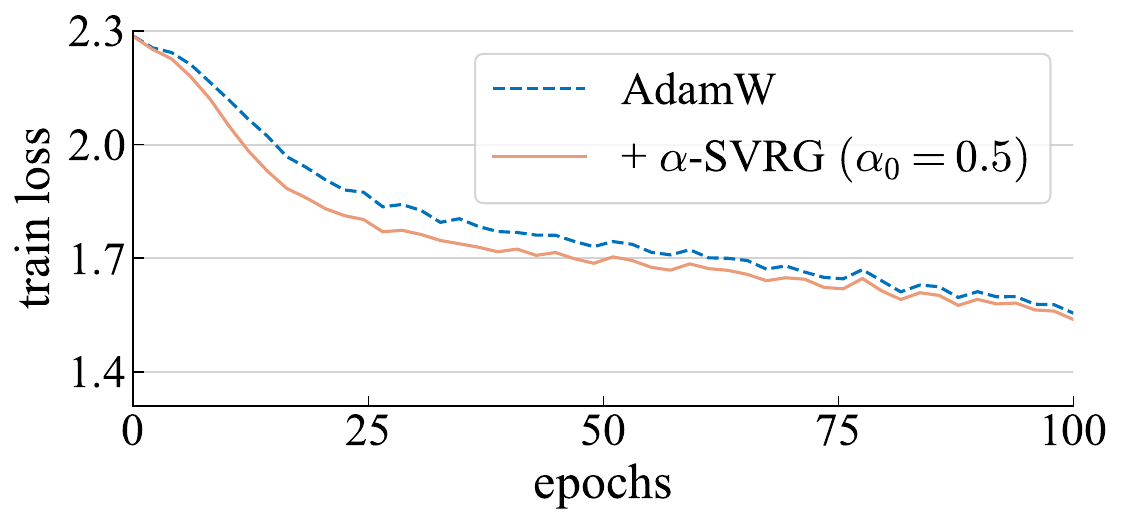}~
    \includegraphics[width=.33\linewidth]{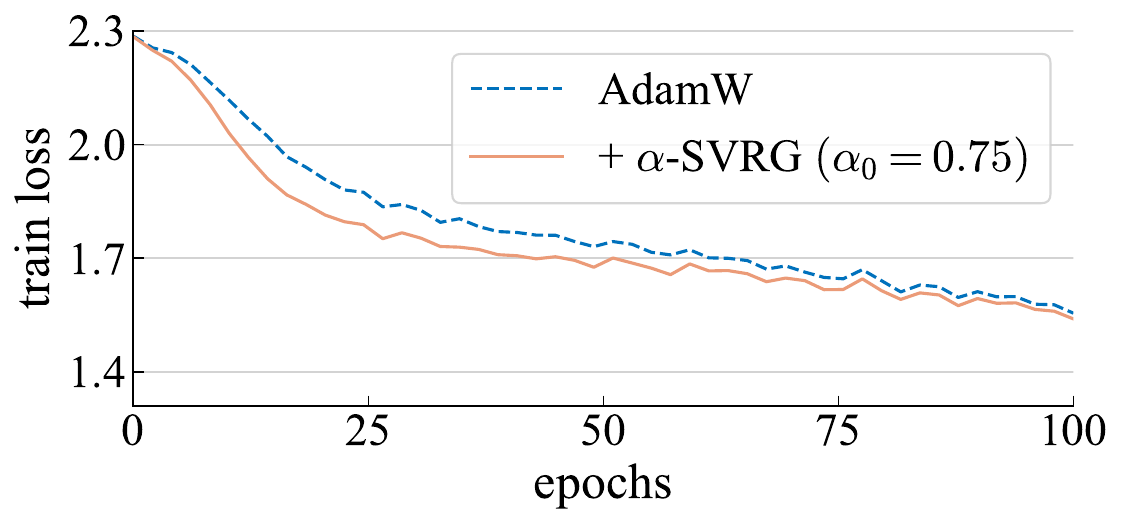}~
    \includegraphics[width=.33\linewidth]{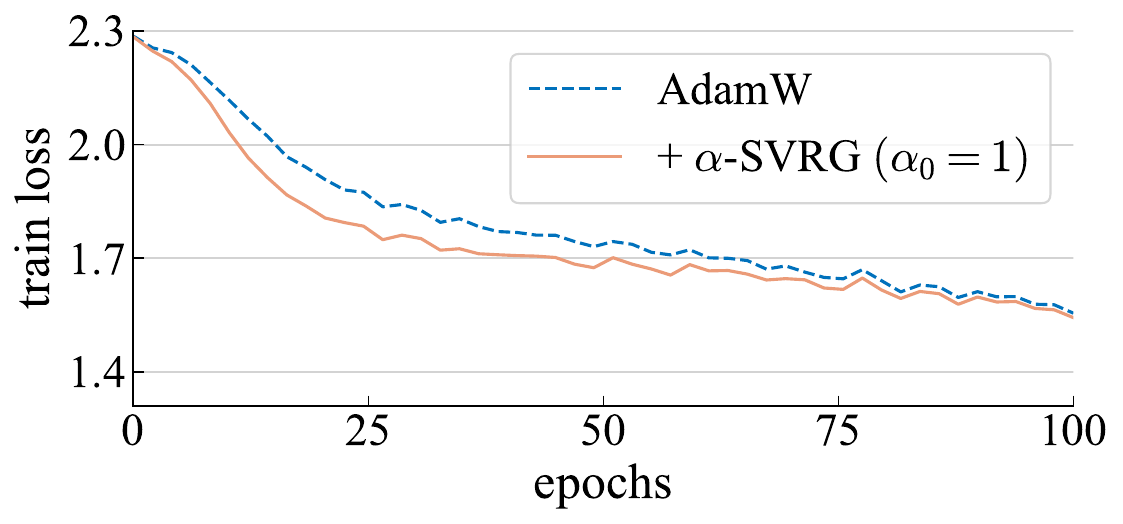}
    \vspace{-1.5em}
    \caption{SVHN}
\end{subfigure}
\begin{subfigure}[h]{\linewidth}
    \centering
    \includegraphics[width=.33\linewidth]{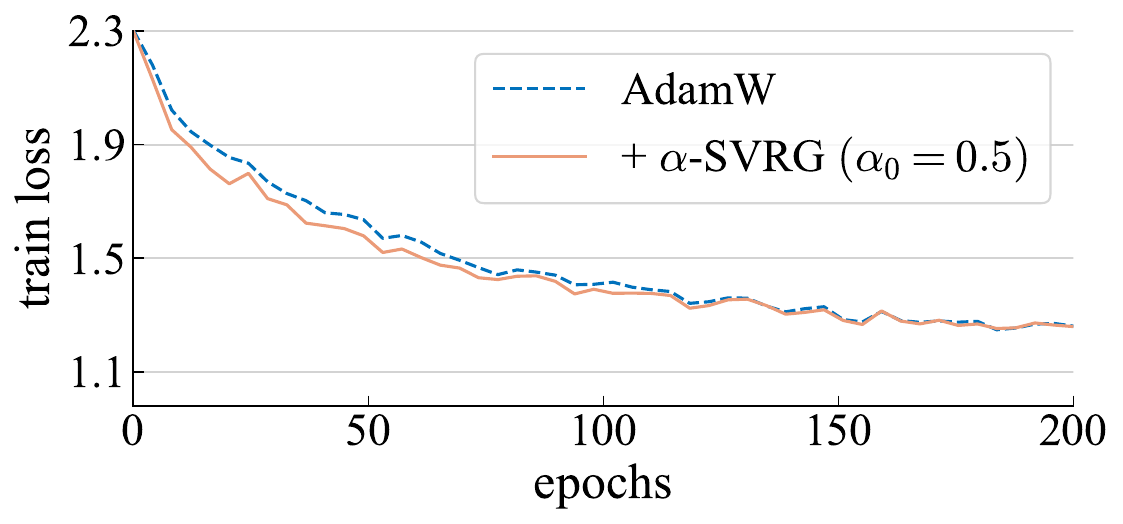}~
    \includegraphics[width=.33\linewidth]{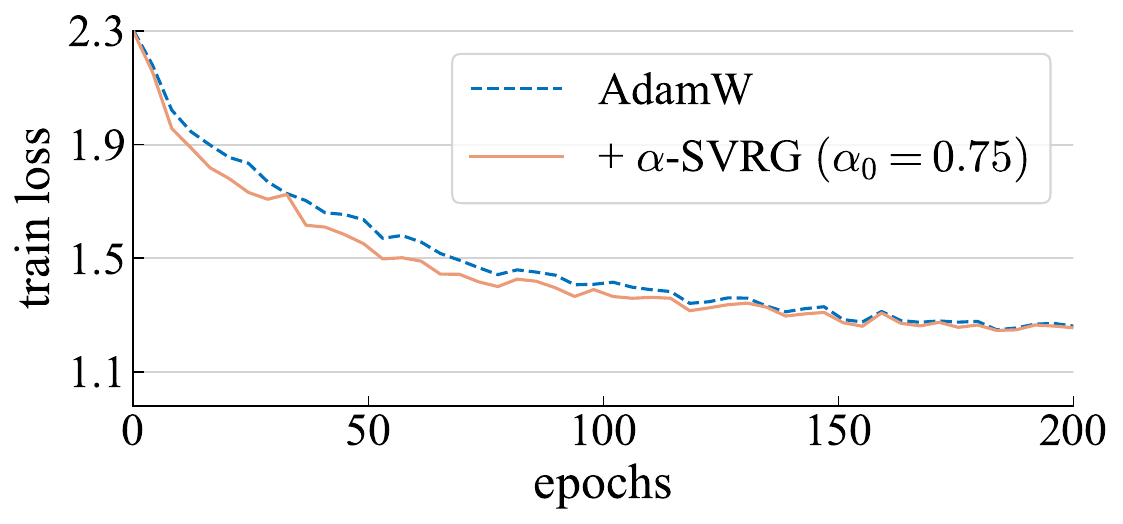}~
    \includegraphics[width=.33\linewidth]{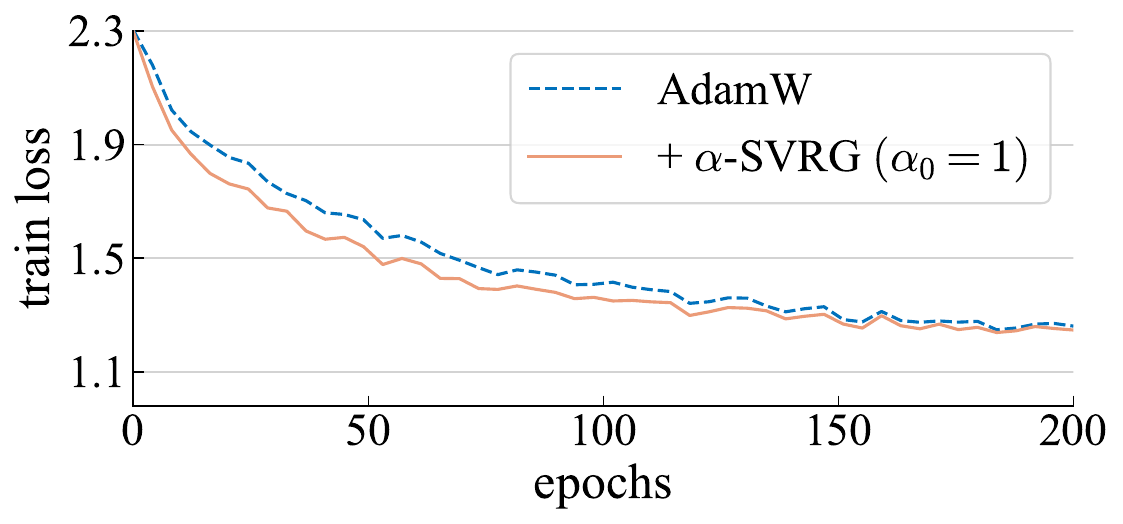}
    \vspace{-1.5em}
    \caption{EuroSAT}
\end{subfigure}
\caption{\textbf{Training loss with AdamW and {$\boldsymbol{\alpha}$-SVRG} on smaller classification datasets (Table~\ref{tab:last}).}}
\label{fig:convergence1}
\end{figure}

We compare the training loss convergence of the AdamW baseline and $\alpha$-SVRG on ImageNet-1K (Figure~\ref{fig:convergence2}) and small classification datasets (Figure~\ref{fig:convergence1}). It is observed that {\approach} can consistently decrease training loss and deliver faster convergence.

\clearpage
\newpage
\section{Pseudocode for \approach}
\label{appendix:pseudocode}
For clarity, we adopt the following notations: $s$ as the epoch index, $t$ as the iteration index within each epoch, $i_t$ as the sampled index at iteration $t$, $T$ as the epoch length, $M$ as the iteration length within each epoch, $n$ as the total number of training data, and $\eta^s_t$ as the learning rate. We illustrate {\approach} with SGD in Algorithm~\ref{alg:svrg} and with AdamW in Algorithm~\ref{alg:svrg2}: 
\begin{algorithm}[h]\small
   \caption{$\alpha$-SVRG with SGD$\left(\vtheta^0_0,T, M, \{\{\eta^s_t\}_{t=0}^{M-1}\}_{s=0}^{T-1}, \{\{\alpha_t^s\}_{t=0}^{M-1}\}_{s=0}^{T-1},\lambda\right)$}
   \label{alg:svrg}
   \setstretch{1.2} 
\begin{algorithmic}[0]
   \STATE {\bfseries Input:} initialized model parameters $\vtheta^0_0$, epoch length $T$, iteration length within each epoch $M$, learning rates $\{\{\eta^s_t\}_{t=0}^{M-1}\}_{s=0}^{T-1}$, scheduled coefficients $\{\{\alpha_t^s\}_{t=0}^{M-1}\}_{s=0}^{T-1}$, weight decay $\lambda$
   \STATE $\vtheta^{0}_\text{past} = \vtheta^0_0$
   \FOR{$s=0$ {\bfseries to} $T-1$}
   \STATE $\nabla f(\vtheta^s_\text{past}) = \frac{1}{n} \sum_{i=1}^n \nabla f_{i}(\vtheta^s_\text{past})$
   \FOR{$t=0$ {\bfseries to} $M-1$}
   \STATE $i_t\overset{\mathrm{iid}}{\sim}\text{Uniform}\{1,\cdots,n\}$ 
   \STATE $\vg^s_t =  \nabla f_{i_t}(\vtheta^s_t) - \alpha^s_t \left(\nabla f_{i_t}(\vtheta^s_\text{past}) - \nabla f(\vtheta^s_\text{past})\right)$ 
   \STATE $\vtheta^s_{t+1} = \vtheta^s_{t} - \eta^s_t \vg^s_t - \lambda \vtheta^s_t$
   \ENDFOR
   \STATE $\vtheta^{s+1}_\text{past} = \vtheta^{s}_M$
   \ENDFOR
   \STATE {\bfseries Output:} final model $\vtheta^{T-1}_M$.
\end{algorithmic}
\end{algorithm}

\begin{algorithm}[h]\small
   \caption{$\alpha$-SVRG with AdamW$\left(\vtheta^0_0,T, M, \{\{\eta^s_t\}_{t=0}^{M-1}\}_{s=0}^{T-1}, \{\{\alpha_t^s\}_{t=0}^{M-1}\}_{s=0}^{T-1},\beta_1,\beta_2,\lambda\right)$}
   \label{alg:svrg2}
   \setstretch{1.2} 
\begin{algorithmic}[0]
   \STATE {\bfseries Input:} initialized model parameters $\vtheta^0_0$, epoch length $T$, iteration length within each epoch $M$, learning rates $\{\{\eta^s_t\}_{t=0}^{M-1}\}_{s=0}^{T-1}$, scheduled coefficients $\{\{\alpha_t^s\}_{t=0}^{M-1}\}_{s=0}^{T-1}$, momentums $\beta_1,\beta_2$, weight decay $\lambda$
   \STATE $\vtheta^{0}_\text{past} = \vtheta^0_0$
   \STATE $\vm=\vv=0$
   \FOR{$s=0$ {\bfseries to} $T-1$}
   \STATE $\nabla f(\vtheta^s_\text{past}) = \frac{1}{n} \sum_{i=1}^n \nabla f_{i}(\vtheta^s_\text{past})$
   \FOR{$t=0$ {\bfseries to} $M-1$}
   \STATE $i_t\overset{\mathrm{iid}}{\sim}\text{Uniform}\{1,\cdots,n\}$ 
   \STATE $\vg^s_t =  \nabla f_{i_t}(\vtheta^s_t) - \alpha^s_t \left(\nabla f_{i_t}(\vtheta^s_\text{past}) - \nabla f(\vtheta^s_\text{past})\right)$ 
   \STATE $\vm = \beta_1 \vm + (1-\beta_1)\vg^s_{t}$
   \STATE $\vv = \beta_2 \vv + (1-\beta_2)(\vg^s_{t})^2$
   \STATE $\vtheta^s_{t+1} = \vtheta^s_{t} - \eta^s_t \frac{\vm}{\sqrt{\vv}+\epsilon} - \lambda \vtheta^s_t$
   \ENDFOR
   \STATE $\vtheta^{s+1}_\text{past} = \vtheta^{s}_M$
   \ENDFOR
   \STATE {\bfseries Output:} final model $\vtheta^{T-1}_M$.
\end{algorithmic}
\end{algorithm}

\end{document}